\documentclass{article}

\PassOptionsToPackage{numbers, compress}{natbib}

\usepackage[main, final]{neurips_2026}

\usepackage[utf8]{inputenc} 
\usepackage[T1]{fontenc}    
\usepackage{hyperref}       
\usepackage{url}            
\usepackage{booktabs}       
\usepackage{amsfonts}       
\usepackage{amsmath}
\usepackage{nicefrac}       
\usepackage{microtype}      
\microtypesetup{verbose=false}  
\usepackage{xcolor}         
\usepackage{graphicx}
\usepackage{float}          
\usepackage{algorithm}
\usepackage{algpseudocode}
\newcommand{\todo}[1]{\textcolor{magenta}{#1}}
\usepackage{mathtools}
\usepackage{multirow} 
\usepackage{xcolor}

\definecolor{myviolet}{rgb}{0.6, 0.4, 0.8}

\title{On the Design Space of Discrete Diffusion Online Adaptation for Molecular Optimization}

\author{
  \begin{tabular}[t]{c}
    Trevor Chen$^{1\ast}$ \enspace Ariel Dai$^{1\ast}$ \enspace Jason Yang$^1$ \enspace Riccardo De Santi$^2{^,}^3$ \\[0.3em]
    Daniel Khalil$^1$ \enspace Wenda Chu$^1$ \enspace Nate Gruver$^4$ \enspace Pranav Murugan$^4$ \\[0.3em]
    Alexander F.~G.~Goldberg$^4$ \enspace Maruan Al-Shedivat$^4$ \enspace Yisong Yue$^1$ \\[0.5em]
    \normalfont\normalsize $^1$Caltech \quad $^2$ ETH Zurich \quad $^3$ ETH AI Center \quad $^4$Genesis Molecular AI \\
  \end{tabular}
}

\begin{document}

\maketitle
\begin{NoHyper}
\renewcommand{\thefootnote}{}
\footnotetext{$^\ast$Equal contribution. \quad Code: \url{https://github.com/trevorbchen/DDAMO}. Results: \url{https://huggingface.co/trevorbchen/DDAMO}. Correspondence: \texttt{yyue@caltech.edu} }
\renewcommand{\thefootnote}{\arabic{footnote}}
\end{NoHyper}

\vspace{-0.1in}
\begin{abstract}

\looseness -1 
Molecular optimization often starts from a pretrained generative model that captures a broad prior over valid molecular structures. At test time, however, the goal is not to sample from this prior, but to use a limited oracle budget to shift generation toward task-specific high-reward molecules. We study this adaptation problem for discrete diffusion models. Each online round couples several choices. The loop must decide which candidates to evaluate, how rewards become model updates, which feedback to reuse, and how far to move beyond the pretrained prior. These choices have mostly been studied in isolation, leaving open whether they complement one another, become redundant, or interfere inside a full online adaptation loop. We conduct controlled studies across six small-molecule binding-affinity tasks and three protein-fitness tasks. We find that acquisition, reward shaping, and model debiasing provide complementary routes to higher reward, especially for small molecules. Replay further stabilizes learning, while validity penalties keep small-molecule exploration on the valid molecular manifold. Together, these findings point to a practical recipe for feedback-efficient molecular optimization: online fine-tuning with acquisition, reward shaping, debiasing, replay, and validity control. This recipe outperforms offline fine-tuning and inference-time search baselines under matched oracle-call budgets and GPU-hour accounting. The gains are largest when high-reward candidates require larger shifts from the pretrained prior.

\end{abstract}

\section{Introduction}

Discrete diffusion models have become a powerful tool for molecular generation, such as for small molecules \citep{leeGenMolDrugDiscovery2025, schiffSimpleGuidanceMechanisms2025} and proteins \citep{wangDiffusionLanguageModels2024}, because they can model realistic chemical structures directly while producing diverse candidates.
In molecular design, however, unconditional generation is rarely useful. 
A pretrained generator $p_{\theta_0}$ encodes a broad molecular prior whose probability mass approximates the pretraining distribution, while a downstream design task defines a different target: sampling molecules achieving high rewards for a task-specific oracle $r(x)$. This oracle, which is typically costly to evaluate, might be a learned binding-affinity model \citep{jiangFlashAffinityBridgingAccuracySpeed}, a structure-prediction, a docking pipeline \citep{passaro2025boltz2}, or a wet-lab assay. Under a fixed evaluation budget, the typical goal of molecular optimization is to generate a molecule $x^*$ best-optimizing the given reward function:
\begin{equation}
    x^* \in \arg\max_{x \in \mathcal{X}} \; r(x) \label{eq:design_objective}
\end{equation}
where $\mathcal{X}$ denotes the valid chemical space under consideration.
This formulation corresponds to top-$1$ discovery objective of the unknown reward function $r$, a problem also denoted by best-arm identification~\citep[e.g.,][]{lattimore2020bandit}. To tackle such generative optimization problems, the generator must be steered toward task-relevant regions of chemical space while preserving prior validity \citep{gutjahr2025constrained, HowArtificialIntelligence,  ueharaInferenceTimeAlignmentDiffusion2025}.
A powerful way to achieve this is via online model adaptation \citep{yangSteeringGenerativeModels2025, stoccoGuidingGenerativeModels2025, ueharaFeedbackEfficientOnline2024, rector-brooksSteeringMaskedDiscrete2024, suIterativeDistillationRewardGuided2025, ueharaRewardGuidedIterativeRefinement2025, yangSteeringGenerativeModels2025, stoccoGuidingGenerativeProtein2025, stoccoGuidingGenerativeModels2025, liCAGenMolConditionAwareDiffusion2026, olivecronaMolecularNovoDesign2017, blalockFunctionalAlignmentProtein2025}, as described next.

\textbf{Online Model Adaptation for Optimization:  Loop Structure.} 
\looseness -1 Given a pre-trained model, each round $t$ of online model adaptation alternates between two high-level steps (Fig.~\ref{fig:test_time_feedback}). A candidate pool $C_t$ of size $M$ is sampled from the current generator $p_{\theta_t}$, optionally with inference-time steering, and a batch $S_t$ of size $K$ is selected for oracle evaluation. The resulting pairs $(x,r(x))$ are appended to the evaluated set. Then the observed scores are converted into training rewards $\tilde{r}(x)$, optionally regularized, and used to update $p_{\theta_t}$ into $p_{\theta_{t+1}}$. Oracle feedback therefore has two roles: it determines the current discovery utility and, through model adaptation, changes which candidates the loop will propose in subsequent rounds.




\textbf{A Tightly Coupled Design Problem.}
\looseness -1 The loop exposes tightly coupled choices that are often studied separately: which candidates receive oracle evaluations, how rewards are converted into model updates, and how the generator is pushed beyond biases in the pretrained prior. Previous works show that: Thompson sampling can improve oracle-call allocation through uncertainty-aware exploration~\citep[][]{ueharaFeedbackEfficientOnline2024, lattimore2020bandit}; top-$1$/$K$ or CVaR-style objectives can focus updates on the reward tail~\citep[e.g.,][]{santiFlowDensityControl, wang2026efficient}; and negative-score regularization can debias the generator away from over-occupied modes~\citep[][]{de2025provable, santiFlowDensityControl, santiVerifierConstrainedFlowExpansion2026}.
It is unclear whether these choices for acquisition, feedback reuse, reward shaping, and prior debiasing complement one another, become redundant, or interfere inside a full online adaptation loop.

To address this gap, we conduct an integrated study of representative design choices in online adaptation loops for discrete diffusion based molecule and protein optimization:

\begin{figure}[t]
  \centering
    \includegraphics[width=\textwidth]{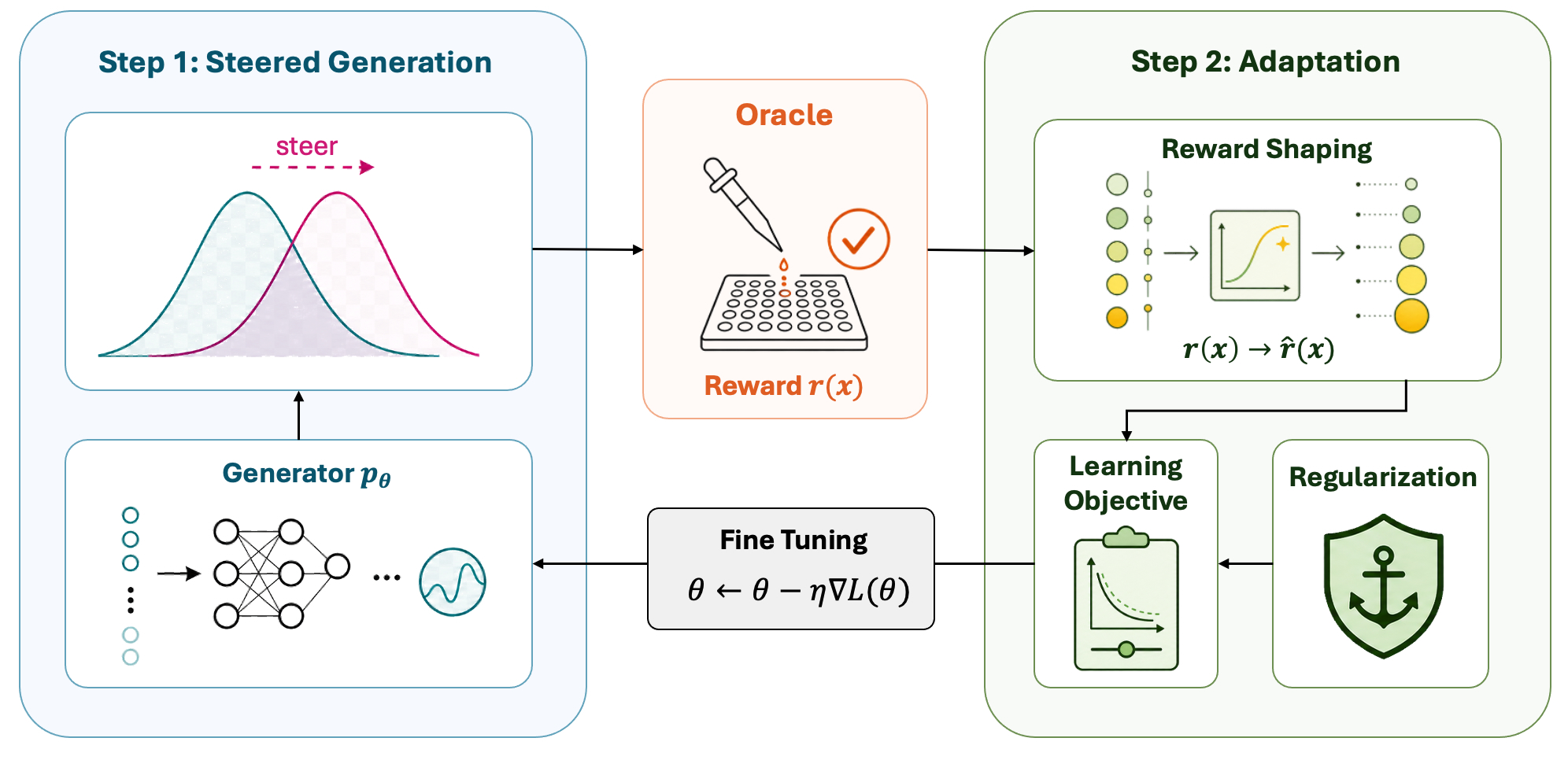}
    \vspace{-0.2in}
      \caption{\textbf{Online feedback loop for molecular optimization.}
Each round consists of two steps:
(1) the current generator $p_{\theta_t}$, updated using oracle feedback in Step~(2) of the previous round, generates candidates for the current round;
(2) the observed oracle rewards are converted into training signals to update the generator from $p_{\theta_t}$ to $p_{\theta_{t+1}}$.
The updated generator then returns to Step~(1), closing the online feedback loop.}
  \label{fig:test_time_feedback}
  \vspace{-0.1in}
\end{figure}

Our experiments on 9 tasks spanning small-molecule binding-affinity and protein-fitness optimization show that these mechanisms provide complementary benefits, especially when high-reward candidates have low likelihood under the pretrained generative prior. 
On small molecules, acquisition, CVaR reward shaping, and Density Entropy Regularization provide the clearest complementary gains. On protein-fitness tasks, ablation effects are weaker, consistent with family-specific priors already covering high-fitness regions. Replay and validity control are primarily for keeping useful feedback across rounds and preventing exploratory updates from leaving the valid molecular manifold. 

\textbf{Overall, our contributions are: }
\begin{itemize}
    \item We provide a controlled empirical study of discrete-diffusion online adaptation for molecular optimization alongside offline fine-tuning and inference-time-search methods under fixed oracle and runtime budgets.

    \item We characterize the interactions among simple \emph{plug-in components} added to the online optimization loop through thorough ablations and mechanistic analysis: DER, CVaR, and Thompson Sampling exhibit compounding interactions while replay and validity control address distinct failure modes.

    \item We identify prior--reward mismatch as a useful diagnostic for when these mechanisms provide the most benefit. Under these observations, we assemble an online active fine-tuning recipe that substantially improves feedback efficiency over baselines.
    
\end{itemize}

\section{Plug-In Components of Online Adaptation} 
\label{sec:design-space}

We frame the loop in Fig.~\ref{fig:test_time_feedback} as a \emph{harness}: a fixed online schedule with swappable components at the points where feedback is collected, transformed, and used for fine-tuning. Each round generates candidates, selects a subset for oracle evaluation, converts oracle scores into training rewards, regularizes the update, and fine-tunes the generator. As a \emph{harness}, it is finetuner-agnostic and domain-agnostic: any discrete generator with objective $\mathcal{L}_{\text{ft}}$ that consumes per-sample log-rewards and an off-policy buffer can be used; we instantiate it with DDPP-LB~\citep{rector-brooksSteeringMaskedDiscrete2024} and VIDD~\citep{suIterativeDistillationRewardGuided2025}.

We organize the components by their role in the feedback loop. \emph{Steering strategies} affect which candidates receive oracle evaluations (Sec.~\ref{sec:design-explore}); \emph{adaptation strategies} determine how observed rewards shape the model update (Sec.~\ref{sec:adaptation_design}); and \emph{objective-mismatch corrections} keep optimization aligned with accumulated feedback and valid molecules as exploration shifts the sampling distribution $p_{\theta_t}$ (Sec.~\ref{sec:design-objmismatch}). For each category, we instantiate one representative choice adapted from prior work (Table~\ref{tab:methods}). The per-knob ablations in Sec.~\ref{sec:results} then test whether these choices retain distinct value when composed.

\begin{table}[h]
\centering
\small
    \begin{tabular}{@{}p{4.4cm}p{9.3cm}@{}}
\toprule
\textbf{Knob} & \textbf{Description} \\
\midrule
$+$ CVaR~\citep{santiFlowDensityControl, wang2026efficient}            & Replace $r$ with $(r - Q_\tau)_+/(1-\tau)$, $\tau{=}0.8$ \\
$+$ Density Entropy Reg.~\citep{de2025provable}  & Add $-\gamma \log p_\theta(x)$ to log-reward, $\gamma{=}1$ \\
$+$ Thompson~\citep{thompson1933likelihood, chapelle2011empirical}        & $J{=}10$ MLP ensemble; oracle top-$K$ by $s_i \sim \mathcal{N}(\mu_i, \sigma_i^2)$ \\
$+$ Replay buffer~\citep{mnih2015human}   & Fixed-capacity FIFO with priority eviction; stratified sampling \\
$+$ Invalid penalty~\citep{wangFineTuningDiscreteDiffusion2024, ueharaELEGANT2024}  & Assign $r {=} r_{\text{inv}}$ to SMILES that fail RDKit parsing \\
\bottomrule
\end{tabular}
\caption{\textbf{\emph{Harness} knobs: plug-in components evaluated in this paper.} Each knob layers on top of finetuner and is ablated independently in Section~\ref{sec:results}, then evaluated jointly in the final recipe of Sec.~\ref{sec:results-method}.}
\label{tab:methods}
\vspace{-0.2in}
\end{table}

\subsection{Steering Strategies}\label{sec:design-explore}

\begin{figure}[b]
  \centering
  \begin{minipage}[t]{0.64\textwidth}
    \vspace{0pt}
    \centering
    \includegraphics[width=\linewidth]{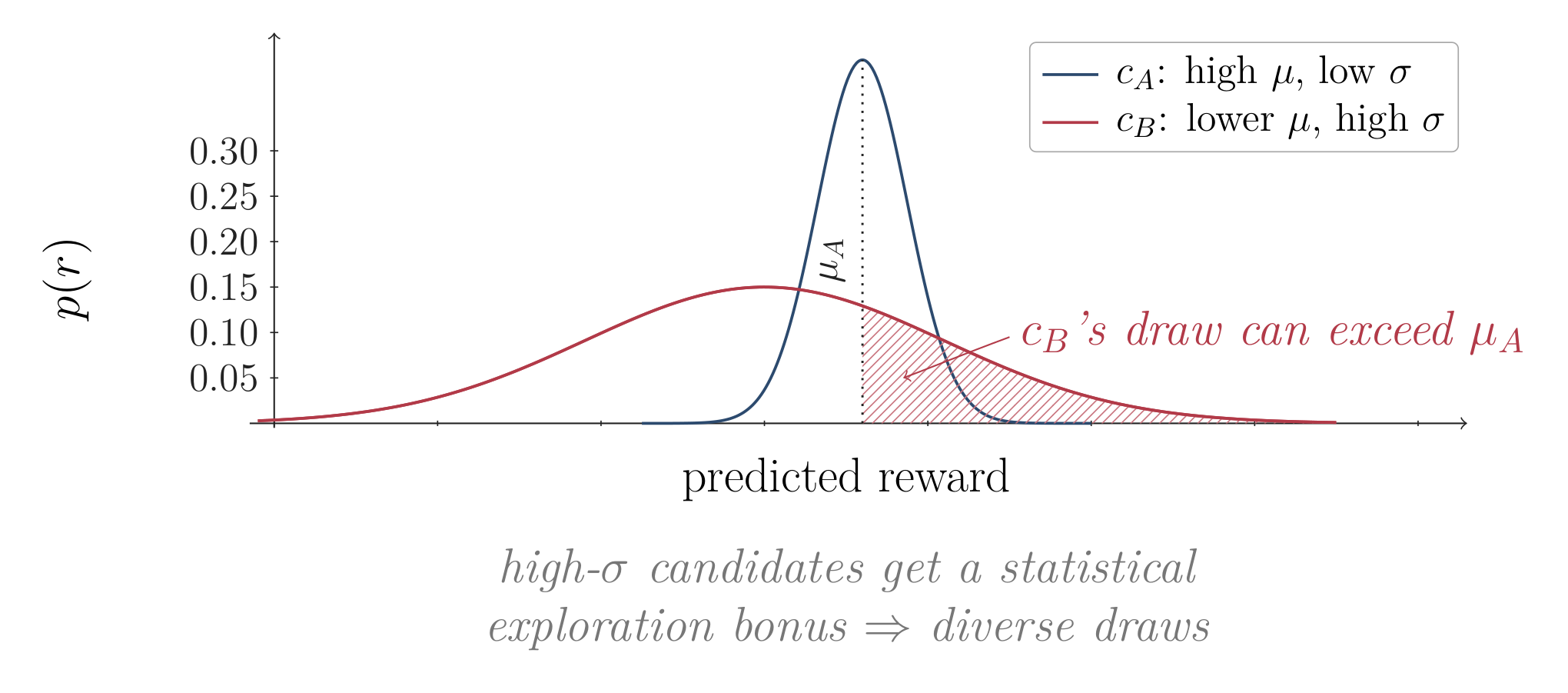}
  \end{minipage}
  \hfill
  \begin{minipage}[t]{0.35\textwidth}
    \vspace{0pt}
    \caption{\textbf{Thompson sampling.} Although $c_A$ has the higher mean prediction, $c_B$ can be selected because its larger uncertainty gives it nonzero probability of drawing above the greedy threshold.}
    \label{fig:diag_thompson_body}
  \end{minipage}
\end{figure}

\textbf{Thompson Sampling Acquisition.} Because the oracle budget is finite, the loop must choose a small subset of generated candidates for evaluation. The simplest rule evaluates the candidates with the highest predicted reward under a surrogate model. This greedy rule is purely exploitative: it spends oracle calls where the surrogate already predicts high reward. We also consider uncertainty-aware acquisition using Thompson sampling \citep{thompson1933likelihood, russo2018tutorial}. Given an ensemble surrogate with mean $\mu_i$ and disagreement $\sigma_i$ for candidate $x_i\in C_t$, Thompson sampling draws a sampled reward estimate:
\[
\hat r_i = \mu_i + \varepsilon_i \sigma_i, \qquad \varepsilon_i \sim \mathcal{N}(0,1),
\]
and evaluates the top-$K$ candidates by $\hat r_i$, defining the selected batch $S_t$. Fig.~\ref{fig:diag_thompson_body} depicts the intuition behind how Thompson sampling balances the explore/exploit trade-off.  Candidates with high uncertainty can outrank candidates with higher predicted mean but lower uncertainty, i.e., this scheme incentivizes to explore seemingly lower quality but high uncertainty candidates. 

\subsection{Adaptation Strategies}\label{sec:adaptation_design}

\begin{figure}[htbp]
  \centering
  \begin{minipage}[c]{0.49\textwidth}\centering
    \includegraphics[width=\textwidth]{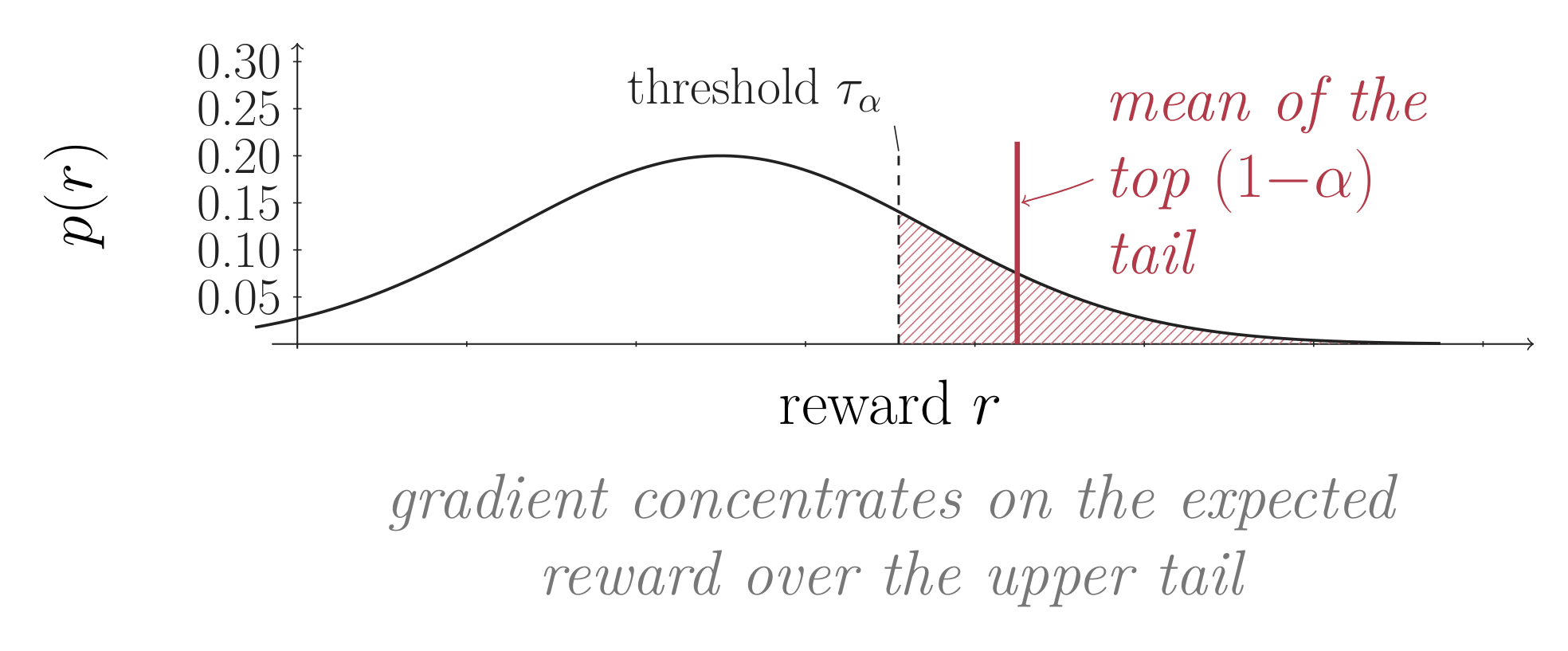}\\
    \textbf{(a)} CVaR Shaping
  \end{minipage}\hfill
  \begin{minipage}[c]{0.49\textwidth}\centering
    \includegraphics[width=\textwidth]{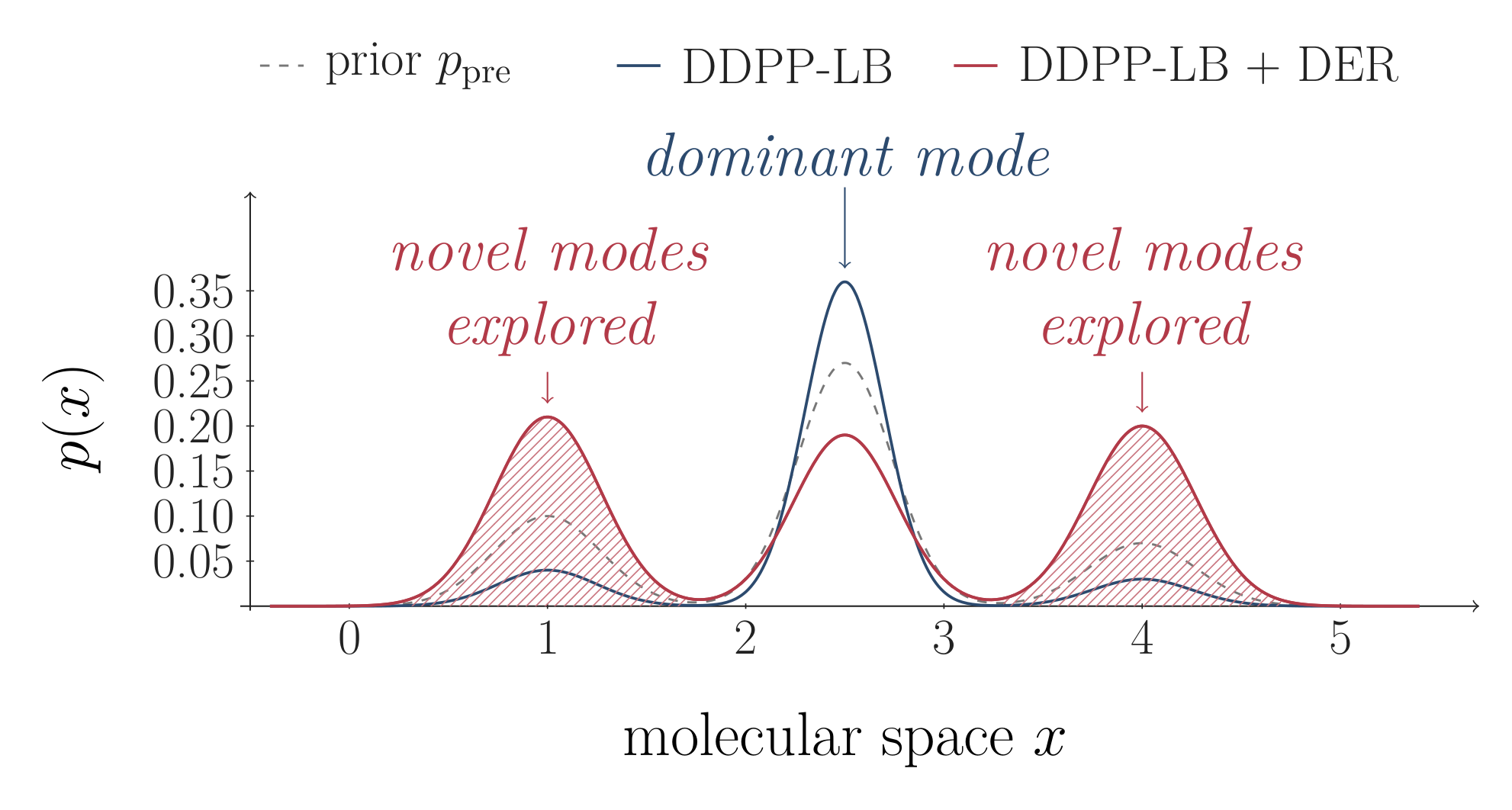}\\
    \textbf{(b)} Density Entropy Regularization
  \end{minipage}
  \caption{\textbf{Reward-shaping and model debiasing} \textbf{(a)} CVaR upper-tail shaping: the right-tail of the reward distribution above the $\tau$-quantile (dashed) is shaded, and the solid red bar marks $\mathrm{CVaR}_{1-\tau} = \mathbb{E}[r \mid r \geq Q_\tau]$, the expected reward conditional on being in that tail. Optimization concentrates gradient on this tail expectation, not on a hard top-$k$ selection. \textbf{(b)} Density Entropy Regularization (model debiasing): subtracting the current generator's diffusion-internal likelihood $\log p_{\theta_t}(x)$ from the log-reward redistributes gradient mass off densely-occupied modes and toward less-frequented regions, exposing minor modes the unshaped reward would have kept under-explored.}
  \label{fig:diag_reward_shaping}
  \label{fig:diag_cvar_body}
  \label{fig:diag_debiasing_body}
\end{figure}

\textbf{CVaR-Style Reward Shaping.}
\looseness -1 A second lever acts on the reward itself after oracle evaluation and before it enters the gradient. We write the resulting shaped training reward as $\tilde r(x)$. Because the design objective in Eq. \eqref{eq:design_objective} is controlled by the upper tail rather than the average generated molecule, we adapt the model via a CVaR-style \citep{rockafellar2000cvar, rockafellar2002cvar} objective (Fig.~\ref{fig:diag_reward_shaping}(a)), as introduced in prior work on diffusion model adaptation~\citep{santiFlowDensityControl, wang2026efficient}. This objective allows to sacrifice the average sample reward to enhance the quality of top-samples:
\[
\mathrm{CVaR}_{1-\tau}[r(x)]
=
\mathbb{E}\!\left[
r(x) \mid r(x) \ge Q_\tau(r)
\right].
\]

\textbf{Density Entropy Regularization.}
CVaR reward shaping focuses the update on the upper tail, but it does not distinguish between high-reward regions that the generator already samples frequently and high-reward regions that remain underexplored. To expose this distinction independently of CVaR, practitioners can apply a density-entropy regularizer~\citep{de2025provable, santiFlowDensityControl, santiVerifierConstrainedFlowExpansion2026} that uses the current model density as a measure of how occupied a region is under the generator and shapes the raw log-reward as:
\begin{equation}
\log \tilde r_{\mathrm{debias}}(x)
=
\log r(x) - \gamma \log p_{\theta_t}(x),
\label{eq:debias}
\end{equation}
illustrated in Fig.~\ref{fig:diag_reward_shaping}(b), where $\log p_{\theta_t}(x)$ is the marginal. In continuous-diffusion analyses~\citep{de2025provable, santiFlowDensityControl}, the corresponding negative-score gradient $\nabla_x \log p_{\theta_t}(x)$ can be directly approximated by the available score network,  but in masked discrete diffusion it is estimated by the marginalization:
\[
p_{\theta_t}(x_0) \;=\; \mathbb{E}_{t, \, x_t \sim q(x_t \mid x_0)}\!\bigl[\,p_{\theta_t}(x_0 \mid x_t)\,\bigr] \cdot \text{(noise-schedule weight)},
\]
a sum over an exponentially-large set of noised states $x_t$, making it intractable. For discrete diffusion the proper asymptotic unbiased estimator is a Monte-Carlo ELBO \citep{nie2025llada} (LLaDA, Algorithm~3), which costs $n_{\text{mc}}$ extra forward passes per evaluation.

We replace $\log p_{\theta_t}(x)$ in Eq.~\ref{eq:debias} with the cheapest such estimator: 
\[
\widehat{\log p_{\theta_t}(x)} \;\equiv\; \log p_{\theta_t}(x_0 \mid x_t)
\;=\;
\textstyle\sum_{i \,:\, x_t^i = [\text{MASK}]}\, \log \mathrm{softmax}(\ell_i)\bigl[x_0^i\bigr].
\]
This estimator is a cheap approximation of the diffusion ELBO, so it is not equivalent to the MCMC-estimated continuous negative-score regularizer or to the LLaDA-style $L/l$-weighted estimator in the limit. We choose this estimator  for computational reasons: the alternatives cost $n_{\text{mc}}\!\times$ extra forward passes per gradient step, which is prohibitive across thousands of online finetuning steps. Despite the approximation, we find empirically that it induces the intended distribution shift and that the added noise leads to more exploration and overall better performance. we find empirically that it achieves comparable oracle-call efficiency to LLaDA-Alg.\,3 variants with $n_{\text{mc}} \in \{1, 4, 8, 16\}$ while being substantially faster per GPU hour. Post-hoc evaluation with the unbiased LLaDA-Alg.\,3 estimator ($n_{\text{mc}}{=}32$) confirms that the resulting NLL distributions from the cheap estimator follows closely to the LLaDA estimator with wider exploration (Appendix Figs.~\ref{fig:negscore_estimator_curves} and~\ref{fig:negscore_nll_dist}).
Intuitively, this term favors candidates that are high reward but not already high likelihood under the generator, complementing Thompson-sampling acquisition (acquisition vs.\ reward-shape act on the loop at different points; see Appendix~\ref{app:knob-interactions} and Fig.~\ref{fig:diag_debiasing_body}).
The estimated value is detached, so the parameter gradient enters the standard DDPP-LB / VIDD policy-gradient term as a scalar reward modifier  (Appendix~\ref{app:method-details}).

\subsection{Objective-mismatch corrections}\label{sec:design-objmismatch}

Once exploration starts shifting $p_{\theta_t}$ off the prior's dominant modes, two gaps open between what the gradient actually optimizes and the true objective. The next two components can close them.

\textbf{Replay Buffer.}
A remaining choice is which evaluated samples appear in each update. A purely on-policy update trains only on the latest batch, so useful molecules discovered early stop influencing the generator once the sampling distribution moves on. We therefore optionally train from a replay buffer that mixes fresh candidates with retained high-reward samples \citep{mnih2015human, schaul2016prioritized, olivecronaMolecularNovoDesign2017}. Concretely, the buffer is fixed-capacity FIFO with priority eviction (when full, the lowest-reward sample leaves first, not the oldest), and each minibatch is drawn by stratified sampling across reward bins so high-reward outliers don't get diluted by a larger volume of mid-reward fresh candidates.

\textbf{Invalid-output penalty.}
To keep optimization on the molecular manifold, candidates that fail validity checks, ie RDKit parsing \citep{rdkit} in the case of small molecules, receive a fixed penalty $r_{\mathrm{inv}}<0$ and enter training as low-reward examples. The mismatch this corrects is between the oracle's reward (defined for any string) and the actual objective (high reward on a valid molecule); the penalty closes that gap by giving the gradient an explicit signal to pull $p_{\theta_t}$ back toward the valid molecular manifold.

\section{Experimental Setup}
\label{sec:experimental-setup}\label{sec:exp-setup}

\begin{figure}[htbp]
  \centering
  \vspace{-0.1in}
  \includegraphics[width=\textwidth]{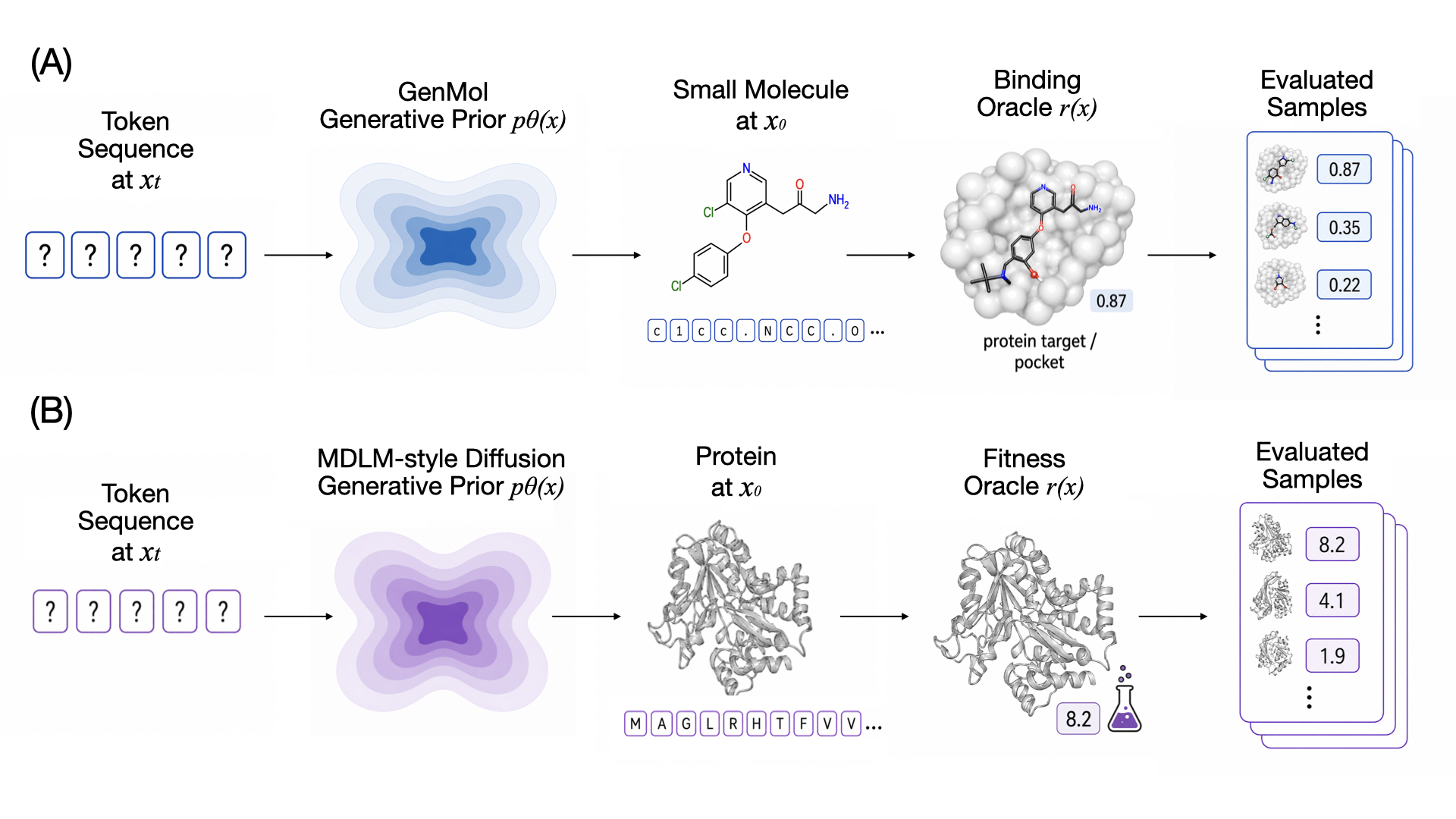}
  \vspace{-0.4in}
  \caption{In this study, we explore several tasks across two different domains, optimizing \textbf{(A)} small molecules for binding to protein targets and \textbf{(B)} proteins for various fitness metrics.}
  \vspace{-0.1in}
  \label{fig:diag_domain}
\end{figure}

\textbf{Tasks and oracles.}
We evaluate whether the design choices in Section~\ref{sec:design-space} improve discovery utility of finetuning methods under a fixed test-time budget. We study nine molecular optimization tasks across two domains. (Fig.~\ref{fig:diag_domain}) The first domain is small-molecule optimization for binding affinity to six protein targets (2VT4, 5SDV, 6CM4, 7BKC, 7C7M, 7YLL from \citep{zhangODesignWorldModel2025}) with GenMol \citep{leeGenMolDrugDiscovery2025} as the generator and FlashAffinity (FA) \citep{jiangFlashAffinityBridgingAccuracySpeed} as the primary oracle, with selected comparisons re-run under Boltz-2 \citep{passaro2025boltz2} to verify conclusions hold for an oracle two orders of magnitude more expensive and more accurate. While many studies approach this from a structure-based generative design perspective \citep{zhangODesignWorldModel2025, zengPropMolFlowPropertyguidedMolecule2026, schneuingStructurebasedDrugDesign2022,  guan3DEquivariantDiffusion2023, rekeshSynCoGenSynthesizable3D2025, adamsShEPhERDDiffusingShape2025}, we use discrete diffusion due to the increased validity, synthesizability, and diversity of generations offered by these models. The second domain is protein fitness optimization across three families (GB1, CreiLOV, TrpB) with a per-family MDLM-style discrete diffusion generator \citep{sahooSimpleEffectiveMasked2024} and a per-family fitness surrogate trained from experimental measurements as the oracle \citep{yangSteeringGenerativeModels2025}. For finetuning, we use either DDPP \citep{rector-brooksSteeringMaskedDiscrete2024} or VIDD \citep{suIterativeDistillationRewardGuided2025}. The complete environment is detailed in Appendix~\ref{app:repro-recipe}.

\textbf{Rewards.} In both domains the oracle \emph{defines} the reward the algorithm optimizes, and we treat it as ground truth in place of an expensive assay. For a molecule $x$ and target $p$, the reward is the molecular-weight-corrected, negated FA value head, $r(x) = -\mathrm{MWcorr}\bigl(\mathrm{FA}_{\text{val}}(x, p)\bigr)$, so that larger means stronger predicted binding; it is an unbounded affinity value rather than a normalized score, which is why absolute rewards differ across targets and we only ever compare within a target. For proteins the reward is the per-family fitness predictor: predicted IgG-Fc binding/enrichment fitness for GB1, predicted fluorescence fitness for CreiLOV, and predicted catalytic fitness for tryptophan formation from indole and serine for TrpB. Higher is better in all cases.

\textbf{Experimental protocol.} The high-level stages in Fig.~\ref{fig:test_time_feedback} define our \emph{full method} configuration for online adaptation, which serves as the upper-bound reference point for every ablation in Sec.~\ref{sec:results}. Each round generates $M$ candidates from the current $p_{\theta_t}$, selects the top-$K$ candidates ranked by Thompson-sampled surrogate scores for oracle evaluation, applies the invalid-SMILES penalty to candidates that fail RDKit before they reach the oracle, refreshes the CVaR threshold $\tau$ from the buffer, and runs $G$ fine-tuning gradient steps on stratified mini-batches with both reward-shape knobs (CVaR $+$ Density Entropy Regularization) applied to the log-reward. The ensemble surrogate is refit on the accumulated pairs at the end of each round. Pseudocode for the loop is given in Appendix~\ref{app:diagrams} as Algorithm~\ref{alg:method}.

\textbf{Evaluation protocol.}
Each method is evaluated under the same oracle-evaluation budget $B$ and random seeds. We report performance both as a function of cumulative oracle calls and as a function of GPU hours. Our primary metrics measure the upper tail of the evaluated set $\mathcal{D}_B$. We primarily report the best reward discovered so far, $\max_{x\in\mathcal{D}_B} r(x)$, as well as top-$k$ and top-$k\%$ means. To compare across tasks, we report the \emph{normalised reward}
\begin{equation}
\mathrm{NR}_t(x) \;=\; \frac{r_t(x)-C_t}{F_t-C_t},
\label{eq:lift}
\end{equation}
where $r_t(x)$ is the raw score of method $x$ on target $t$ at the evaluated budget, and $C_t$ and $F_t$ are target-specific reference endpoints computed over the methods in the corresponding comparison. Because raw FA rewards are unbounded and vary in scale across targets, this within-paper normalisation maps each target's reference interval to $[0,1]$, allowing aggregation across heterogeneous tasks. Since it is relative to the evaluated method set, it is intended for within-paper aggregation rather than cross-paper comparison. Alongside normalised reward, which evaluates vertical slices of the trajectory at a fixed budget $B$ to measure discovery quality, we report oracle-call speedup, which evaluates horizontal slices at a fixed target score (specifically, the strongest baseline's terminal score) to measure feedback efficiency. Because reward optimization can exploit oracle artifacts or leave the valid molecular manifold, we additionally track secondary metrics over a sliding window of generated candidates (parameters in Appendix~\ref{app:traces}): SMILES validity, pairwise Tanimoto diversity \citep{bajusz2015tanimoto}, QED \citep{bickerton2012qed}, and SA \citep{ertl2009sa}. We compare the full online loop against unconditional sampling, inference-time search variants, offline fine-tuning, and online fine-tuning without plug-in components (the untreated/vanilla
DDPP-LB or VIDD baseline).

\textbf{Advanced search.}
In addition to the core acquisition choice above, we also test whether the candidate-generation step benefits from inference-time search. Instead of plain ancestral sampling from the current generator, the loop can propose candidates using beam search \citep{ramesh2025noisetrajectory}, MCTS \citep{tangTR2D2TreeSearch2025}, or discrete Feynman-Kac correction \citep{hasanDiscreteFeynmanKacCorrectors2026, skretaFeynmanKacCorrectorsDiffusion2025} guided by the reward surrogate \citep{yangSteeringGenerativeModels2025}. In recent years, there have been significant related efforts combining inference time guidance with a fixed generator, which we do not fully explore here \citep{didi2026scaling, wangFineTuningDiscreteDiffusion2024, singhalGeneralFrameworkInferencetime2025, liDerivativeFreeGuidanceContinuous2024, nisonoffUnlockingGuidanceDiscrete2025, gruverProteinDesignGuided2023,  xiongProteinGuideOntheflyProperty2026, luProVADAGenerationSubcellular2025, ueharaRewardGuidedIterativeRefinement2025, liDynamicSearchInferenceTime2025, jianGeneralBindingAffinity2026, yangHitLeadDiscovery, tangPepTuneNovoGeneration2024, tangTR2D2TreeSearch2025, yangSteeringGenerativeModels2025}.

\newpage
\section{Results}\label{sec:results}

\begin{figure}[H]
  \centering
  
  \includegraphics[width=0.90\textwidth]{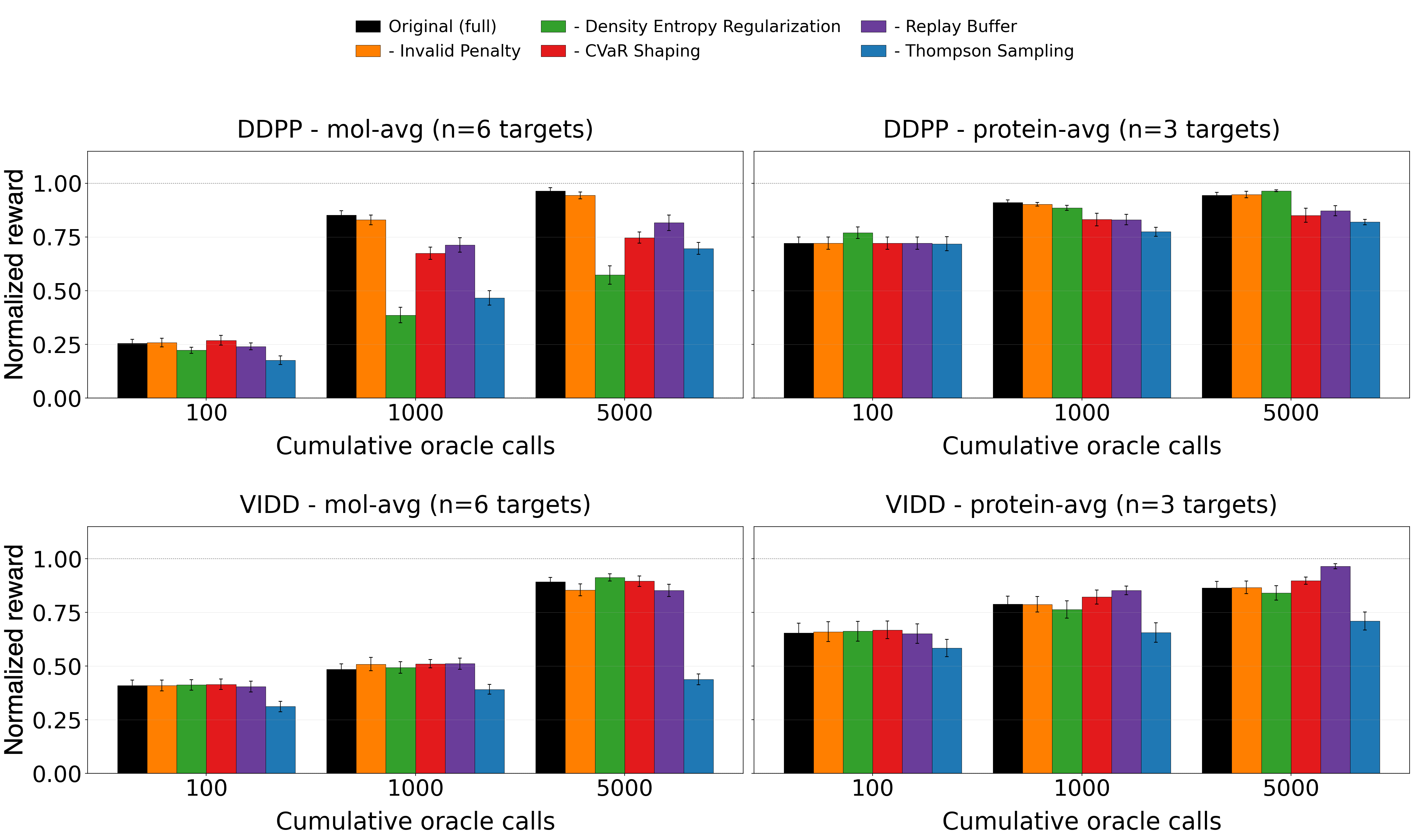}
   \caption{\textbf{Leave-one-out ablations reveal complementary roles most clearly on small-molecule tasks.} Normalised reward at 100/1k/5k calls for DDPP-LB (top) and VIDD (bottom), averaged over molecule targets (left) and protein families (right). Small-molecule effects are strongest for Density Entropy Regularization; protein effects are weaker, with Thompson sampling clearest. Error bars: $\pm1$ SEM over target--seed runs; Appendix~\ref{app:knobs} gives tables.}
\label{fig:barplot_ablation}
\end{figure}

Our main finding is that the online-loop components are complementary, although not equally so across domains (Fig.~\ref{fig:barplot_ablation}). On small molecules, acquisition, CVaR reward shaping, and Density Entropy Regularization produce the clearest reward gains, while replay and validity penalties stabilize the exploratory loop. On protein-fitness tasks, ablation effects are weaker and Thompson sampling gives the clearest consistent gain, suggesting that the family-specific priors already cover high-fitness regions more closely. Sec.~\ref{sec:results-method} breaks down the controlled ablations, and Sec.~\ref{sec:analysis} analyzes how these components shift the fine-tuned generator's distribution.

\subsection{Individual harness design components complement each other}\label{sec:results-method}

\textbf{Steering Design Choices.}
Thompson sampling gives the clearest acquisition-side gain in the main leave-one-out ablations (Fig.~\ref{fig:barplot_ablation}). In the isolated Thompson ablation, its strongest effect appears in the top-10\% mean rather than in top-1 alone, consistent with its role in diversifying which candidates receive oracle calls and shifting the generated distribution toward higher-reward modes (Appendix Fig.~\ref{fig:thompson_ablation}).
In contrast, adding inference-time search to the online loop (best-of-$N$, beam, MCTS, DFKC, or SMC) does not improve over plain online fine-tuning at fixed wall budget (Appendix Fig.~\ref{fig:tweedie_active}). The reason is primarily compute allocation: search proposes $L \times M$ candidates per round instead of $M$, and the extra forward passes needed to score and prune them dominate the wall budget, leaving little time for fine-tuning gradient steps (Appendix Figs.~\ref{fig:search_overhead}, \ref{fig:tweedie_active_compute}, \ref{fig:compute_breakdown}). Under this budget, each gradient step extracts more reusable signal from an oracle call than one-shot search does.

\textbf{Component Interactions.} The complete $2 \times 2 \times 2$ factorial in Table~\ref{tab:factorial} demonstrates that DER, CVaR, and Thompson acquisition exhibit compounding synergy rather than simple additivity. We interpret this pattern as a gating interaction: DER acts as a gate because Thompson and CVaR operate relative to the current generator's support $p_{\theta_t}$. Without DER, $p_{\theta_t}$ remains confined to high-density prior modes where surrogate disagreement is low and upper-quantile updates simply sharpen local modes. Conversely, DER alone disperses proposals broadly into low-likelihood regions without selective filtering. Once DER pushes $p_{\theta_t}$ into unfamiliar off-prior space, Thompson sampling is required to translate the resulting surrogate uncertainty into non-redundant oracle evaluations, while CVaR is required to suppress off-prior exploratory noise and concentrate gradient steps on the newly discovered upper tail.

\begin{table}[H]
\centering
\small
\begin{tabular}{@{}cccr@{}}
\toprule
DER & CVaR & Thompson & Score \\
\midrule
on  & on  & on  & $+1.13$ \\
on  & off & on  & $+0.54$ \\
on  & on  & off & $-0.20$ \\
on  & off & off & $-0.43$ \\
off & on  & on  & $-0.54$ \\
off & off & on  & $-0.80$ \\
off & on  & off & $-0.86$ \\
off & off & off & $-1.06$ \\
\bottomrule
\end{tabular}

\vspace{6pt}
\caption{\textbf{All eight combinations of DER, CVaR, and Thompson (DDPP-LB).} Mean best oracle score at $1{,}000$ oracle calls, averaged over $30$ target--seed pairs. Turned on alone from the all-off cell ($-1.06$), DER is worth $+0.63$, Thompson $+0.26$, and CVaR $+0.20$; their individual gains sum to $+1.09$, falling well short of the $+2.19$ joint improvement.}
\label{tab:factorial}
\end{table}
\vspace{-22pt}

\textbf{Objective-Mismatch Corrections.} 
Replay and invalid-output penalties have smaller and less uniform effects on top-1 reward (Fig.~\ref{fig:barplot_ablation}), but they stabilize failure modes induced by exploration. The replay buffer has its clearest effect on top-10\% mean (Appendix Fig.~\ref{fig:buffer_ablation}), consistent with aligning the gradient with the time-averaged distribution of evaluated data rather than only the current on-policy batch. The invalid-output penalty prevents a sharper failure: without a validity penalty ($r_{\text{inv}}{=}0$), validity collapses within the first hour as the model concentrates mass on out-of-distribution but high-surrogate-reward strings~\citep{wangFineTuningDiscreteDiffusion2024, ueharaELEGANT2024}. This collapse is driven primarily by the likelihood penalty in Density Entropy Regularization (Appendix~\ref{app:knobs}), so validity control is best viewed as a guardrail for aggressive exploration rather than as a standalone reward booster.
\begin{figure}[H]
  \centering
  \includegraphics[width=0.90\textwidth]{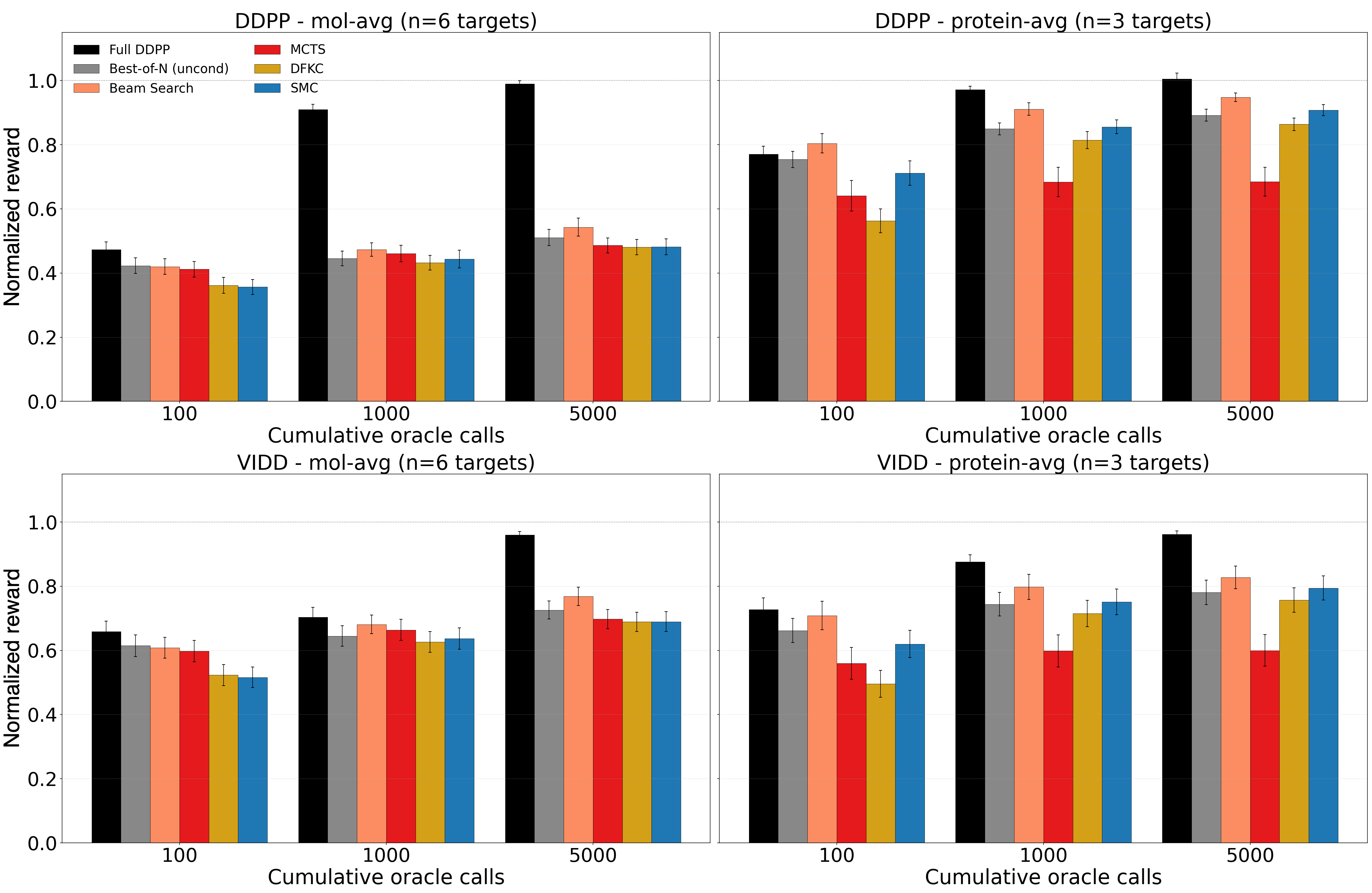}
  \caption{\textbf{Online adaptation outperforms inference-time search most clearly on small-molecule tasks.} Full loop (black) versus five search baselines at matched oracle-call budgets. Protein terminal gaps are smaller despite a $16\times$ call-efficiency gain. Error bars: $\pm1$ SEM over target--seed runs; normalised reward per Eq.~\ref{eq:lift}.}
  \label{fig:barplot_inference_search}
\end{figure}
\textbf{Online Adaptation Outperforms Inference-Time Search.} 
Fig.~\ref{fig:barplot_inference_search} compares the full online harness against inference-time search baselines that keep the generator fixed during selection (best-of-$N$, beam, MCTS, DFKC, and SMC). The full loop dominates these baselines on small-molecule tasks because oracle feedback is reused through model updates rather than spent only on one-shot candidate selection. Averaged across the six small-molecule targets, the harness reaches the best search baseline's final score in $\approx 58\times$ fewer oracle calls; on protein-fitness tasks the corresponding speedup is $\approx 16\times$. Replacing DDPP-LB with VIDD as $\mathcal{L}_{\text{ft}}$ gives similar trends (Appendix~\ref{app:vidd-ablation}). 
Mechanistically, the advantage of online adaptation stems from resolving both the distributional and computational bottlenecks. Distributionally, inference-time search relies on lookahead rollouts or noisy discrete Tweedie estimates (Appendix Fig.~\ref{fig:tweedie_search}) to filter trajectories from a frozen prior $p_{\text{pre}}$; when high-affinity ligands reside far outside the prior's dominant modes, reweighting static samples cannot traverse the density gap, whereas gradient updates progressively shift the proposal distribution $p_{\theta_t}$ toward the high-reward basin. Computationally, while pure search discards feedback after each selection step, online adaptation amortizes each oracle query into persistent parameter updates, maintaining its advantage across $100\times$ oracle-cost rescalings and full Boltz-2 evaluations (Appendix Fig.~\ref{fig:boltz_surrogate}, Fig.~\ref{fig:boltz_oracle_1x}, Fig.~\ref{fig:oracle_rescale}).


\subsection{Understanding performance improvements through distribution shifts in \texorpdfstring{$p_{\theta_t}$}{p\_theta\_t}}\label{sec:analysis}
\begin{figure}[H]
  \centering
  \includegraphics[width=0.90\textwidth]{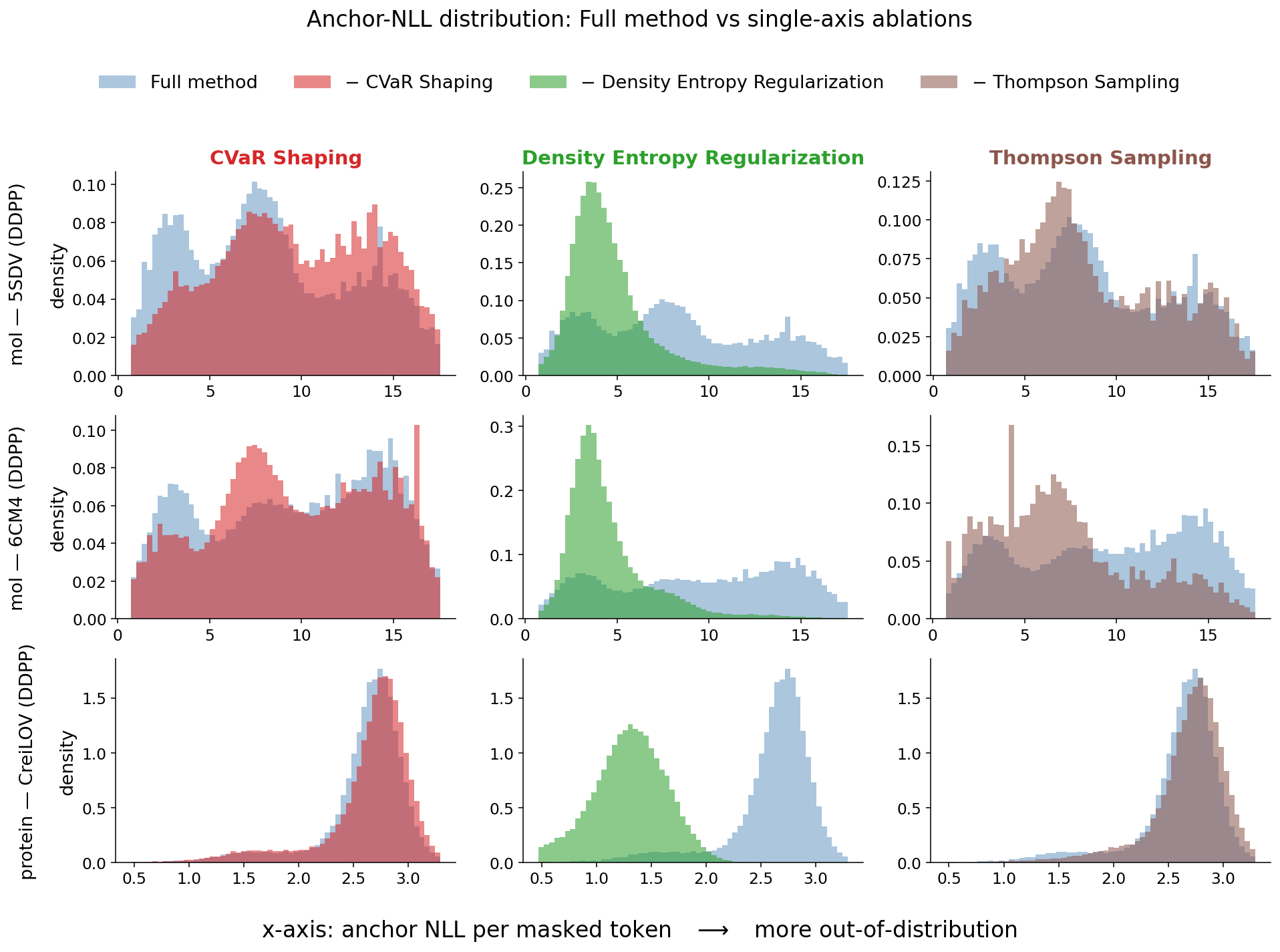}
  \caption{\textbf{Distribution shifts in $p_{\theta_t}$ explain performance gains.} Run-end anchor-NLL for the full method versus leave-one-out ablations. Density Entropy Regularization produces the strongest off-prior shift, while protein shifts are weaker. Additional 2VT4R results appear in Appendix Fig.~\ref{fig:anchor_nll}.}
  \label{fig:distribtuion_shift_main}
\end{figure}

The domain dependence in Fig.~\ref{fig:barplot_ablation} is largely explained by how far adaptation must move the generator away from its pretrained prior. Fig.~\ref{fig:distribtuion_shift_main} measures this shift using anchor NLL under the frozen pretrained model, $-\log p_{\text{pre}}(x)$. Density Entropy Regularization has the strongest and most direct effect on this distribution: removing it collapses mass back toward high-density regions of the pretrained prior. This matches its role in the objective, where the likelihood penalty discourages repeatedly reinforcing modes the current generator already occupies.

CVaR shaping and Thompson sampling affect distribution shift more conditionally. CVaR focuses the gradient on the upper reward tail, while Thompson sampling changes which uncertain candidates receive oracle calls. Either mechanism can move the generator farther from the prior when high-reward candidates lie in low-density regions, but neither has a fixed direction of shift. In Fig.~\ref{fig:distribtuion_shift_main}, the no-CVaR and full-method densities nearly overlap on protein-CreiLOV, where the prior already appears to cover the high-reward basin. On small-molecule targets such as 5SDV and 6CM4, removing CVaR or Thompson produces a visible change in the high-NLL tail, indicating that reward shaping and acquisition can help expose off-prior high-reward regions.

Once acquisition and reward shaping move $p_{\theta_t}$ into less prior-dense regions, the main failure mode shifts from insufficient exploration to target mismatch. A naive on-policy update trains only on the current batch, so earlier high-reward discoveries can stop influencing the generator. The replay buffer addresses this by aligning the gradient with the reward-tilted distribution over all evaluated data, not just the latest on-policy samples. A second mismatch appears when reward optimization moves probability mass off the valid molecular manifold. The invalid-SMILES penalty gives these failures a low-reward training signal, pulling $p_{\theta_t}$ back whenever exploration produces invalid strings. These components are therefore best understood as guardrails for the exploratory loop rather than as the main drivers of distribution shift.

This distributional view explains why the ablation effects are stronger on small-molecule tasks and weaker on protein-fitness tasks. The small-molecule generator is a broader prior, so high-affinity candidates often require larger shifts away from its dominant modes. The protein generators are family-specific, so high-fitness variants appear closer to regions already covered by the pretrained prior. As a result, reward shaping and debiasing are less decisive on proteins, while acquisition can still help by allocating oracle calls more effectively. This also explains why search baselines and ablations converge more quickly on the protein tasks: when the prior already places mass near high-reward modes, adapting the generator and selecting from it become closer in effect.

\section{Conclusion}
This work studies how to use limited oracle feedback for molecular optimization with pretrained discrete diffusion models. Our main finding is that online adaptation benefits from combining components that act on different parts of the feedback loop. Acquisition, CVaR reward shaping, and Density Entropy Regularization provide the clearest reward gains on small-molecule tasks, while replay and invalid-output penalties stabilize the exploratory loop. Together, these components form a practical recipe that outperforms offline fine-tuning and inference-time search baselines under matched oracle-call budgets and GPU-hour accounting.

The results also clarify when this recipe is most useful. Small-molecule binding optimization often requires moving beyond high-density regions of a broad pretrained prior, so shaping and debiasing matter more. Protein-fitness tasks show weaker ablation effects, consistent with family-specific priors already covering high-fitness regions more closely; in that regime, acquisition remains useful because oracle-call allocation still matters, but large distribution shifts are less central. Future work should extend this recipe to settings with less reliable learned oracles, multi-objective design goals, and other discrete-generation domains with non-differentiable feedback. 

\textbf{Acknowledgments.} This work was primarily conducted by researchers at Caltech, who were supported in part by NSF \#2505096, and gifts from Genesis Therapeutics, Point72, and OpenAI. Jason Yang was supported in part by a Google Fellowship. Riccardo De Santi was supported by and ETH AI Center doctoral fellowship, the Swiss
National Science Foundation under NCCR Catalysis (grants 180544 \& 225147) and NCCR Automation (grant 51NF40 180545), and Centres of Competence in Research funded by the Swiss National Science Foundation.

\newpage

\bibliographystyle{unsrtnat}
\bibliography{main}


\newpage

\appendix

\section*{Guide to the Appendix}
The appendix follows the logic of the paper: definitions and reproducibility first, then method illustrations, controlled component evidence, search comparisons, task-level results, and finally the alternative-oracle and qualitative checks.
\begin{itemize}
\item \textbf{Experimental definitions and reproducibility (Appendix~\ref{app:method-details}).} Normalization, budget accounting, backbones and oracles, implementation details, hyperparameters, the reproduction recipe, and stochasticity.
\item \textbf{Method and baseline illustrations (Appendix~\ref{app:diagrams}).} The active loop, fine-tuning and exploration components, inference-time search families, and computational shortcuts.
\item \textbf{Component ablations and diagnostics (Appendix~\ref{app:component-analysis}).} Individual knobs, secondary metrics, distribution-shift diagnostics, the joint factorial, the untreated baseline, and cross-finetuner results.
\item \textbf{Search baselines and compute accounting (Appendix~\ref{app:search}).} Online versus offline adaptation, frozen-generator search, search inside the online loop, and oracle-cost sensitivity.
\item \textbf{Task-level and aggregate results (Appendix~\ref{app:results-detail}).} Per-task curves, domain aggregates, GPU-hour views, and pointers to the canonical main-text bar plots.
\item \textbf{Alternative oracle and qualitative examples (Appendix~\ref{app:external-validation}).} Boltz-2 experiments and representative valid molecules.
\end{itemize}

\clearpage

\section{Experimental Definitions and Reproducibility}\label{app:method-details}

This appendix records implementation specifics that are not stated in the main body. Hyperparameter values are collected in Table~\ref{tab:hparams}.

\subsection{Normalised Reward}\label{app:lift-fraction}

The normalised reward is defined in Eq.~\ref{eq:lift} in Sec.~\ref{sec:exp-setup}. The target-specific endpoints $C_t$ and $F_t$ are computed over the figure-specific method set: component ablations together with \texttt{best\_of\_n} for Fig.~\ref{fig:barplot_ablation}, and the full configuration together with search baselines for Fig.~\ref{fig:barplot_inference_search}. Per-bar means and symmetric SEMs are pooled over target--seed draws.

%
%
%
\subsection{Wall-time accounting}

Wall time for a single run decomposes as
\begin{equation}
t_{\text{wall}} \;=\; N \cdot c_{\text{oracle}} + C_{\text{method}},
\label{eq:wall-decomp}
\end{equation}
where $N$ is the number of oracle calls, $c_{\text{oracle}}$ is the per-call oracle cost, and $C_{\text{method}}$ collects all method-side compute (generation, fine-tuning steps, surrogate updates). Reporting against both axes (oracle calls and GPU hours) reveals whether one technique dominates another regardless of oracle cost, or only past a threshold $c_{\text{oracle}}^\star$ at which the two curves cross. The retroactive $100\times$ oracle-cost rescaling used throughout the appendix figures simulates a switch from a fast oracle (FA) to a slow oracle (Boltz-2-class) by scaling $c_{\text{oracle}}$ in this decomposition without re-running the experiments.

\subsection{Backbone and oracle internals}

The GenMol checkpoint we use is the released MDLM-style masked discrete diffusion model over SAFE-tokenized SMILES \citep{noutahiGottaBeSAFE2023, sahoo2024mdlm} with vocabulary size 1882 (1880 SAFE tokens plus mask and end-of-sequence). The forward noise schedule and architecture are unchanged from the release; only the policy weights $\theta$ are updated. A FlashAffinity call decomposes as: parse the SMILES with RDKit; embed a 3D conformer with ETKDGv3 (one retry on failure); extract per-atom features with the FA torchdrug parser; score the (pocket, ligand) pair with the released FA checkpoint. Per-pocket ESM3 protein features are pre-computed once and cached. Each FA call takes tens of milliseconds on the A100; Boltz-2 calls take roughly $20$ seconds on the same hardware, two orders of magnitude slower.

\subsection{Run logs}

\begin{figure}[H]
  \centering
    \includegraphics[width=0.85\textwidth]{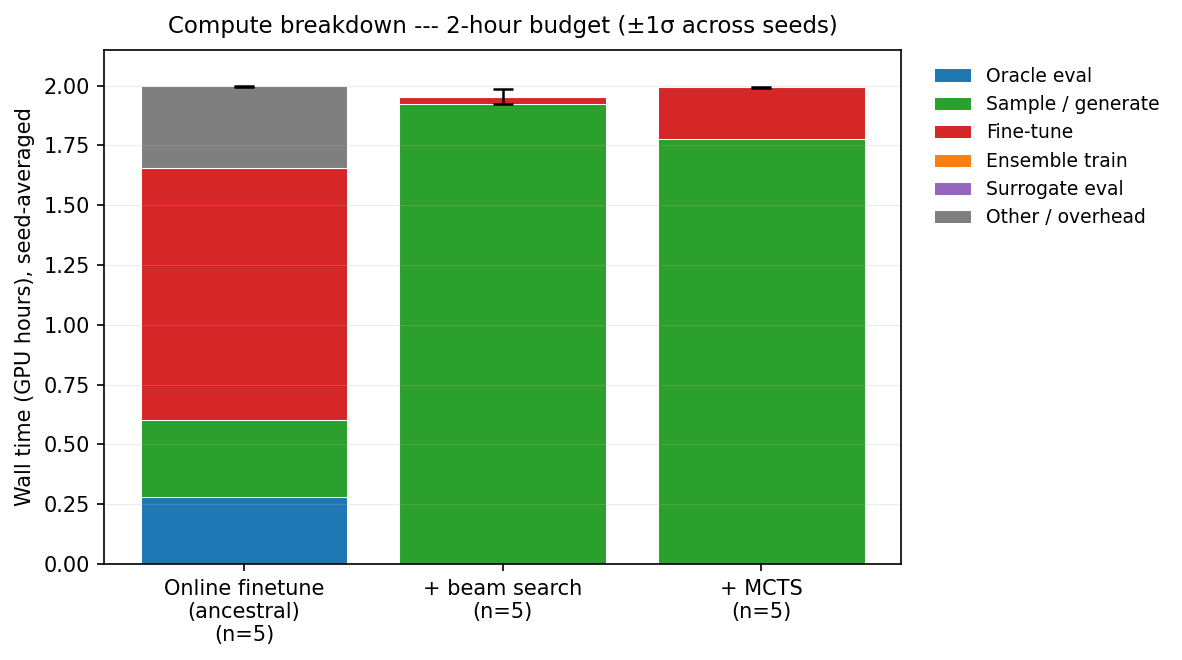}
  \caption{Compute breakdown under a 2-hour wall budget for plain online finetuning (left) versus online finetuning augmented with beam search (middle) or MCTS (right). All numbers are seed-averaged ($n{=}5$). Plain online finetune spends roughly half its budget on fine-tuning ($\sim$1.0\,hr); both search variants spend almost the entire budget on candidate generation, leaving nearly no time for fine-tuning gradient steps ($\sim$0.04\,hr for beam, $\sim$0.21\,hr for MCTS). The reduction in fine-tuning wall time is the proximate reason search-augmented online finetuning fails to improve over the plain variant.}
  \label{fig:search_overhead}
\end{figure}

The compute-breakdown plots (Appendix Figs.~\ref{fig:search_overhead} and~\ref{fig:compute_breakdown}) and the retroactive $100\times$ oracle-cost rescaling are computed offline from two log files written during every run. \texttt{oracle\_timeline.jsonl} has one line per oracle call (cumulative call count, SMILES, score, best-so-far, elapsed seconds, RDKit-derived QED and SA, epoch index). \texttt{active\_loop\_log.jsonl} has one line per active-loop round (round number, top-1 best-so-far $f^\star$, mean ensemble disagreement $\sigma$, Spearman $\rho$ between ensemble and oracle on a held-out $20\%$ of the buffer, cumulative wall seconds, and per-phase wall times $t_{\text{gen}}, t_{\text{surrogate}}, t_{\text{oracle}}, t_{\text{ddpp}}, t_{\text{ensemble}}$).

\subsection{Implementation subtleties}

Three details are not derivable from Algorithm~\ref{alg:method} or Table~\ref{tab:hparams} and matter for reproducing our numbers.

\paragraph{Gradient detach in model debiasing.} $\log p_\theta(x)$ in Eq.~\ref{eq:debias} is treated as a constant: gradient is not propagated through the debiasing term. Passing gradient through it turns the loss into an entropy-regularized policy gradient and breaks the posterior-matching guarantee of DDPP-LB.

\paragraph{Thompson noise scope.} At acquisition we draw one $\varepsilon \sim \mathcal{N}(0, 1)$ per candidate, shared across all $J$ ensemble heads, not one per (candidate, head) pair. The per-candidate scalar acts on the ensemble disagreement $\sigma_i$ and is what produces the diversity push described in Sec.~\ref{sec:results}.

\paragraph{CVaR threshold refresh.} The threshold $\tau$ is recomputed once per active-loop epoch from the empirical $q$-quantile of buffer rewards, not per gradient step. Below-threshold samples remain in the gradient with their shaped reward floored at $\epsilon = 10^{-8}$, which under the DDPP log-reward becomes a strong negative signal that pushes $q_\theta$ away from them rather than dropping them entirely.

\subsection{Hyperparameters}

Table~\ref{tab:hparams} lists the hyperparameter values used throughout the controlled experiments and the final-comparison runs. Defaults follow the recommended settings from the original DDPP-LB and VIDD references where applicable; values for the active-loop wrappers ($M, K, G$, ensemble size and training schedule) follow our active-loop reference implementation. We did not run an exhaustive per-method hyperparameter sweep. The design space we ablate has thirteen-plus methods (Table~\ref{tab:methods}) crossed with five seeds and a two-hour FA budget per cell, and the cost grows by another order of magnitude under the Boltz-2 oracle; tuning each cell independently would multiply that cost by the size of the per-method search grid and would also confound the design-space comparison, since improvements would no longer be attributable to the knob under study versus to a more aggressive sweep on one cell than another. We instead control everything except the knob being ablated, hold all other hyperparameters at the active-loop reference defaults, and run every cell at the same wall budget. The trade-off is explicit: we measure the effect of each design choice at sensible-default settings rather than at each method's peak-tuned setting. Per-knob sensitivity to the hyperparameter most likely to matter (the invalid-SMILES penalty magnitude) is reported separately in Appendix Fig.~\ref{fig:penalty_sweep}.

\begin{table}[H]
\centering
\small
\begin{tabular}{@{}l l l@{}}
\toprule
\textbf{Group} & \textbf{Hyperparameter} & \textbf{Value} \\
\midrule
\multicolumn{3}{@{}l}{\textit{Wall budget and seeds}} \\
& Seeds                         & $\{0, 1, 2, 3, 42\}$ \\
& Wall budget per seed (FA)     & 2 hours \\
& Wall budget per seed (Boltz-2) & 10 hours \\
& Hardware                      & 1$\times$ NVIDIA A100 80GB \\
\midrule
\multicolumn{3}{@{}l}{\textit{Active loop}} \\
& Generation batch $M$          & 1000 \\
& Oracle batch $K$              & 25 \\
& Fine-tuning steps per round $G$ & 50 \\
& Buffer capacity               & 10\,000 \\
& Buffer eviction               & priority (lowest reward first) \\
& Fresh-fraction                & 0.25 \\
\midrule
\multicolumn{3}{@{}l}{\textit{Reward shaping}} \\
& CVaR quantile $q$             & 0.8 \\
& Reward floor                  & $10^{-8}$ \\
& Model-debiasing weight $\gamma$ & 1.0 \\
& Invalid-SMILES penalty $r_{\text{inv}}$ & $-5$ \\
\midrule
\multicolumn{3}{@{}l}{\textit{Surrogate ensemble (Thompson)}} \\
& Ensemble size $J$             & 10 \\
& Hidden width                  & 256 \\
& Training epochs / round  & 100 \\
& Learning rate                 & $10^{-3}$ \\
& Mini-batch size               & 256 \\
\midrule
\multicolumn{3}{@{}l}{\textit{Fine-tuning (DDPP-LB)}} \\
& Backbone learning rate        & $5 \times 10^{-5}$ \\
& LogZ head learning rate       & $5 \times 10^{-4}$ \\
& EMA decay                     & 0.999 \\
& Backbone warmup (LogZ-only steps) & 200 \\
\midrule
\multicolumn{3}{@{}l}{\textit{Search baselines (when used as a knob)}} \\
& Beam: $N$ beams, $L$ branches, $K$ steps & $N{=}8$, $L{=}4$, $K{=}5$ \\
& MCTS: $c_{\text{UCT}}$, $L$, $K$         & $1.0$, $4$, $5$ \\
& DFKC: particles, $\beta$, score\_start   & 8, 2.0, 0.5 \\
\bottomrule
\end{tabular}
\caption{\textbf{Hyperparameters used across all reported experiments.} Values are the active-loop reference defaults; only the knob being ablated is varied within a comparison.}
\label{tab:hparams}
\end{table}

\begin{table}[H]
\centering
\small
\begin{tabular}{@{}llrr@{}}
\toprule
\textbf{Hyperparameter} & \textbf{Values tested} & \textbf{Best score} & \textbf{Top-100} \\
\midrule
Ensemble size $J$ & 3, 5, \textbf{10} & $0.20$ & $0.07$ \\
Invalid penalty $r_{\text{inv}}$ & 0, $-1$, $-3$, $\mathbf{-5}$, $-7$ & $0.36$ & $0.38$ \\
Buffer capacity & 100, 1000, \textbf{unlimited} & $0.13$ & $0.21$ \\
CVaR quantile $q$ & 0.60, 0.70, \textbf{0.80}, 0.90 & $0.14$ & $0.16$ \\
Oracle batch $K$ & 10, \textbf{25}, 50 & $0.25$ & $0.32$ \\
Debias weight $\gamma$ & 0.25, 0.5, \textbf{1.0}, 2.0 & $1.02$ & $1.01$ \\
\bottomrule
\end{tabular}
\caption{\textbf{Hyperparameter sensitivity.} Largest absolute change relative to the default (bold) in best score and top-$100$ mean, averaged over $3$ targets with $3$ runs per value per target. The sweep's default cell runs an unbounded buffer; this matches the $10{,}000$ capacity of Table~\ref{tab:hparams} in effect, because no run evaluates more than $8{,}750$ candidates, so the cap never binds and never evicts. The mean within-cell standard deviation is $0.22$, so the five small effects are of the same order as run-to-run variation. The debiasing weight is the exception, and it sets a threshold rather than requiring tuning: the score rises from $\gamma{=}0.25$ to $\gamma{=}1.0$ and does not rise further at $\gamma{=}2.0$, so any $\gamma \ge 1$ is equivalent --- consistent with $\gamma{=}1$ putting $\log r$ and $\log p_\theta$ on the same scale. $K$ is set by the ratio $K/M$ of oracle batch to generation batch rather than by tuning: performance is flat from $K{=}10$ to $K{=}25$ and drops at $K{=}50$, where the surrogate is barely filtering the pool.}
\label{tab:hpsens}
\end{table}

These sweeps do consume oracle calls. They were run on $3$ targets at $3$ seeds and were \emph{not} used to retune per-target settings: the configuration reported in Table~\ref{tab:hparams} is held fixed across all six molecular targets, all three protein families, and all seeds, so the reported results do not benefit from per-target tuning.

\subsection{End-to-end reproduction recipe}\label{app:repro-recipe}

This subsection lists, in order, every concrete asset and step required to reproduce one cell of our experimental table; combined with Algorithm~\ref{alg:method} (the abstract loop) and Table~\ref{tab:hparams} (the numerical settings), this is sufficient to rebuild the environment and re-run.

\paragraph{Software environment.} Python 3.10. PyTorch 2.x with CUDA 12.x. RDKit 2024.x (used for SAFE encoding/decoding, validity checking via \texttt{MolFromSmiles}, and ETKDGv3 conformer embedding). NumPy 1.x. Hydra for configuration. \textbf{boltz} package version 2.2.1 (PyPI) for the Boltz-2 oracle path.

\paragraph{Pretrained assets (downloaded once, cached locally).} (i) The \textbf{GenMol} masked discrete-diffusion checkpoint from \citet{leeGenMolDrugDiscovery2025} (ICML 2025; arXiv:2501.06158), released by the authors at \url{https://github.com/NVIDIA-BioNeMo/genmol}; we use the second-vocab-fix release shipped as \texttt{model\_v2.ckpt} (1397\,MB, vocabulary $1882$ = $1880$ SAFE tokens plus mask and end-of-sequence). Forward noise schedule and architecture are unchanged from the release; only the policy weights $\theta$ are updated. The model was pretrained on the \textbf{SAFE dataset V2} (\url{https://huggingface.co/datasets/datamol-io/safe-drugs}); we do not retrain. (ii) The \textbf{FlashAffinity} (FA) affinity-regression checkpoint
\texttt{FlashAffinity/checkpoints/value\_2.ckpt}
from \citet{jiangFlashAffinityBridgingAccuracySpeed},
together with its torchdrug-style featurizer; we use the regression
(value) head, apply the released molecular-weight correction,
and negate the corrected prediction. (iii) For Boltz-2 oracle runs, the released Boltz-2 weights downloaded via the \texttt{boltz} package's bootstrap (structure checkpoint \texttt{boltz2\_conf.ckpt} + affinity head \texttt{boltz2\_aff.ckpt}), \citet{passaro2025boltz2}. (iv) Per-pocket \textbf{ESM3} protein-feature tensors, computed once per PDB target and cached to disk (\texttt{FlashAffinity/data/protein\_test/repr/esm3.lmdb}) so that no oracle call recomputes them.

\paragraph{Per-target inputs.} Each PDB target (2VT4, 5SDV, 6CM4, 7BKC, 7C7M, 7YLL) contributes the receptor structure plus the binding-pocket specification used by the FA / Boltz-2 oracle. We use the publicly released coordinates without modification.

\paragraph{Seed handling.} A run is identified by \texttt{(method, target, seed)} with seed $\in \{0, 1, 2, 3, 42\}$. At process start the seed is propagated to: \texttt{torch.manual\_seed}, \texttt{torch.cuda.manual\_seed\_all}, \texttt{numpy.random.seed}, Python's built-in \texttt{random.seed}, and the RDKit conformer-embedding seed passed to \texttt{ETKDGv3}.

\paragraph{Per-cell launcher.} Each cell is dispatched by a single launcher that takes \texttt{(method, target, seed, wall\_budget\_sec)} as arguments. The launcher loads the GenMol checkpoint, attaches the configured component slots from Sec.~\ref{sec:design-space} (acquisition rule, reward-shape modifier, replay strategy, validity penalty, finetuning objective), opens the two log files described in App.~\ref{app:method-details} (\texttt{oracle\_timeline.jsonl} and \texttt{active\_loop\_log.jsonl}), and runs the loop in Algorithm~\ref{alg:method}.

\paragraph{Per-round step counts.} Each active-loop epoch generates $M{=}1000$ candidate molecules from the current $q_\theta$, scores them with the surrogate ensemble, picks the top-$K{=}25$ for oracle evaluation under Thompson sampling, evaluates them, pushes each $(x, r)$ pair into the buffer (with $r{=}r_{\text{inv}}$ for RDKit-failed candidates), refreshes the CVaR threshold $\tau$ from the buffer's current $q$-quantile, and runs $G{=}50$ DDPP-LB gradient steps on stratified mini-batches drawn from the buffer with fresh-fraction $0.25$. The ensemble surrogate is refit at the end of each epoch on all accumulated $(x, r)$ pairs for $100$ inner training epochs at learning rate $10^{-3}$ and mini-batch size $256$.

\paragraph{Wall-budget enforcement.} A monotonic clock is checked after each round. If the elapsed wall time exceeds the configured \texttt{wall\_budget\_sec} (2~h for FA, 10~h for Boltz-2; a $4$-h cap was applied to the small-molecule batch1 runs documented in our latest data drop), the loop terminates after writing one final \texttt{active\_loop\_log.jsonl} entry and a \texttt{results.json} summary. If the runner is killed externally, only the per-epoch and per-call log files are guaranteed to be flushed; the partial-summary case is described in App.~\ref{app:method-details}.

\paragraph{Output artifacts.} Every cell produces a directory \texttt{<TARGET>/<METHOD>\_s<SEED>/} containing exactly three files: \texttt{results.json} (post-run summary or partial stub), \texttt{active\_loop\_log.jsonl} (one JSON line per epoch), and \texttt{oracle\_timeline.jsonl} (one JSON line per oracle call). All paper figures are produced from these three file types alone; no additional artifacts are needed for plot regeneration.

\subsection{Sources of stochasticity beyond the seed}\label{app:stochasticity}

Setting the seed described above pins all explicitly-RNG-driven sampling decisions: which $M$ candidates the diffusion sampler draws each epoch, the Thompson noise $\varepsilon$ at acquisition, mini-batch composition during fine-tuning, ensemble weight initialization, and the conformer seed handed to RDKit. Several other sources of variation are \emph{not} pinned by the seed, and we list them here so that bit-level reproducibility is not expected even between two runs with identical \texttt{(method, target, seed)}.

\paragraph{GPU floating-point nondeterminism.} cuBLAS / cuDNN convolution and matmul algorithms make algorithm-selection decisions based on workspace memory and tensor shapes, and several reduction kernels (e.g., \texttt{scatter\_add\_}) use atomic adds whose summation order is not deterministic across kernel launches. We do not enable \texttt{torch.use\_deterministic\_algorithms(True)} because doing so would substitute slower kernels and change the wall-time profile that our compute-breakdown plots are measured against. The practical effect is small bit-level drift in $q_\theta$'s logits across nominally-identical runs, which can cascade into different sampled $x$ at later epochs.

\paragraph{ETKDGv3 conformer embedding.} The RDKit conformer-embedding routine has its own random restart policy when the first attempt fails to converge. We pass a fixed embedding seed (derived from the run seed), but the number of restarts and the resulting conformer can still differ across re-evaluations of the same SMILES if the underlying force-field convergence is borderline. Most molecules are unaffected; a small fraction (typically $<1\%$) yield slightly different FA scores on re-evaluation.

\paragraph{Async oracle ordering.} Oracle calls are issued in batches of $K$ but completed asynchronously when the oracle is GPU-bound (Boltz-2 in particular). The order in which $(x, r)$ pairs land in the buffer therefore depends on per-call latency, not just on the order they were dispatched. This affects: (i) the buffer's priority-eviction tie-breaks (lowest reward first; ties broken by arrival order), (ii) the empirical $q$-quantile that the CVaR threshold refresh uses (since the threshold is computed from buffer state at refresh time), and (iii) which candidates appear in which mini-batch under stratified sampling.

\paragraph{Wall-clock-driven decisions.} The active loop is wall-budget-bounded, not round-count-bounded. Two runs with the same seed on different hardware (or on the same hardware under different system load) will execute a different number of epochs before the budget cutoff fires, and therefore terminate at different points along the $q_\theta$ trajectory. The 4-hour cap on the small-molecule batch1 runs makes this effect visible directly in the per-target completion table reported in App.~\ref{app:knobs}.

\paragraph{DataLoader worker ordering.} When the buffer's mini-batch sampler runs with PyTorch \texttt{DataLoader} workers $> 0$, the order in which prefetched batches arrive at the training loop depends on worker-process scheduling. We use a single-worker loader for the fine-tuning step to remove this source, but the surrogate-ensemble retraining loop uses multiple workers for throughput; ensemble outputs are therefore not bit-exact across runs.

\paragraph{Practical implication.} For all reported metrics in the paper (top-1, top-10\%, validity, etc.) we average across five seeds and report $95\%$ CI bands. The bit-level drift above is dominated by the seed-driven variation in every comparison we have inspected; the conclusions in Sec.~\ref{sec:results} (which knobs are complementary, where the failure modes fire) are stable under the level of nondeterminism listed here. We surface these sources nonetheless because anyone trying to bit-match a single curve from our logs will not be able to, even with the seed and full hyperparameter set in hand.

\clearpage

\section{Method and Baseline Illustrations}\label{app:diagrams}

This appendix collects schematic diagrams for the methods and concepts introduced in Sec.~\ref{sec:design-space} and Sec.~\ref{sec:results}. Each diagram is referenced from the body where the corresponding concept is first introduced; this section serves as a single place to look up the visual.

\subsection{Active loop}

\begin{figure}[H]
  \centering
  \includegraphics[width=0.55\textwidth]{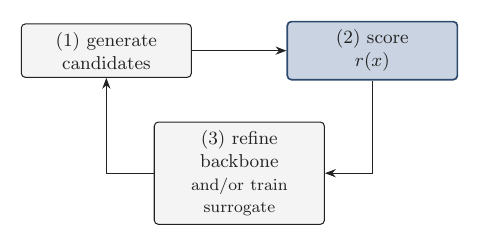}
  \caption{\textbf{Active loop.} The high-level structure shared by every fine-tuning method we evaluate: generate candidates, score them under the oracle, then refine the backbone $p_\theta$ and/or the surrogate before generating again.}
  \label{fig:diag_active}
\end{figure}

{\begin{algorithm}[H]
\caption{Online active loop with the five composable knobs (finetuner-agnostic). Schematic in Fig.~\ref{fig:test_time_feedback}.}
\label{alg:method}
\begin{algorithmic}[1]
\Require Pretrained $p_{\text{pre}}$, oracle $r(\cdot)$, fine-tuning loss $\mathcal{L}_{\text{ft}}$, wall budget $T_{\text{wall}}$
\Require Generation batch $M$, oracle batch $K$, fine-tuning steps per round $G$
\Require Reward-shape: CVaR quantile $q$, debiasing $\gamma$, invalid penalty $r_{\text{inv}}$; ensemble size $J$
\State Initialize $p_{\theta_0} \leftarrow p_{\text{pre}}$;~ buffer $\mathcal{B} \leftarrow \emptyset$;~ ensemble $\{\hat{r}_{\phi_j}\}_{j=1}^{J}$ random
\While{wall time $< T_{\text{wall}}$}
    \State $C_t \leftarrow$ sample $M$ candidates from $p_{\theta_t}$
    \State $(\mu_i, \sigma_i) \leftarrow$ ensemble mean/std on $x_i \in C_t$
    \State $\hat r_i \sim \mathcal{N}(\mu_i, \sigma_i^2)$ \Comment{\emph{(4) Thompson sampling}}
    \State $S_t \leftarrow$ top-$K$ of $C_t$ by $\hat r_i$
    \For{$x \in S_t$}
        \State $r_x \leftarrow r_{\text{inv}}$ if $x$ fails RDKit parse \textbf{else} $r(x)$ \Comment{\emph{(3) invalid-SMILES penalty}}
        \State $\mathcal{B} \leftarrow \mathcal{B} \cup \{(x, r_x)\}$ \Comment{\emph{(5) replay buffer}}
    \EndFor
    \State $\tau \leftarrow Q_q(\{r : (\cdot, r) \in \mathcal{B}\})$
    \For{$g = 1, \ldots, G$}
        \State sample mini-batch $\mathcal{B}_b \subset \mathcal{B}$
        \State $r^*(x) \leftarrow (r_x - \tau)_+ / (1-q)$ \Comment{\emph{(1) CVaR shaping}}
        \State $\log \tilde{r}(x) \leftarrow \log r^*(x) - \gamma \log p_{\theta_t}(x)$ \Comment{\emph{(2) model debiasing}}
        \State update $p_{\theta_t}$ via $\mathcal{L}_{\text{ft}}$ on $\mathcal{B}_b$ using $\log \tilde{r}$ as the per-sample reward
    \EndFor
    \State refit ensemble $\{\hat{r}_{\phi_j}\}$ on $\mathcal{B}$
\EndWhile
\State \Return $\arg\max_{x \in \mathcal{B}} r_x$
\end{algorithmic}
\end{algorithm}}

\subsection{Fine-tuning objective and exploration knobs}

\begin{figure}[H]
  \centering
  \includegraphics[width=0.95\textwidth]{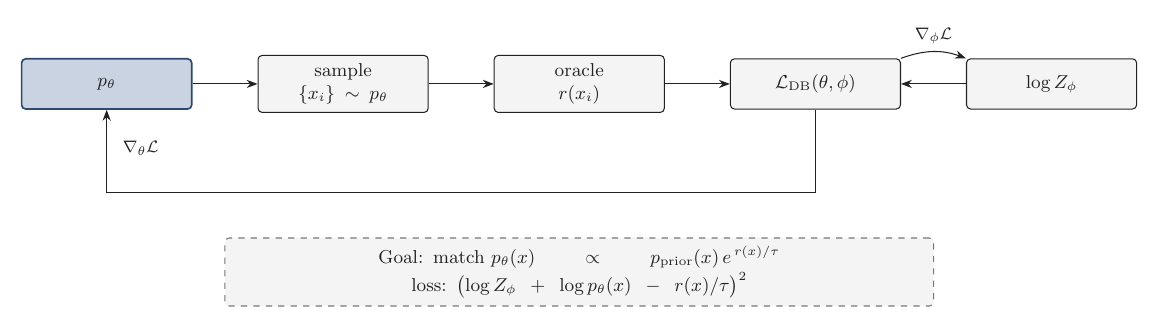}
  \caption{\textbf{DDPP-LB loss.} Trajectory-balance squared residual matching $q_\theta$ to the reward-tilted posterior $p^\star \propto p_{\text{pre}} \cdot r$, with a learned partition-function head $\hat{Z}_\phi$ trained jointly with $\theta$.}
  \label{fig:diag_ddpp}
\end{figure}

\begin{figure}[H]
\centering
\includegraphics[width=0.85\textwidth]{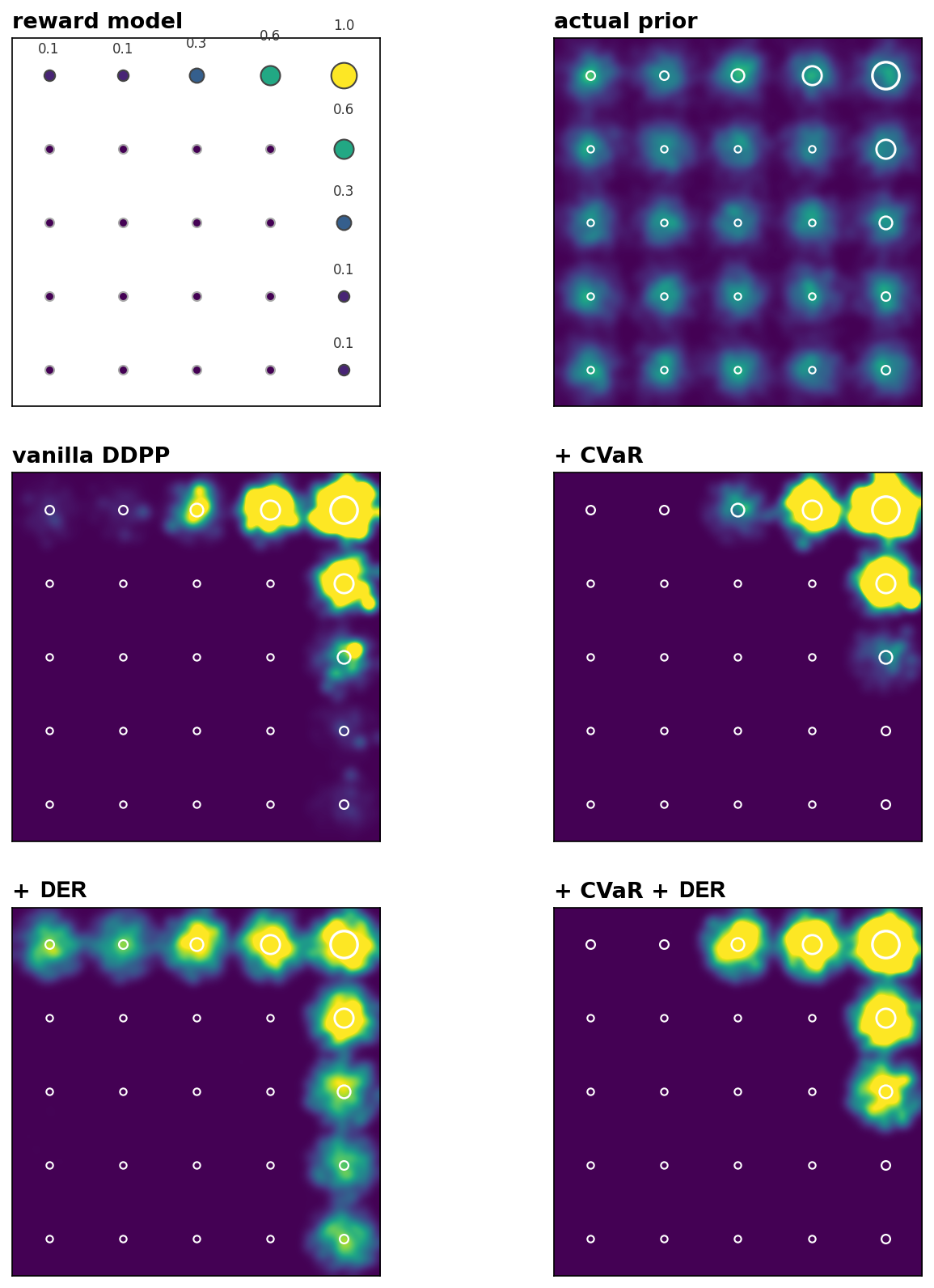}
\caption{\textbf{Reward shapes on a synthetic 25-mode landscape.} (a) Prior $p_{\text{anchor}}$: 25 roughly Gaussian modes. (b) Each mode is assigned a score in $[0,1]$. (c) Vanilla DDPP-LB with raw reward $r(x)$ puts mass on modes proportional to their reward weight (ideal under unlimited sampling, but compute-inefficient under limited oracle calls). (d) CVaR shaping concentrates most mass on the $r \geq 0.3$ modes, enabling deeper search of the top-reward modes. (e) Model debiasing $\log \tilde r(x)=\log \tilde r_{\mathrm{cvar}}(x) - \gamma \log p_{\theta_t}(x)$ flattens probability mass across rewarded modes, allowing rare high-scoring molecules in less-likely modes to be discovered. (f) CVaR $+$ model debiasing yields a roughly uniform distribution over the top-$\tau$ modes, preventing CVaR's early collapse while preserving its tail focus.}
\label{fig:synth_dist_evolution}
\end{figure}

\begin{figure}[H]
  \centering
  \includegraphics[width=0.65\textwidth]{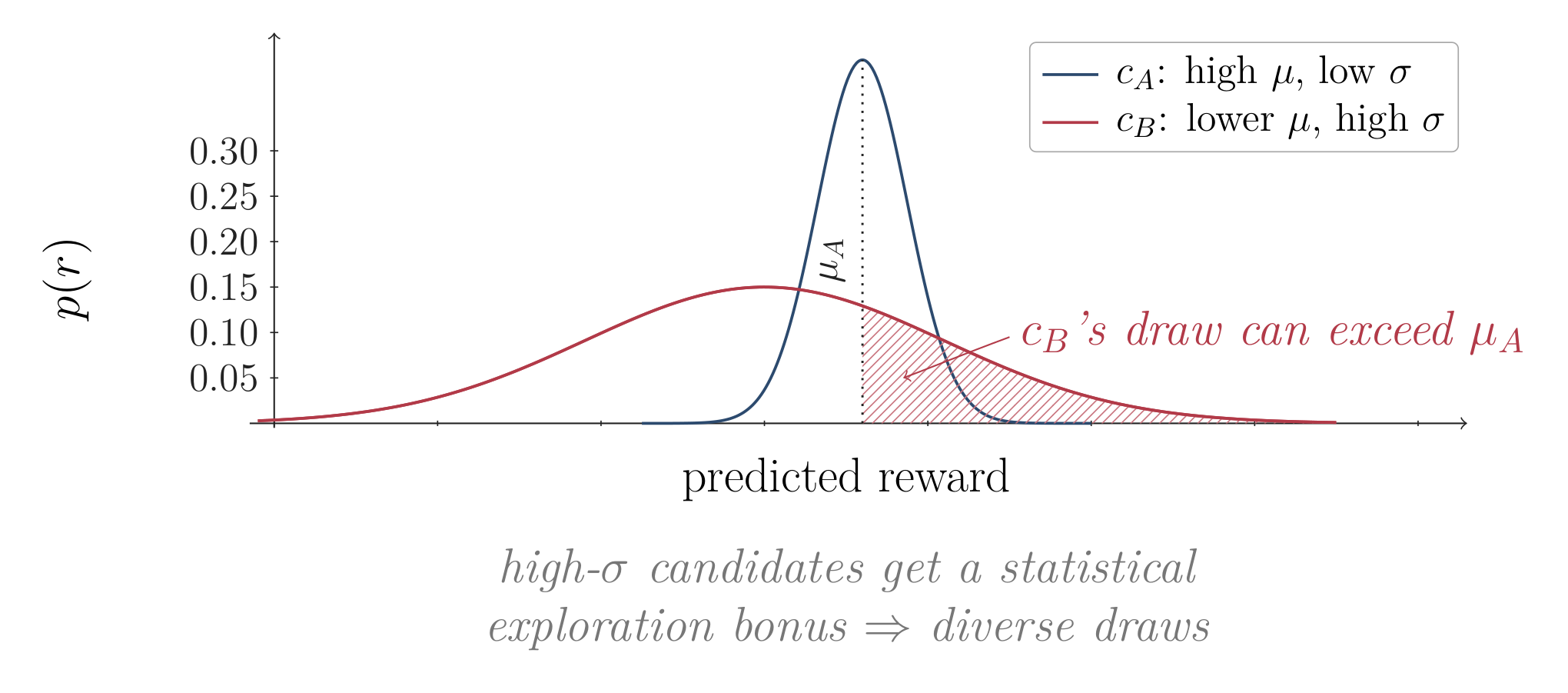}
  \caption{\textbf{Thompson sampling.} Acquisition rule: a per-candidate sampled surrogate reward $\hat r_i = \mu_i + \varepsilon_i \sigma_i$ over the ensemble's mean/disagreement, oracle-evaluating the top-$K$ by $\hat r_i$. Different draws of $\varepsilon$ favor different molecular regions.}
  \label{fig:diag_thompson}
\end{figure}

\begin{figure}[H]
  \centering
  \includegraphics[width=0.95\textwidth]{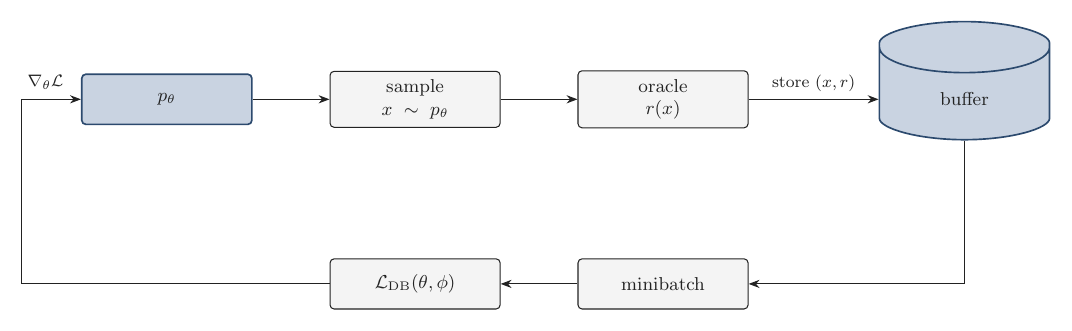}
  \caption{\textbf{Replay buffer.} Each generated $(x, r)$ pair is stored in a fixed-capacity buffer with priority eviction; mini-batches are drawn from the buffer (with a fixed fresh-fraction) so the gradient sees a time-averaged version of the policy distribution rather than the sharp on-policy one.}
  \label{fig:diag_buffer}
\end{figure}

\subsection{Inference-time search families}

\begin{figure}[H]
  \centering
  \includegraphics[width=0.55\textwidth]{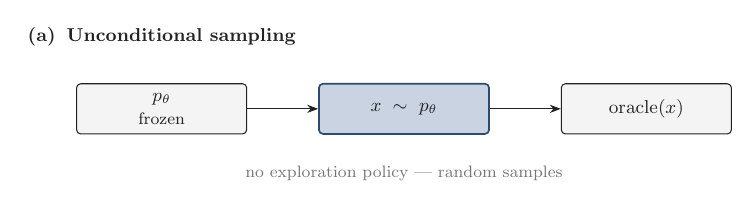}
  \caption{\textbf{Unconditional sampling.} The simplest baseline: sample from $p_\theta$ and score every sample under the oracle, with no exploration policy.}
  \label{fig:diag_uncond}
\end{figure}

\begin{figure}[H]
  \centering
  \includegraphics[width=0.7\textwidth]{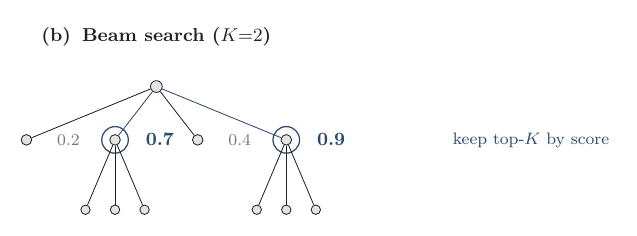}
  \caption{\textbf{Beam search.} At each denoising step, the top-$K$ partial trajectories by score are kept and expanded; the rest are pruned.}
  \label{fig:diag_beam}
\end{figure}

\begin{figure}[H]
  \centering
  \includegraphics[width=0.7\textwidth]{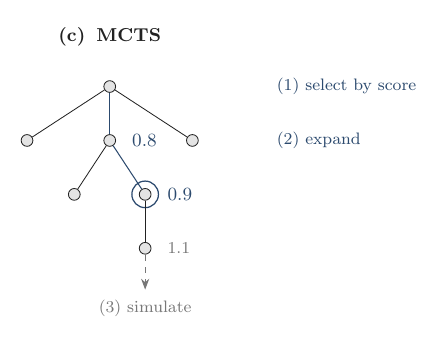}
  \caption{\textbf{Monte Carlo Tree Search.} Each node is a partial denoising state; UCB selects the next expansion based on visit counts and value estimates from rollouts.}
  \label{fig:diag_mcts}
\end{figure}

\begin{figure}[H]
  \centering
  \includegraphics[width=0.7\textwidth]{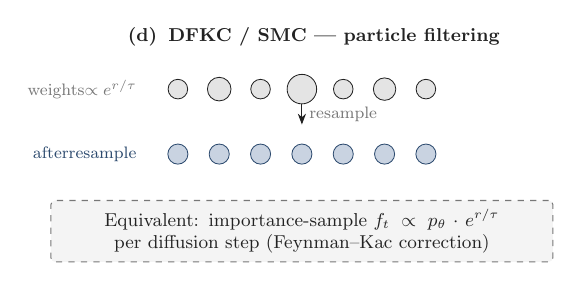}
  \caption{\textbf{Discrete Feynman-Kac correctors (DFKC) / SMC.} A particle population is reweighted at each denoising step by a tilt proportional to the (estimated) reward, then resampled when effective sample size drops below threshold.}
  \label{fig:diag_dfkc}
\end{figure}

\subsection{Surrogate and Tweedie shortcuts}

\begin{figure}[H]
  \centering
  \includegraphics[width=0.7\textwidth]{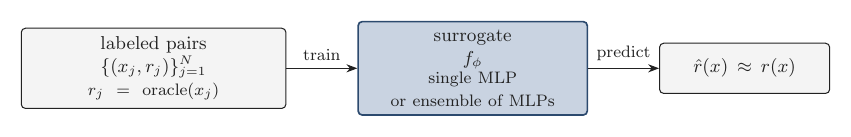}
  \caption{\textbf{Surrogate screening.} A cheap learned surrogate $\hat{r}(x_t)$ scores intermediate states in place of the expensive oracle, reducing per-output oracle cost from $O(L)$ to $O(1)$.}
  \label{fig:diag_surrogate}
\end{figure}

\begin{figure}[H]
  \centering
  \includegraphics[width=0.7\textwidth]{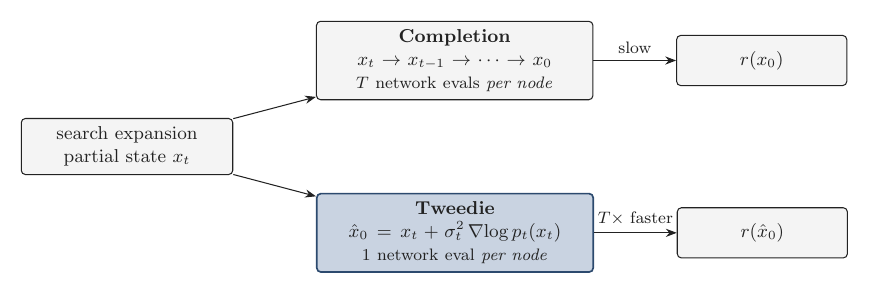}
  \caption{\textbf{Tweedie estimator.} The posterior-mean denoised state $\hat{x}_0 = \mathbb{E}[x_0 \mid x_t]$ is computed in a single denoising pass and scored directly, replacing a full rollout \citep{efron2011tweedie, chung2023dps}.}
  \label{fig:diag_tweedie}
\end{figure}

\clearpage

\section{Component Ablations and Diagnostics}\label{app:component-analysis}

This section moves from controlled single-component comparisons to diagnostics, joint effects, and transfer across fine-tuning objectives.

\subsection{Roles of Acquisition and Reward Shaping}\label{app:knob-interactions}

This subsection collects the longer descriptions of how Thompson-sampling acquisition and model-debiasing reward shaping individually act on the loop, moved here from Sec.~\ref{sec:design-explore} for brevity.

\begin{figure}[H]
  \centering
  \includegraphics[width=0.95\textwidth]{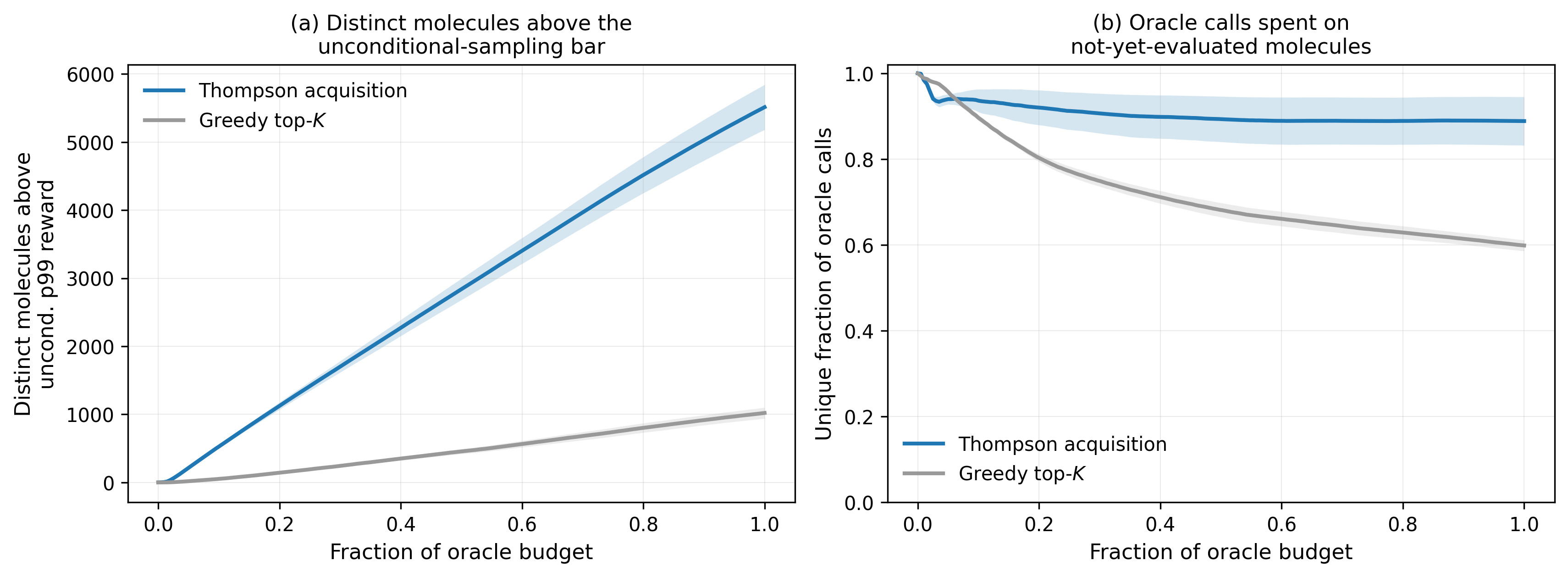}
  \caption{\textbf{What the uncertainty bonus buys at the acquisition step.} Full DDPP-LB loop (Thompson acquisition) vs.\ the same loop with Thompson disabled, which falls back to greedy top-$K$ by the surrogate's predicted mean; everything else is identical, so the contrast isolates the acquisition rule. Mean over six small-molecule targets $\times$ five seeds, shaded band $\pm 1$ SEM over targets. \textbf{(a)} Distinct molecules found above a per-target bar set by the $99$th percentile of everything the unconditional best-of-$N$ baseline ever evaluated on that target; the bar is external to both methods being compared, so it does not flatter either. Thompson finds $4.5$--$7\times$ more such molecules at the shared final budget. \textbf{(b)} Fraction of oracle calls spent on molecules not already evaluated, which needs no threshold at all. Greedy acquisition keeps re-querying its current favourites and spends roughly $40\%$ of its budget on repeats ($55$--$64\%$ unique, vs.\ $65$--$99.7\%$ for Thompson), which is the mechanism behind the gap in (a).}
  \label{fig:thompson_acquisition}
\end{figure}

\paragraph{Thompson sampling vs.\ deterministic top-$\mu$ selection.} The contrast with deterministic top-$\mu$ selection is sharp (Fig.~\ref{fig:diag_thompson_body}): under deterministic selection a candidate $c_B$ with $\mu_B < \mu_A$ never gets oracle-evaluated, no matter how uncertain it is, so the training data $(x, r(x))$ that downstream finetuning sees is concentrated on the current surrogate argmax. As the ensemble disagreement $\sigma_B$ grows away from zero, the Thompson rule smoothly transitions from this hard-greedy regime to a stochastic exploration bonus, the standard way to guard against local optima \citep{chapelle2011empirical}. The acquisition step is itself modifying which samples from $p_{\theta_t}$ enter the evaluated set and therefore which data the gradient computes against. Fig.~\ref{fig:thompson_acquisition} measures this directly: against a reward bar set by unconditional sampling rather than by either method, Thompson acquisition clears the bar with $4.5$--$7\times$ more distinct molecules, and the threshold-free view shows why --- greedy selection spends roughly $40\%$ of its oracle budget re-evaluating molecules it has already scored.

\paragraph{Model debiasing as training-dynamics shaping.} At the outer-loop level, $x$ names the candidate being trained on; inside the diffusion objective, this term may be evaluated on the noised or partially denoised state used by the loss. Model debiasing therefore acts on the diffusion training dynamics rather than introducing a separate density estimator over completed molecules. This term favors candidates that are high reward but not already high likelihood under the current generator. It therefore complements Thompson sampling rather than replacing it: Thompson sampling explores uncertainty in the reward surrogate before oracle evaluation, while model debiasing reshapes the training signal after oracle evaluation to avoid repeatedly reinforcing the generator's current modes. Fig.~\ref{fig:diag_debiasing_body} illustrates the redistribution: under the unshaped reward the dominant prior mode keeps absorbing gradient mass; with debiasing the smaller modes get lifted above their prior heights.

\subsection{Individual Component Ablations}\label{app:knobs}

Each figure isolates one design-choice knob from group~(iii) of Table~\ref{tab:methods}, holding the other knobs at their reference defaults (Table~\ref{tab:hparams}). Every figure is a $3 \times 2$ grid: rows are top-1 best-so-far, top-10 mean, and top-10\% mean; columns are cumulative oracle calls (left) and GPU hours (right).

\begin{figure}[H]
  \centering
  \includegraphics[width=0.49\textwidth]{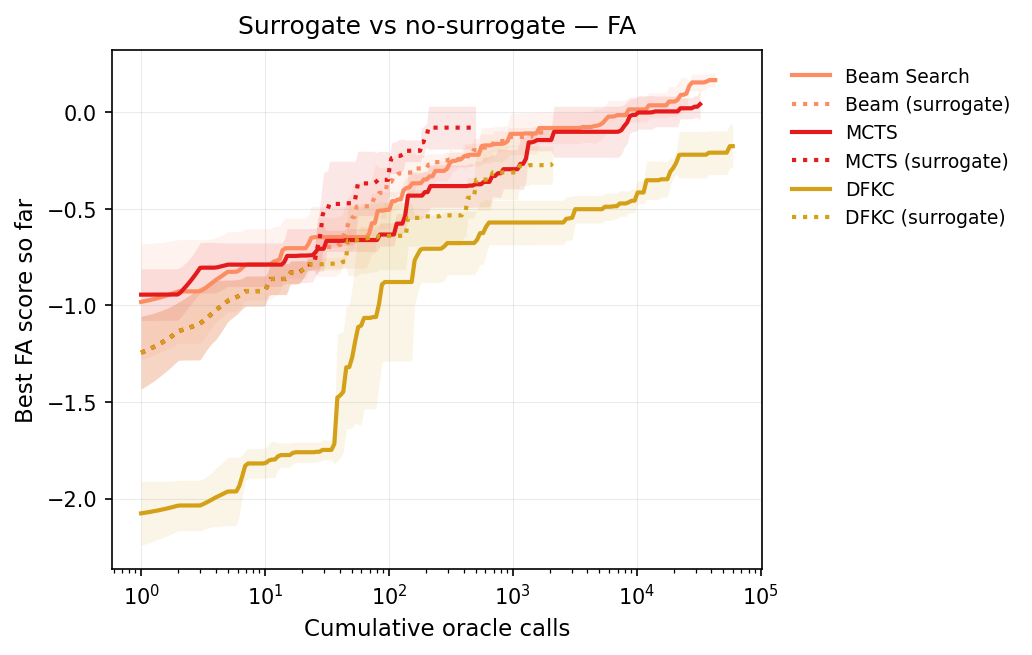}\hfill
  \includegraphics[width=0.49\textwidth]{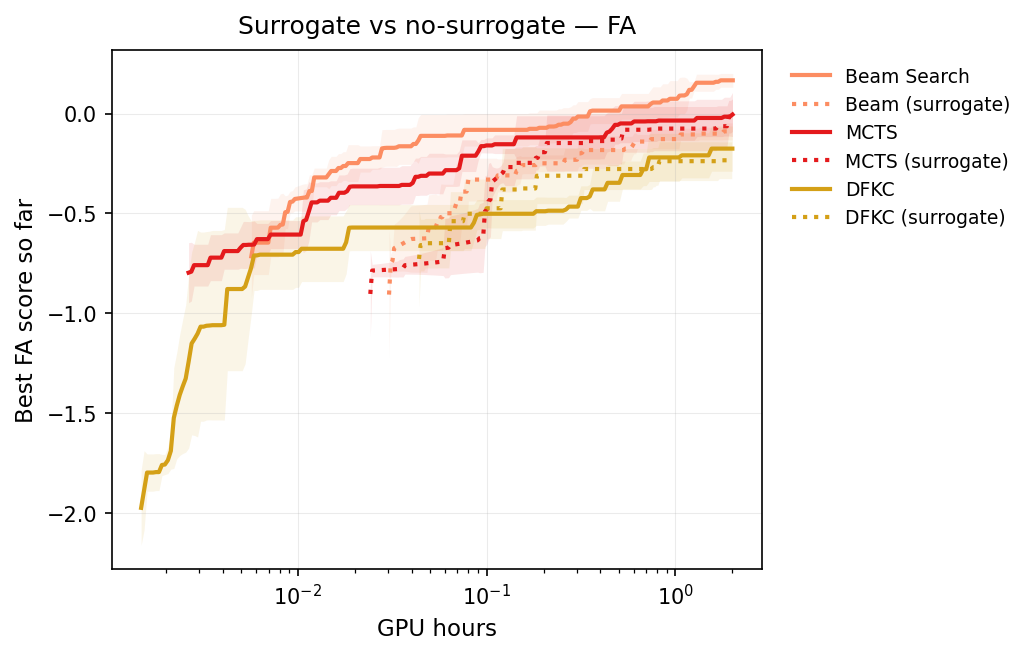}
  \caption{\textbf{Surrogate screening on inference-time search.} Beam, MCTS, and DFKC each compared against their surrogate-pre-filter variant (cheap MLP scores all candidates, oracle only sees the top-$K$). Solid: oracle-on-everything. Dashed: surrogate pre-filter. Seed-averaged ($n{=}5$) top-1 best FA against cumulative oracle calls (left) and GPU hours (right). Surrogate screening dominates plain search on both axes across all three families: each oracle call is spent on a candidate that already cleared the cheap filter, so the per-call yield is higher.}
  \label{fig:surrogate_search}
\end{figure}

\begin{figure}[H]
  \centering
  \includegraphics[width=0.49\textwidth]{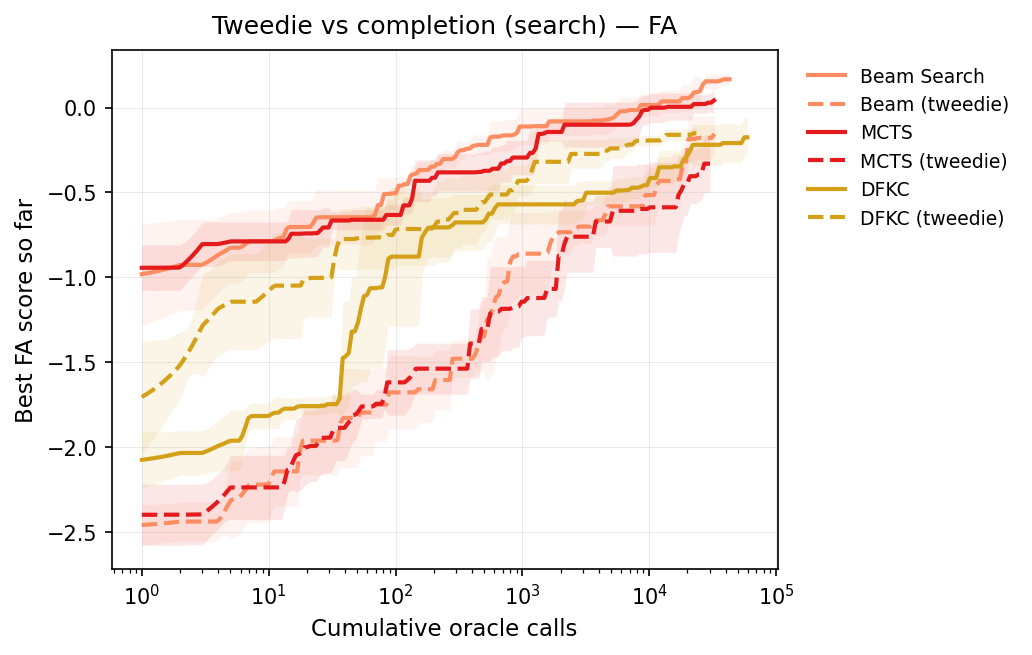}\hfill
  \includegraphics[width=0.49\textwidth]{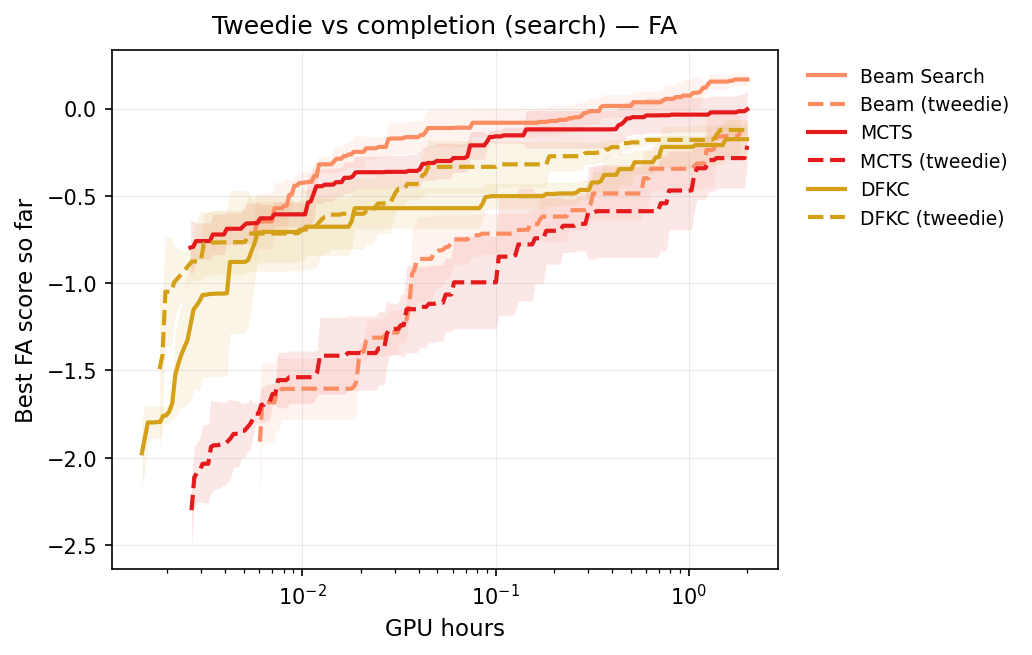}
  \caption{\textbf{Tweedie vs.\ completion (inference-time search).} Same beam/MCTS/DFKC families as above. Solid: completion (full denoising rollout to score). Dashed: Tweedie posterior-mean estimate $\hat{x}_0 = \mathbb{E}[x_0 \mid x_t]$ in place of the rollout. Seed-averaged ($n{=}5$). The Tweedie estimator's noise in the discrete setting is reflected directly in score: every Tweedie variant underperforms its completion counterpart, and the gap is largest for DFKC. The diffusion-pass savings do not translate into top-1 oracle score because the per-step score estimate is too noisy to steer the search.}
  \label{fig:tweedie_search}
\end{figure}

\paragraph{Reading the top-10\% row.} For inference-time search comparisons (Appendix~\ref{app:search}, e.g.\ Figs.~\ref{fig:surrogate_search} and~\ref{fig:tweedie_search}) the underlying generative distribution is fixed and only the selection procedure changes. Top-10\% mean in those figures reflects sampling efficiency from a static distribution and saturates as the upper tail of $p_{\text{pre}}$ is exhausted. In the active ablations below the policy $q_\theta$ is itself updated, so a continued rise in top-10\% mean is direct evidence that the sampling distribution is shifting --- not just that we are picking better samples from the same distribution.

\subsubsection{CVaR reward shaping}

\begin{figure}[H]
  \centering
  \includegraphics[width=0.95\textwidth]{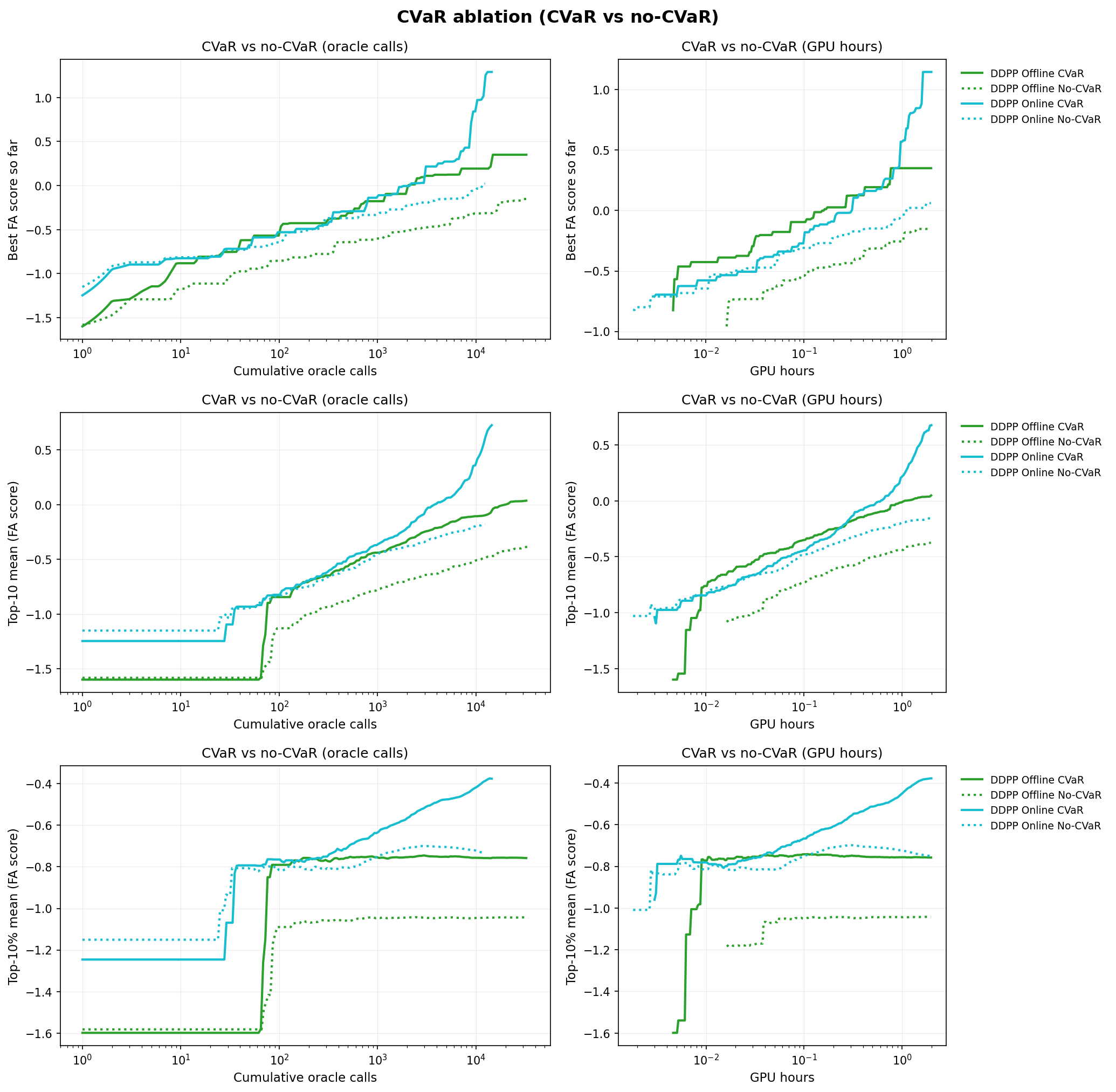}
  \caption{\textbf{CVaR ablation.} Four cells on DDPP-LB: solid = CVaR-on, dotted = CVaR-off; green = offline, teal = online. Seed-averaged ($n{=}5$) with $\pm 1$ std bands. CVaR lifts all three metrics in both regimes; the gain on top-10\% mean (bottom row) for the online~$+$~CVaR cell --- which continues to climb to $\sim 0.5$ rather than saturating --- confirms that the CVaR-shaped gradient shifts the policy distribution itself rather than only re-ranking samples from a fixed distribution.}
  \label{fig:cvar_ablation}
\end{figure}

\subsubsection{Density Entropy Regularization}

\begin{figure}[H]
  \centering
\includegraphics[width=0.95\textwidth]{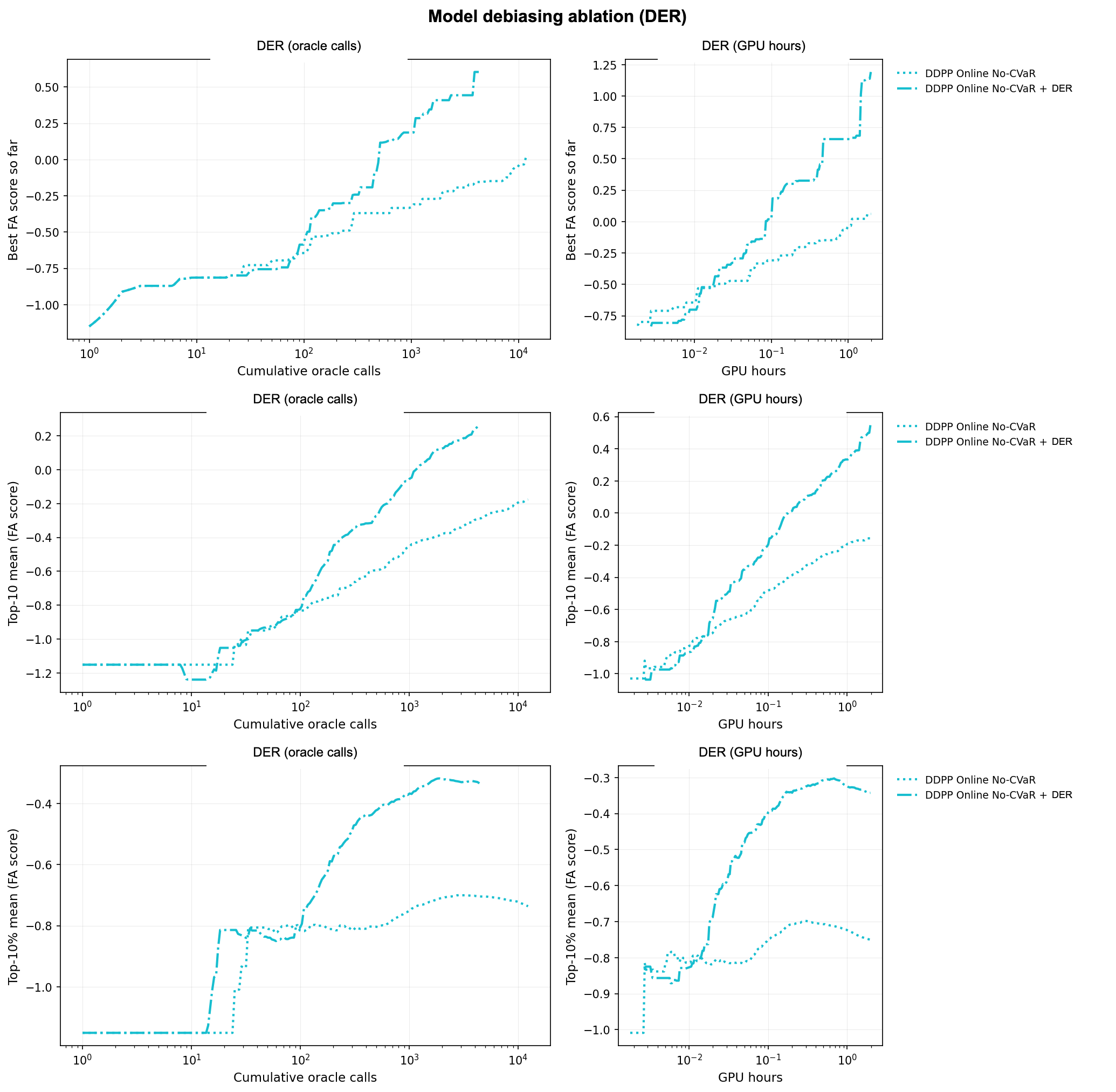}
  \caption{\textbf{Model-debiasing ablation.} Two cells, both online DDPP-LB with CVaR off: solid (with debiasing) vs.\ dotted (without). Seed-averaged ($n{=}5$). Holding CVaR off isolates the debiasing knob from the upper-tail reweighting in Fig.~\ref{fig:cvar_ablation}. Debiasing lifts top-1 by $\sim +0.6$ and top-10\% mean by a comparable margin, again indicating a genuine shift in $q_\theta$ and not just better sample selection.}
  \label{fig:debiasing_ablation}
\end{figure}

\subsubsection{Replay buffer}

\begin{figure}[H]
  \centering
  \includegraphics[width=0.95\textwidth]{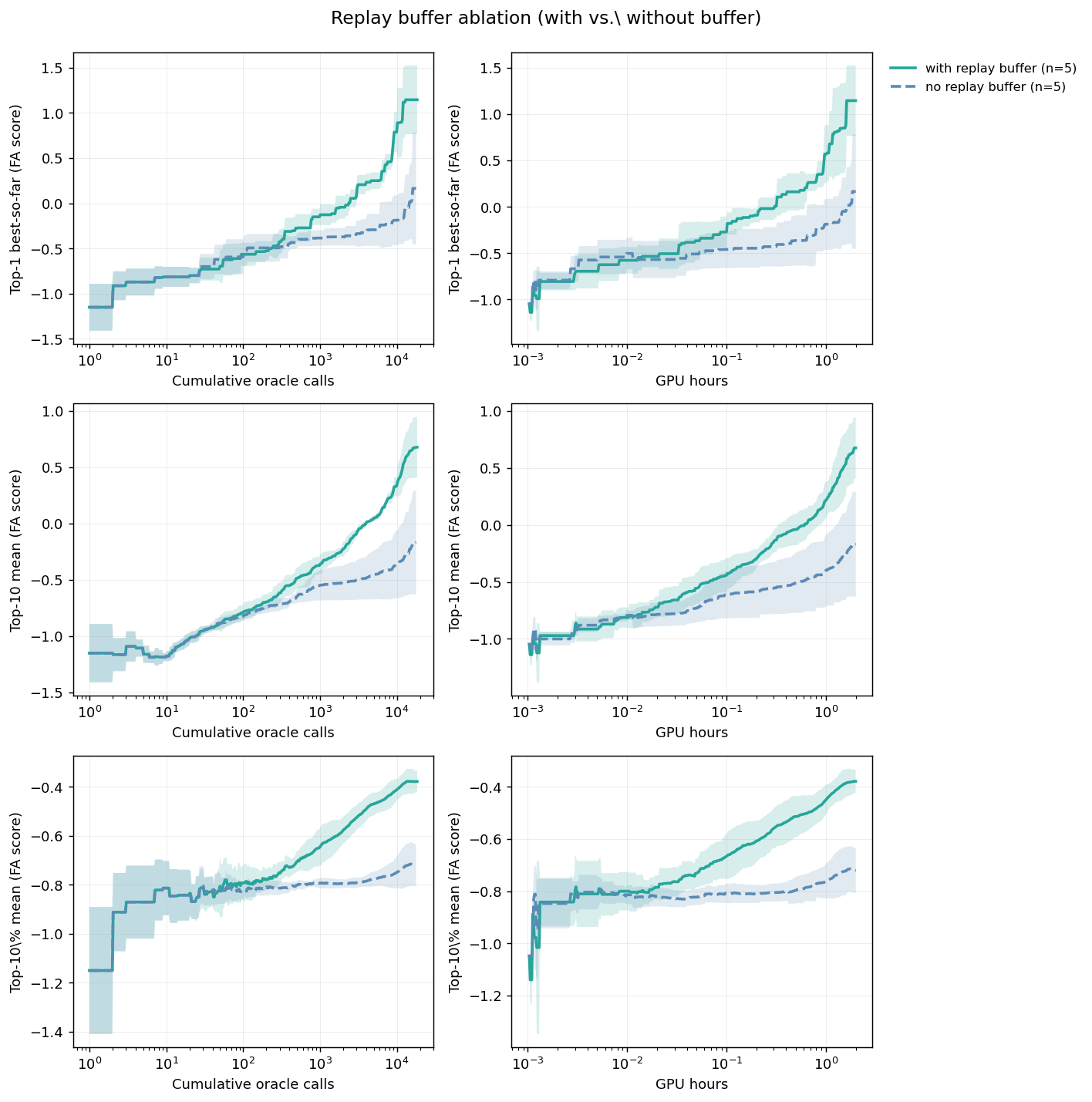}
  \caption{\textbf{Replay-buffer ablation.} Two cells, both online DDPP-LB$+$CVaR: solid (with replay buffer) vs.\ dashed (no buffer). Seed-averaged ($n{=}5$). Top-1 separates the cells modestly. The clearest signal is in top-10\% mean (bottom row), where the buffer-on cell continues to climb past $-0.4$ while the buffer-off cell plateaus near $-0.7$. The buffer is shifting where the policy puts mass over the run, not merely re-presenting a fixed set of samples to the gradient.}
  \label{fig:buffer_ablation}
\end{figure}

\subsubsection{Thompson acquisition}

\begin{figure}[H]
  \centering
  \includegraphics[width=0.95\textwidth]{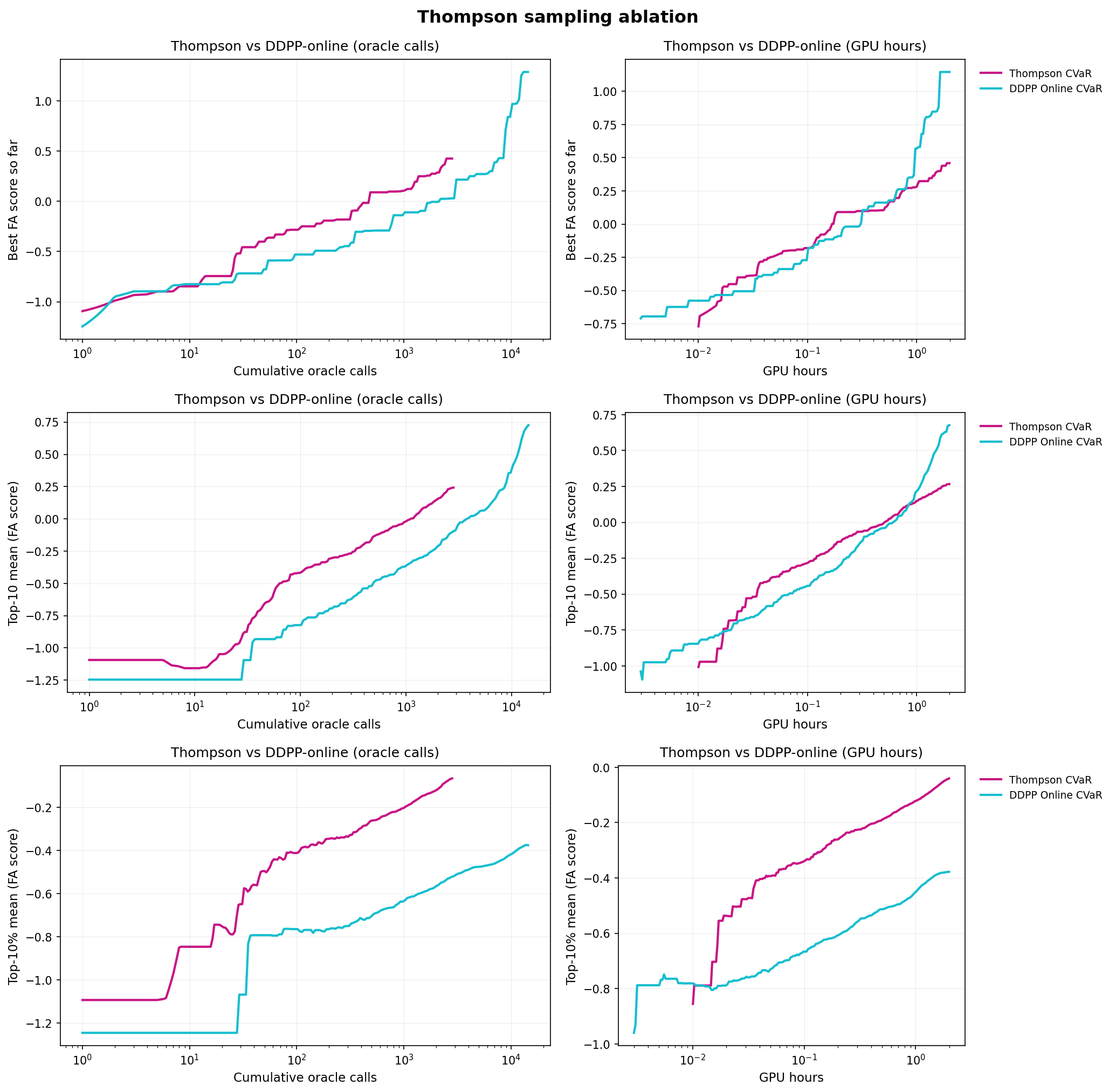}
  \caption{\textbf{Thompson-sampling ablation.} Pink: Thompson on top of online DDPP-LB$+$CVaR. Cyan: same configuration with deterministic top-$K$ selection by ensemble mean instead of Thompson draws. Seed-averaged ($n{=}5$). Top-1 is comparable; top-10\% mean (bottom row) is where the contrast is sharpest --- Thompson reaches $\sim -0.05$ vs.\ $\sim -0.4$ for deterministic selection, a clear distribution shift toward higher-reward modes that comes from Thompson diversifying which candidates the oracle is spent on rather than from any change to the gradient itself.}
  \label{fig:thompson_ablation}
\end{figure}

\subsubsection{Invalid-output penalty}

\begin{figure}[H]
  \centering
  \includegraphics[width=0.95\textwidth]{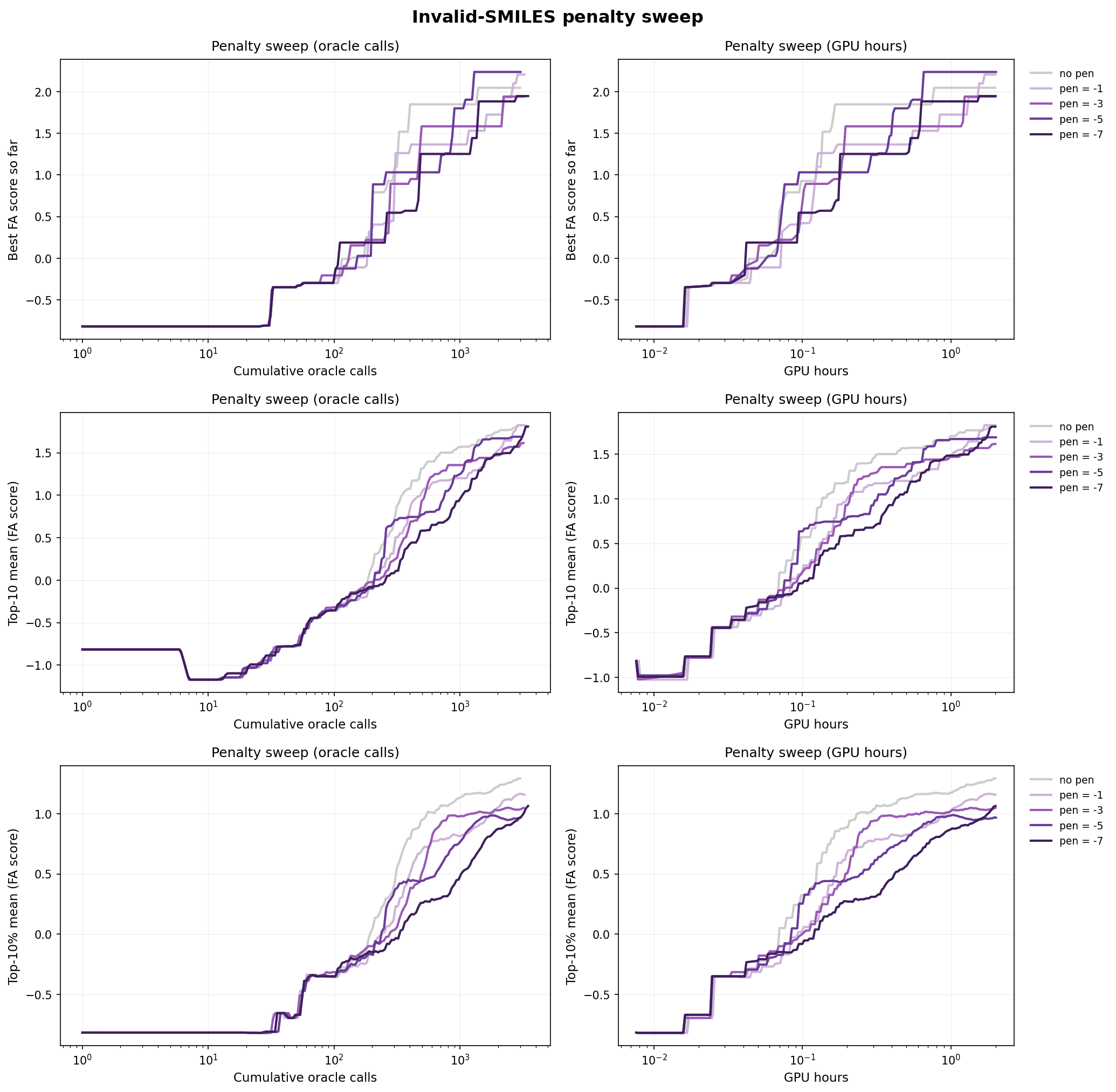}
  \caption{\textbf{Invalid-SMILES penalty sweep $r_{\text{inv}} \in \{0, -1, -3, -5, -7\}$.} On top of Approach~A $+$ Thompson. Light-to-dark purple = increasing penalty magnitude. Top-1 and top-10 narrow the gap between cells over time; top-10\% mean separates them most clearly, with smaller penalties ($-1$, $-3$) reaching slightly higher upper-tail FA but at the cost of validity (Fig.~\ref{fig:penalty_sweep}). $r_{\text{inv}} = -5$ achieves the best validity / top-1 FA trade-off across the sweep. Effect is not very signifcant on regular DDPP-LB but becomes much more significant with other exploratory methods, as seen in the ablation figures.}
  \label{fig:penalty_ablation}
\end{figure}

\begin{figure}[H]
  \centering
    \includegraphics[width=0.85\textwidth]{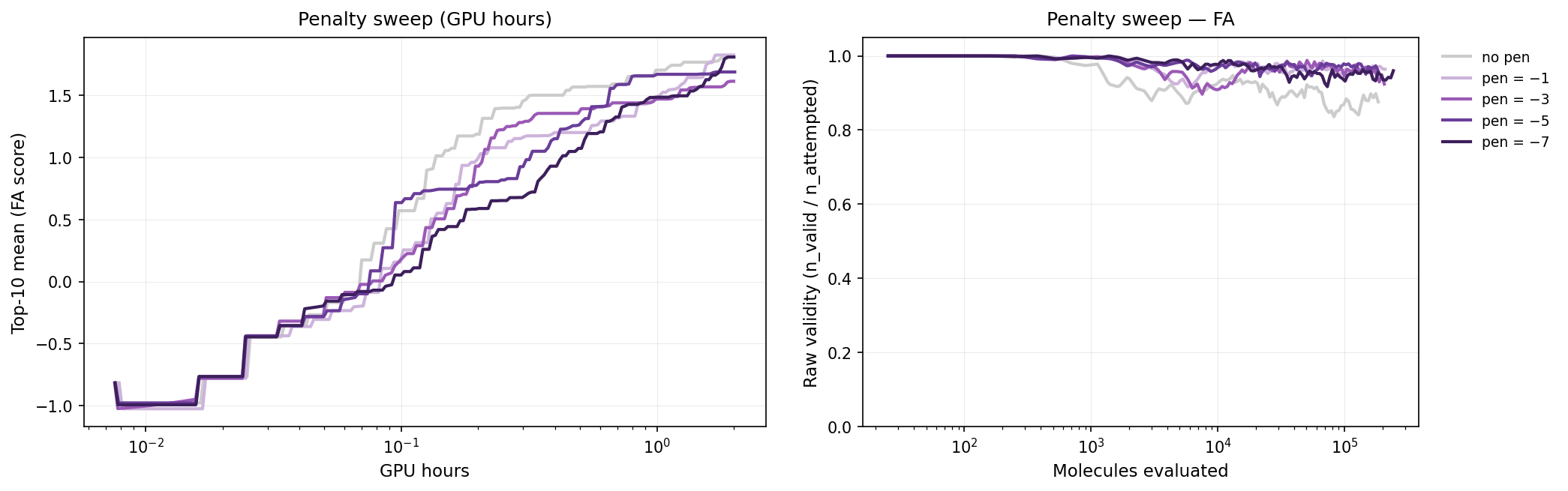}
  \caption{Validity panel paired with top-1 FA for the same penalty sweep as Fig.~\ref{fig:penalty_ablation}, used in the body to motivate the choice of $r_{\text{inv}} = -5$. The $r_{\text{inv}}{=}0$ line shows grammar collapse; all non-zero penalties restore validity.}
  \label{fig:penalty_sweep}
\end{figure}

\subsubsection{Per-family protein ablations}
Table~\ref{tab:protein_knobs} reports the leave-one-out effect per protein family, which the aggregated protein panels of Fig.~\ref{fig:barplot_ablation} average over.

\begin{table}[H]
\centering
\small
\begin{tabular}{@{}lrrr@{}}
\toprule
\textbf{Component removed} & \textbf{GB1} & \textbf{TrpB} & \textbf{CreiLOV} \\
\midrule
Thompson acquisition   & $+1.038$ & $+0.279$ & $+0.090$ \\
CVaR shaping           & $+1.037$ & $+0.042$ & $+0.014$ \\
Replay buffer          & $+0.694$ & $+0.085$ & $+0.042$ \\
Density Entropy Reg.\  & $+0.179$ & $+0.035$ & $-0.184$ \\
Invalid-output penalty & $ 0.000$ & $-0.006$ & $ 0.000$ \\
\bottomrule
\end{tabular}
\caption{\textbf{Per-family leave-one-out effects on protein tasks.} Loss in best fitness from removing each component. Positive means the component helps. Only Thompson acquisition helps on all three families. The invalid-output penalty never fires: the protein sampler emits fixed-length sequences over the canonical amino acids, so invalid sequences cannot occur.}
\label{tab:protein_knobs}
\end{table}

The ordering is per-family rather than per-domain, and tracks prior coverage rather than modality. GB1 behaves like the small-molecule regime --- large gains from Thompson, CVaR, and replay --- and it is also the family whose prior is by far the least diverse, with only $35$ sequence clusters at $80\%$ identity in the training split against $33{,}851$ for TrpB and $185{,}774$ for CreiLOV. Density Entropy Regularization is the weakest knob throughout and turns negative on CreiLOV, which is what the prior-coverage account predicts: DER pushes sampling away from sequences the model already finds likely, and that pays off only when high-fitness sequences are ones the prior considers unlikely. Here the opposite holds --- scored under the frozen prior, high-fitness sequences are the \emph{more} typical ones on all three families (reward--NLL correlations $-0.24$ to $-0.74$) --- so pushing off-prior moves the search away from fitness.

\subsubsection{Molecular length under Density Entropy Regularization}
DER is not length-neutral for molecules. Table~\ref{tab:derlen} shows that the full method shifts evaluated candidates toward longer tokenized strings, especially early in the run, and that the shift persists within matched reward ranges, so it is not merely a by-product of reaching higher-reward regions.

\begin{table}[H]
\centering
\small
\begin{tabular}{@{}lrr@{}}
\toprule
\textbf{Window} & \textbf{Full (DER)} & \textbf{No-DER} \\
\midrule
First $20\%$ of oracle calls & $54.4$ & $39.4$ \\
Last $20\%$ of oracle calls  & $49.6$ & $41.2$ \\
\midrule
\multicolumn{3}{@{}l}{\textit{By FA reward range, last $20\%$}} \\
$<-2$        & $47.0$ & $39.4$ \\
$[-2,-1)$    & $41.1$ & $40.0$ \\
$[-1,0)$     & $45.1$ & $39.7$ \\
$[0,1)$      & $54.3$ & $43.9$ \\
$[1,2)$      & $57.5$ & $56.8$ \\
$\ge 2$      & $55.0$ & $41.9$ \\
\bottomrule
\end{tabular}
\caption{\textbf{Molecule token length under Density Entropy Regularization.} Mean tokenized SMILES length of oracle-evaluated candidates.}
\label{tab:derlen}
\end{table}

This does not by itself indicate a failure mode. DER is intended to improve top-molecule discovery, and the full method does find better top-scoring molecules. Length also \emph{decreases} from the first to the last $20\%$ of oracle calls ($54.4 \to 49.6$), so DER does not simply keep inflating length; it expands the searched molecular regime early and then shapes the sampling posterior within that regime. For proteins the mechanism does not apply at all, since sequence length is fixed by family.

\subsubsection{Density estimator comparison}
\begin{figure}[H]
  \centering
  {\includegraphics[width=\textwidth]{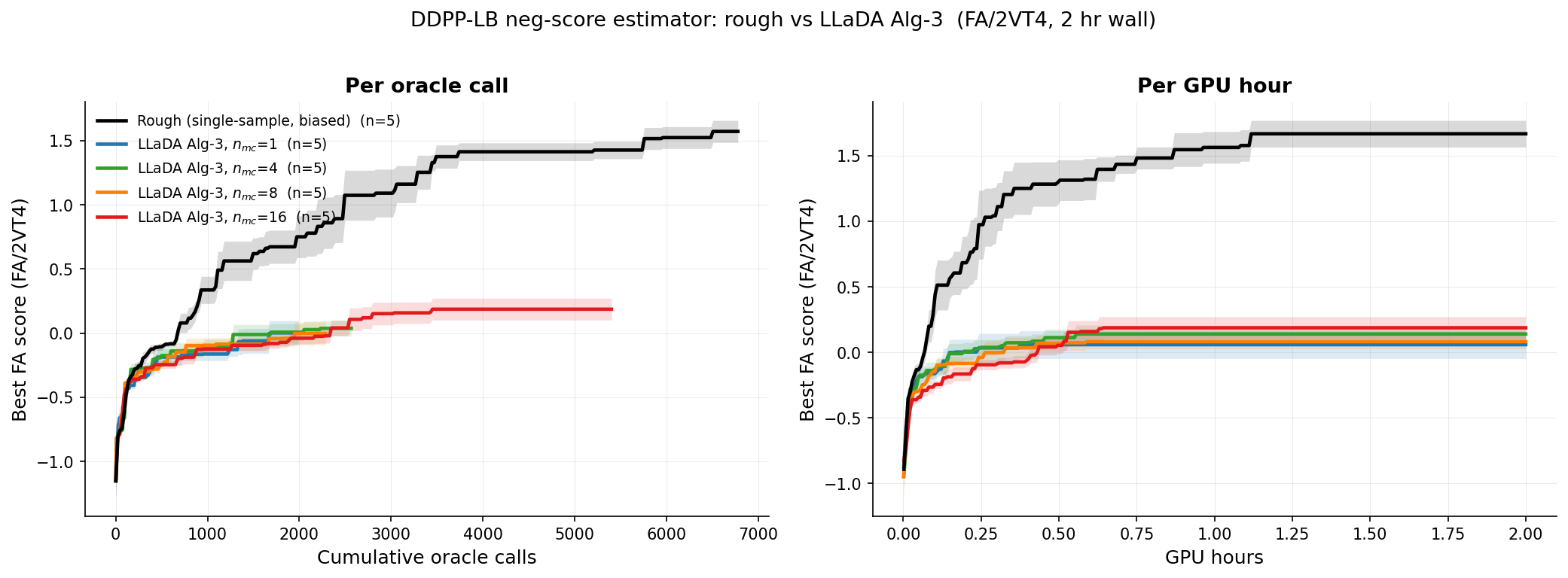}}\\[4pt]
  {\includegraphics[width=\textwidth]{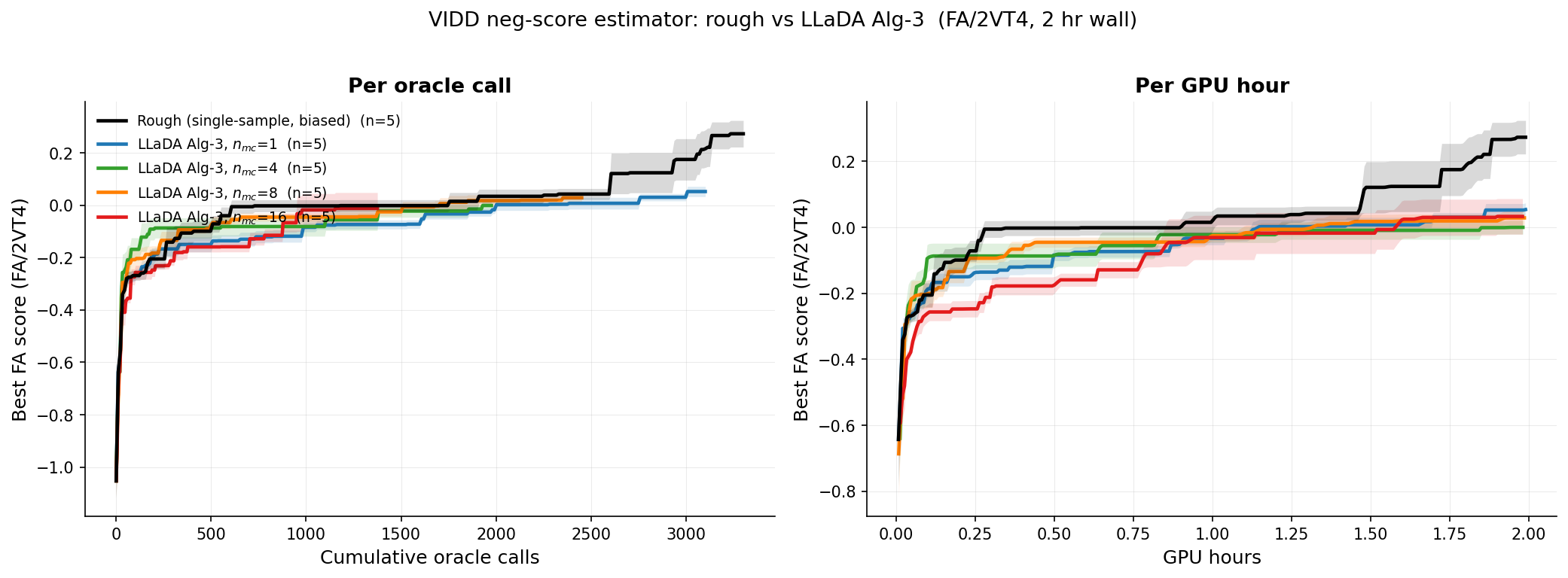}}
  \caption{{\textbf{Best-so-far oracle curves: cheap estimator vs.\ LLaDA-Alg.\,3 variants (FA/2VT4, 2 hr wall).} Each subplot shows best FA score per oracle call (left) and per GPU hour (right) for five density-entropy-regularization variants: single-pass cheap estimator (black) and LLaDA-Alg.\,3 with $n_{\text{mc}} \in \{1,4,8,16\}$ (colored). \textbf{Top:} DDPP-LB. \textbf{Bottom:} VIDD. The cheap estimator matches or exceeds LLaDA variants on oracle calls and is far more or at least equally efficient per GPU hour.}}
  \label{fig:negscore_estimator_curves}
\end{figure}

\begin{figure}[H]
  \centering
  {\includegraphics[width=\textwidth]{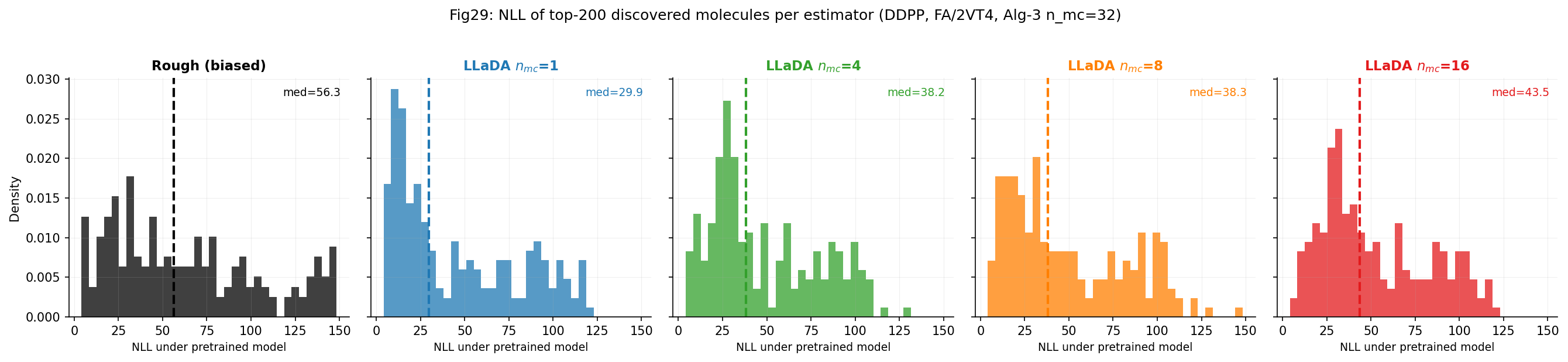}}\\[4pt]
  {\includegraphics[width=\textwidth]{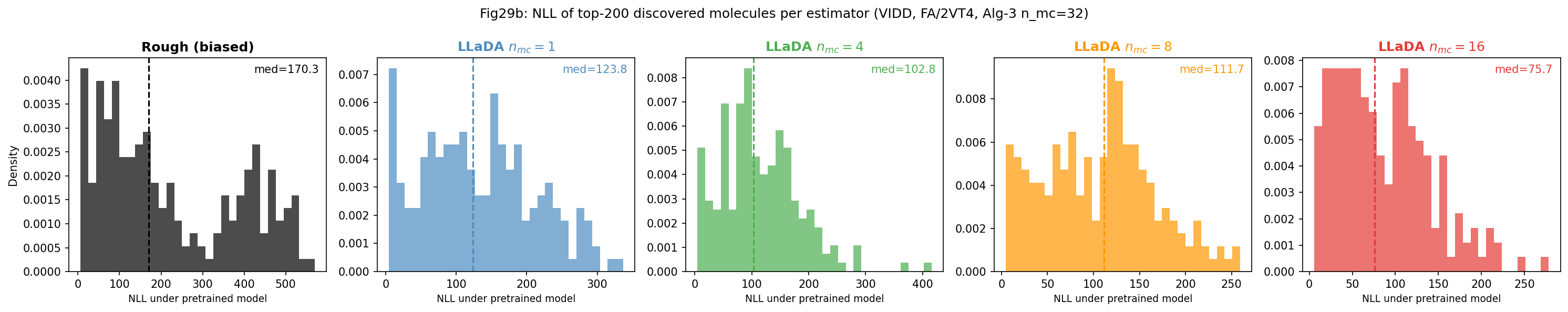}}
  \caption{{\textbf{NLL under the pretrained model for the top-200 discovered molecules, per estimator (FA/2VT4).} NLL evaluated post-hoc via LLaDA-Alg.\,3 ($n_{\text{mc}}{=}32$); dashed line = median. \textbf{Top:} DDPP-LB. \textbf{Bottom:} VIDD. The cheap estimator produces distributions shifted at least as far off the prior as the LLaDA variants, validating that the single-pass approximation achieves the intended exploration pressure.}}
  \label{fig:negscore_nll_dist}
\end{figure}

{\paragraph{Cheap estimator vs.\ LLaDA-Alg.\,3.}
We compare five density-entropy-regularization variants: the single-pass cheap estimator and LLaDA-Alg.\,3 with $n_{\text{mc}} \in \{1, 4, 8, 16\}$, for both DDPP-LB and VIDD on FA/2VT4. Fig.~\ref{fig:negscore_estimator_curves} shows best-so-far oracle score per oracle call and per GPU hour. The cheap estimator matches or exceeds all LLaDA variants on oracle-call efficiency and is substantially faster per GPU hour due to requiring only a single forward pass. Fig.~\ref{fig:negscore_nll_dist} shows the distribution of NLL under the pretrained model for the top-200 discovered molecules per variant, evaluated post-hoc via LLaDA-Alg.\,3 ($n_{\text{mc}}{=}32$). The cheap estimator induces at least as much off-prior shift as the LLaDA variants, confirming that the single-pass approximation is a computationally sound substitute.}

\subsection{Secondary Metrics and Distribution-Shift Diagnostics}\label{app:traces}

These panels report sliding-window means of pairwise Tanimoto diversity, QED, SA, and SMILES validity for the per-knob ablation cells of Appendix~\ref{app:knobs}. Window size 300, stride 200; one curve per cell, x-axis = molecules evaluated.

\begin{figure}[H]
  \centering
  \includegraphics[width=0.49\textwidth]{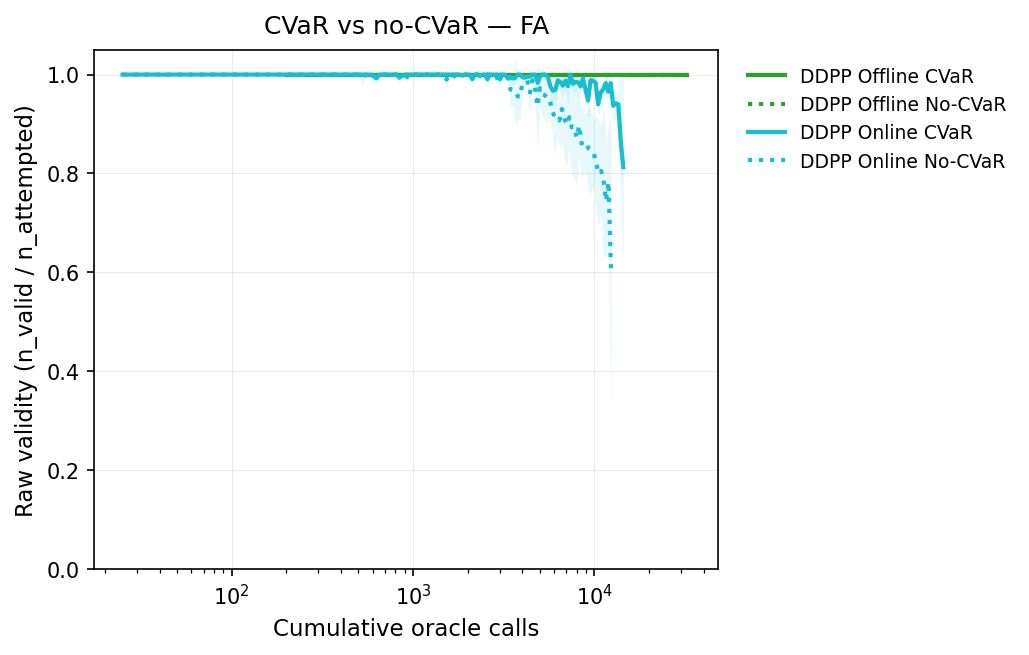}\hfill
  \includegraphics[width=0.49\textwidth]{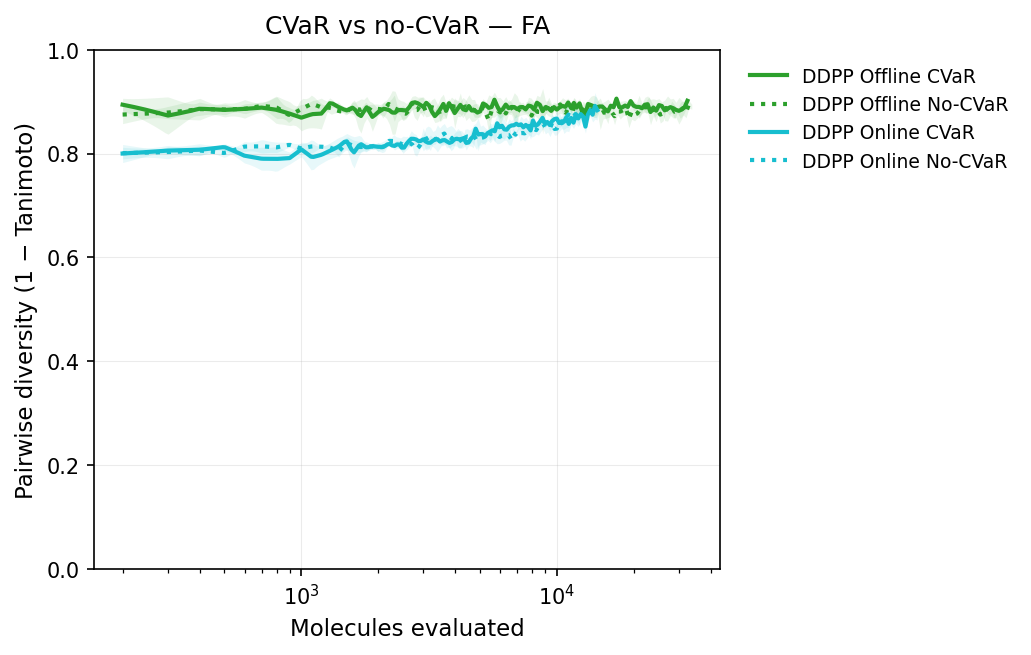}\\[0.5ex]
  \includegraphics[width=0.49\textwidth]{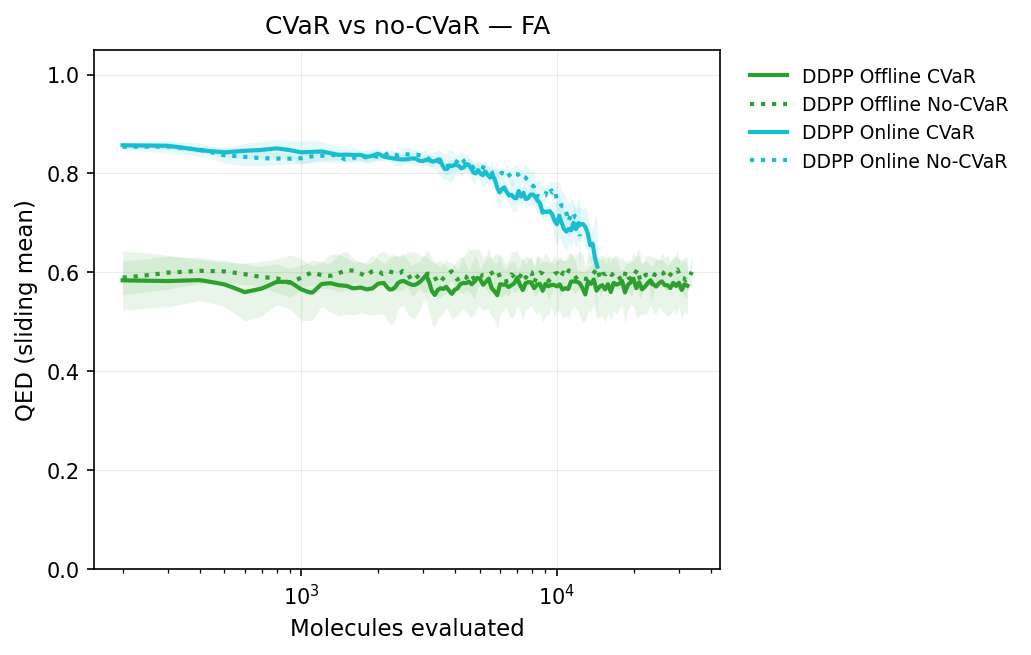}\hfill
  \includegraphics[width=0.49\textwidth]{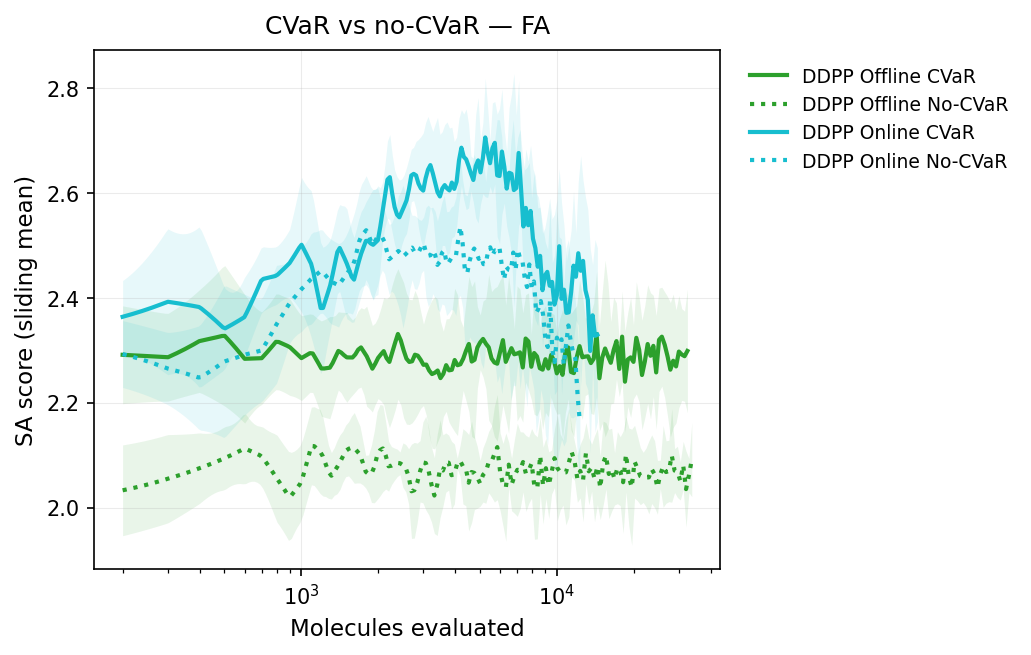}
  \caption{\textbf{CVaR ablation, secondary metrics.} Same four cells as Fig.~\ref{fig:cvar_ablation} (offline / online $\times$ CVaR-on / CVaR-off). Top-left: validity. Top-right: Tanimoto diversity. Bottom-left: QED. Bottom-right: SA. The top-1 FA gains in Fig.~\ref{fig:cvar_ablation} are not accompanied by validity collapse or QED/SA degradation, ruling out reward hacking on the CVaR knob.}
  \label{fig:traces_cvar}
\end{figure}

\begin{figure}[H]
  \centering
  \includegraphics[width=0.49\textwidth]{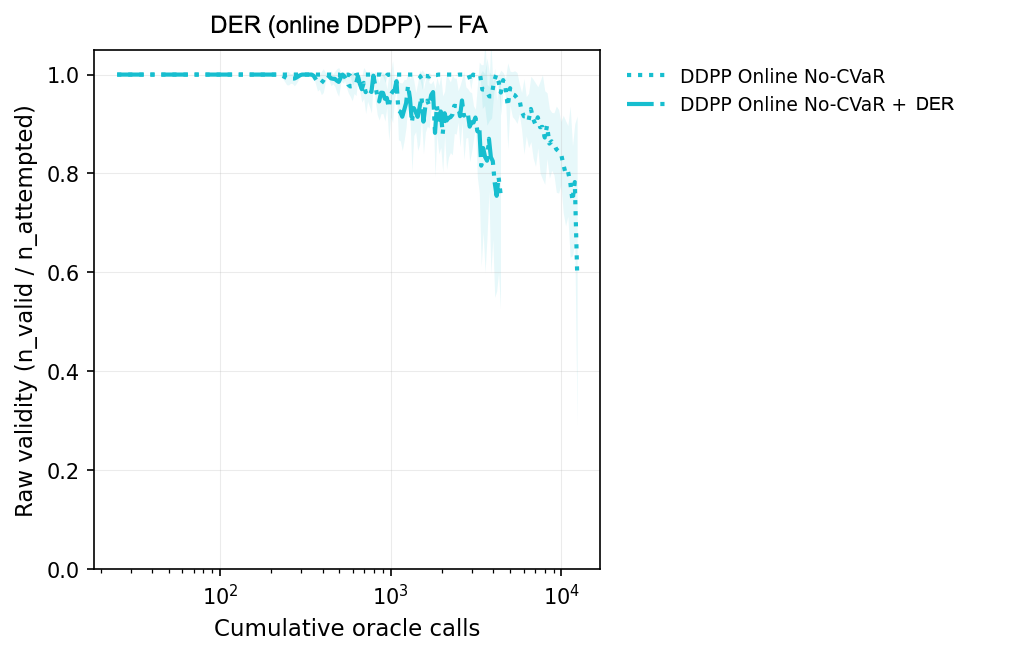}\hfill
  \includegraphics[width=0.49\textwidth]{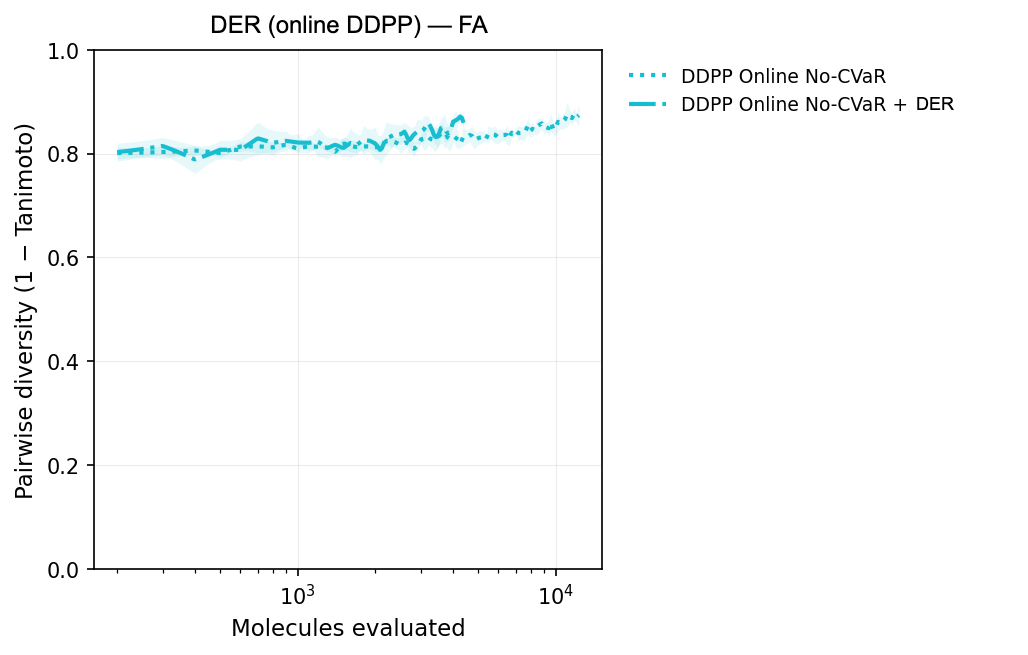}\\[0.5ex]
  \includegraphics[width=0.49\textwidth]{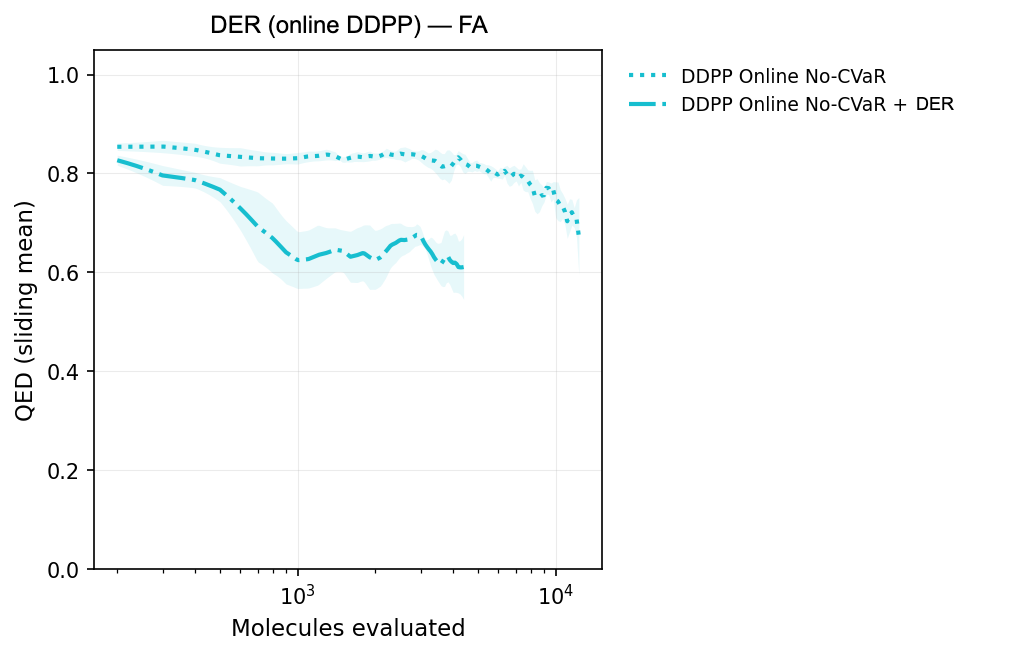}\hfill
  \includegraphics[width=0.49\textwidth]{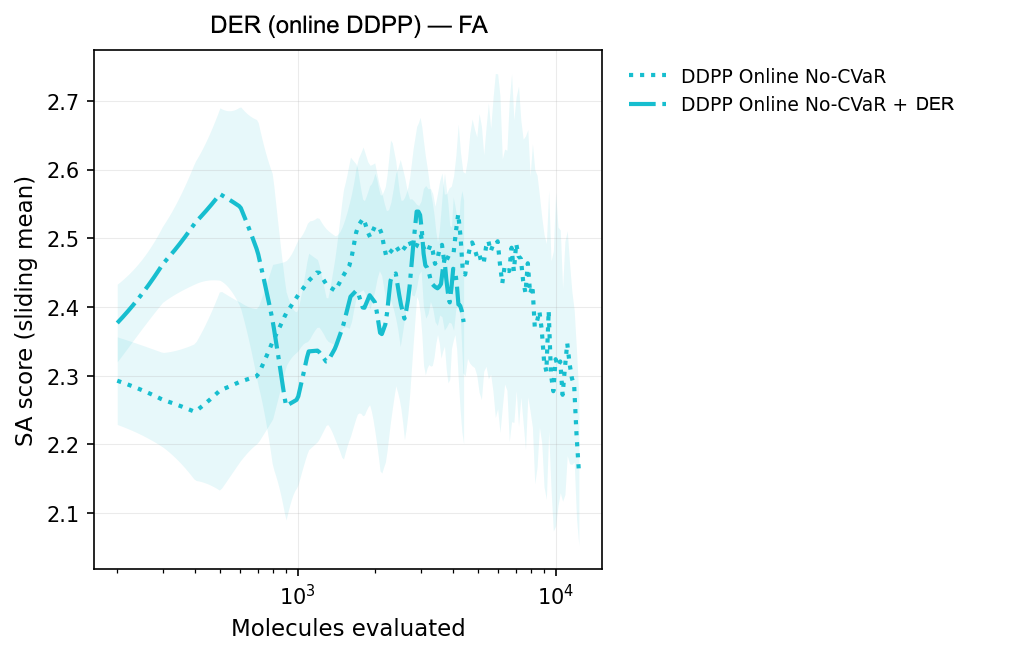}
  \caption{\textbf{Model-debiasing ablation, secondary metrics.} Same two cells as Fig.~\ref{fig:debiasing_ablation} (online no-CVaR with vs.\ without Density Entropy Regularization). Same panel layout as above. The top-1 FA gain from debiasing is not accompanied by validity or QED/SA degradation.}
  \label{fig:traces_debiasing}
\end{figure}

\begin{figure}[H]
  \centering
  \includegraphics[width=0.49\textwidth]{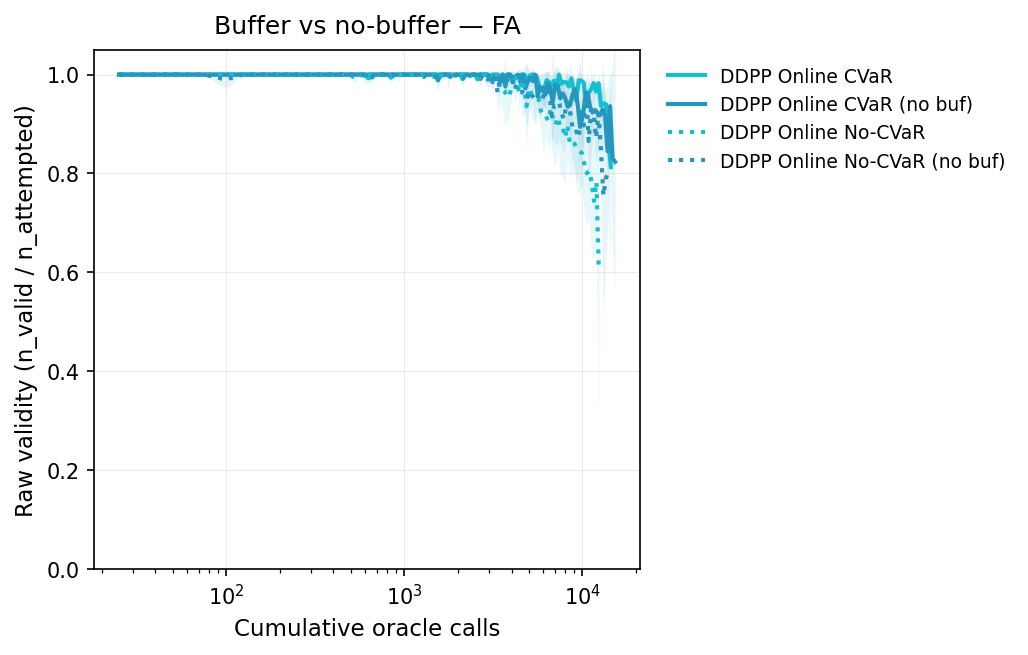}\hfill
  \includegraphics[width=0.49\textwidth]{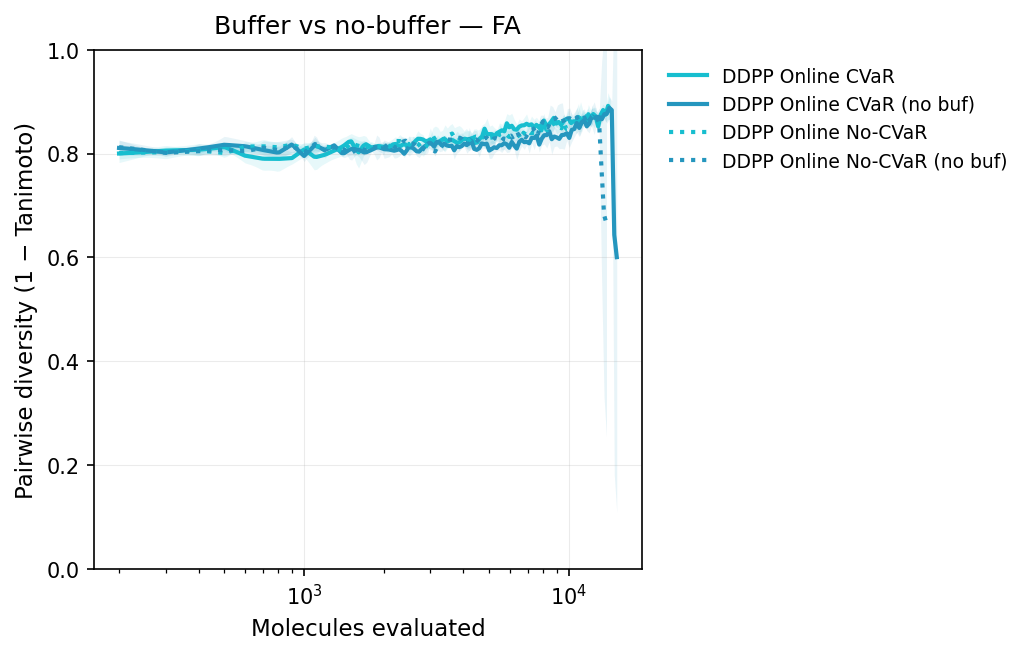}\\[0.5ex]
  \includegraphics[width=0.49\textwidth]{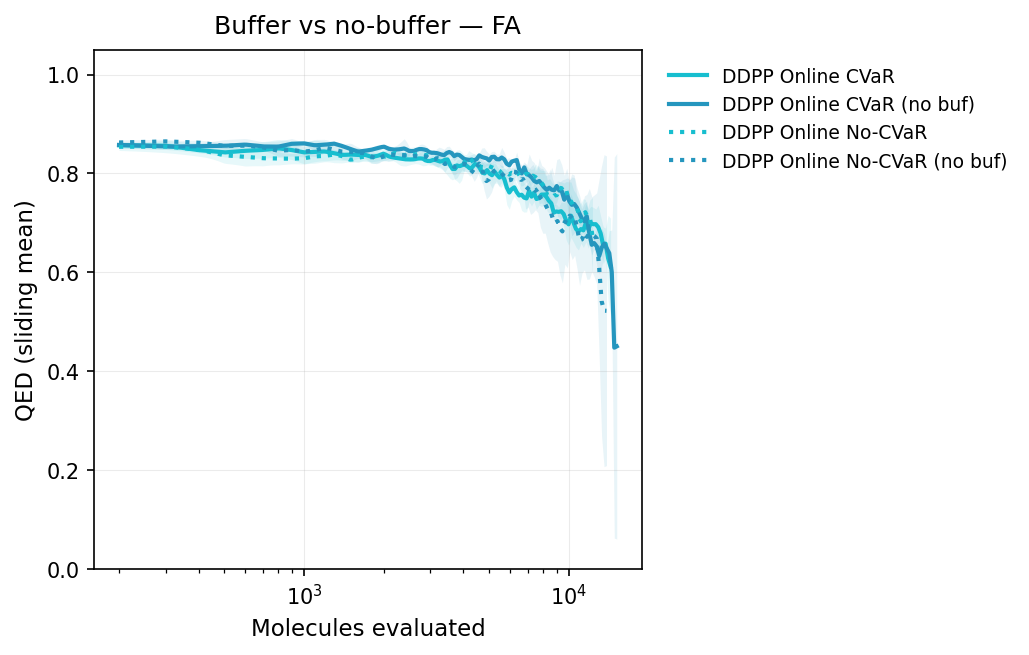}\hfill
  \includegraphics[width=0.49\textwidth]{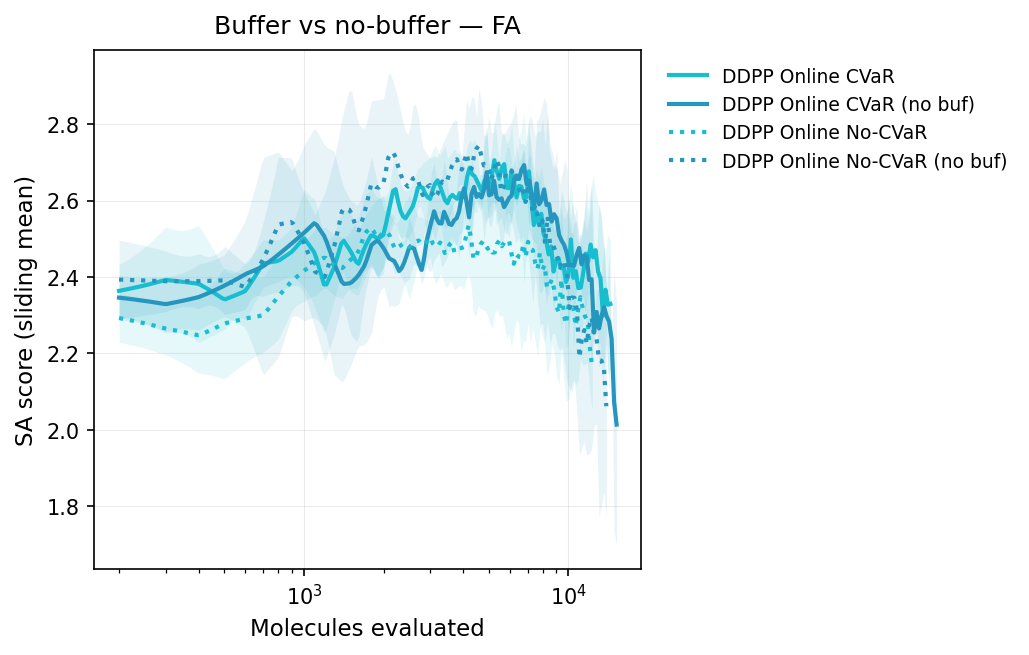}
  \caption{\textbf{Replay-buffer ablation, secondary metrics.} Same four cells as Fig.~\ref{fig:buffer_ablation} (buffer~$\times$~CVaR). Top-left: validity. Top-right: Tanimoto diversity. Bottom-left: QED. Bottom-right: SA. The buffer-on cells preserve diversity and SA more reliably than the buffer-off cells, consistent with the buffer's role as an implicit normalizer over the historical sample distribution (Sec.~\ref{sec:results}).}
  \label{fig:traces_buffer}
\end{figure}

The validity panel for the invalid-SMILES penalty sweep is reported directly in Fig.~\ref{fig:penalty_sweep}; per-cell QED/SA/Tanimoto for that sweep are not informative because the cells differ only in the validity-restoration penalty.

\begin{figure}[H]
  \centering
  \includegraphics[width=\textwidth]{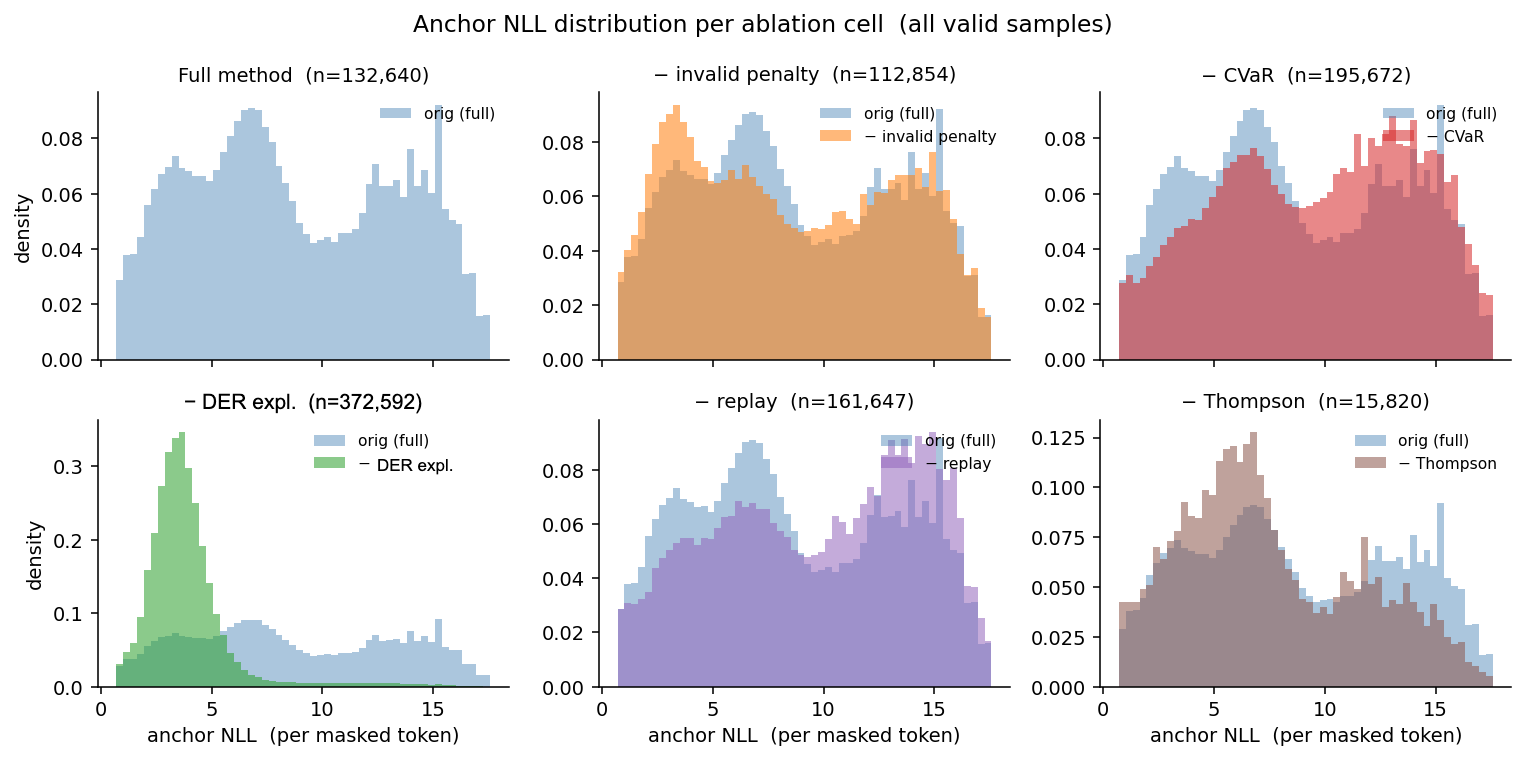}
  \caption{\textbf{Anchor-NLL distribution per leave-one-out ablation cell.} Histogram of per-token anchor NLL $-\log p_{\text{pre}}(x_0 \mid x_t)$ at run end, computed over all valid samples (n in title). Top-left is the full method; the other five panels each remove one knob and overlay the resulting distribution against the full-method reference (blue). Removing the density entropy regularization / debiasing knob (bottom-left) collapses mass back toward the prior (low anchor NLL) far more than removing any other knob, identifying debiasing as the primary driver of distribution shift away from $p_{\text{pre}}$. Removing CVaR (top-right) reduces the high-NLL tail, consistent with CVaR's role in pushing the policy toward the upper-reward region of the distribution. The remaining knobs (invalid penalty, replay buffer, Thompson) leave the bimodal shape qualitatively intact.}
  \label{fig:anchor_nll}
\end{figure}

\paragraph{Run-end anchor-NLL distribution.} The per-knob traces above measure how the secondary metrics evolve over time. A complementary view is the end-of-run snapshot of how far each cell's policy $q_\theta$ has drifted from the frozen prior $p_{\text{pre}}$, measured by the anchor NLL $-\log p_{\text{pre}}(x_0 \mid x_t)$ averaged over masked tokens (Fig.~\ref{fig:anchor_nll}). Cells here are the leave-one-out variants from Section~\ref{sec:results} (full method minus one knob at a time), not the per-knob composition cells of Appendix~\ref{app:knobs}.

\subsection{Joint Component Effects and the Untreated Baseline}\label{app:factorial}

Table~\ref{tab:factorial} in the main text reports the complete $2\times2\times2$ factorial over Density Entropy Regularization (DER), CVaR shaping, and Thompson acquisition. Leave-one-out ablations alone cannot separate additivity from redundancy, because they measure only the marginal effect of removing a component from the full configuration; the factorial resolves this.


The components are neither purely additive nor redundant. DER acts as a gate: with DER off, all four cells sit between $-1.06$ and $-0.54$ regardless of what CVaR and Thompson do, whereas with DER on those same two knobs move the score from $-0.43$ to $+1.13$. CVaR's marginal value is $+0.20$ with DER and Thompson both off, $+0.26$ with Thompson on but DER off, and $+0.59$ with DER on; Thompson shows the same pattern. No component's marginal value falls to zero given the others, which rules out redundancy. This is the operational sense of \emph{complementary} used in Sec.~\ref{sec:results-method}: each component's marginal value depends on which others are active, and that dependence is positive. The pattern is consistent with the distribution-shift analysis of Sec.~\ref{sec:analysis} --- CVaR and Thompson concentrate effort on the upper tail of whatever region the generator currently occupies, so if the policy has not been pushed off the prior's dominant modes there is little tail to concentrate on, and DER is what opens that headroom.

\paragraph{Comparison against the untreated (all-knobs-off) baseline.} Table~\ref{tab:allknobs} compares the full harness against the untreated online DDPP-LB configuration defined in Sec.~\ref{sec:exp-setup}, at a matched budget of $4{,}000$ oracle calls.

\begin{table}[H]
\centering
\small
\begin{tabular}{@{}lrrrrr@{}}
\toprule
 & \textbf{Best} & \textbf{Top-10} & \textbf{Top-100} & \textbf{Median} & \textbf{Mean of all} \\
\midrule
All five knobs on  & $+1.15$ & $+0.89$ & $+0.46$ & $-0.96$ & $-1.09$ \\
All five knobs off & $-0.49$ & $-0.73$ & $-1.55$ & $-2.65$ & $-2.66$ \\
\bottomrule
\end{tabular}
\caption{\textbf{Full harness vs.\ the untreated all-knobs-off baseline (DDPP-LB).} Averaged over $3$ targets at $4{,}000$ oracle calls. The measured top-$100$ gap of $+2.01$ is roughly $7\times$ the mean within-cell standard deviation.}
\label{tab:allknobs}
\end{table}

The gap is not a transient early advantage that more oracle calls would close: the untreated baseline never improves. Averaging the score of newly sampled molecules over successive blocks of $1{,}000$ calls gives $-2.68$, $-2.67$, $-2.65$, $-2.65$ for the all-off configuration against $-1.40$, $-0.94$, $-1.00$, $-1.06$ for the full method, so the all-off block mean is flat across the whole run while the full method rises. On 2VT4 specifically the same per-block comparison gives $-1.21$, $-1.23$, $-1.22$, $-1.22$ for all-off against $+0.10$, $+0.32$, $+0.25$, $+0.22$ for the full configuration.

\begin{figure}[ht]
    \centering
    \includegraphics[width=\linewidth]{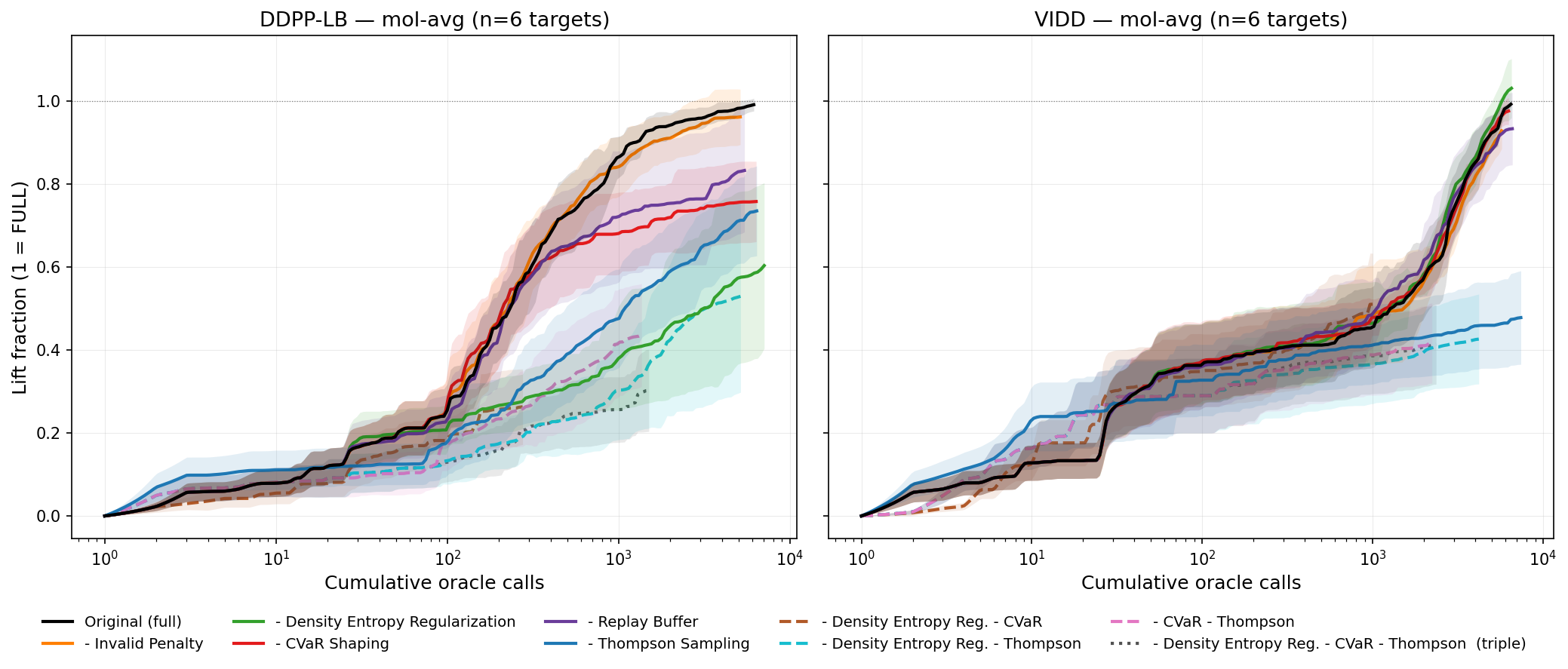}
    \caption{\textbf{Multi-knob ablation: normalised-reward trajectories.} Normalised reward ($1=$ FULL final) vs.\ cumulative oracle calls, averaged across the six FA-oracle small-molecule targets and five seeds; pair removals dashed, triple removal dotted. Removing additional exploration / shaping knobs degrades lift roughly monotonically the triple sits lowest at every budget on DDPP-LB, with VIDD showing the same ordering at compressed spread.}
    \label{fig:multiknob_lift_curves}
\end{figure}

\begin{figure}[ht]
    \centering
    \includegraphics[width=\linewidth]{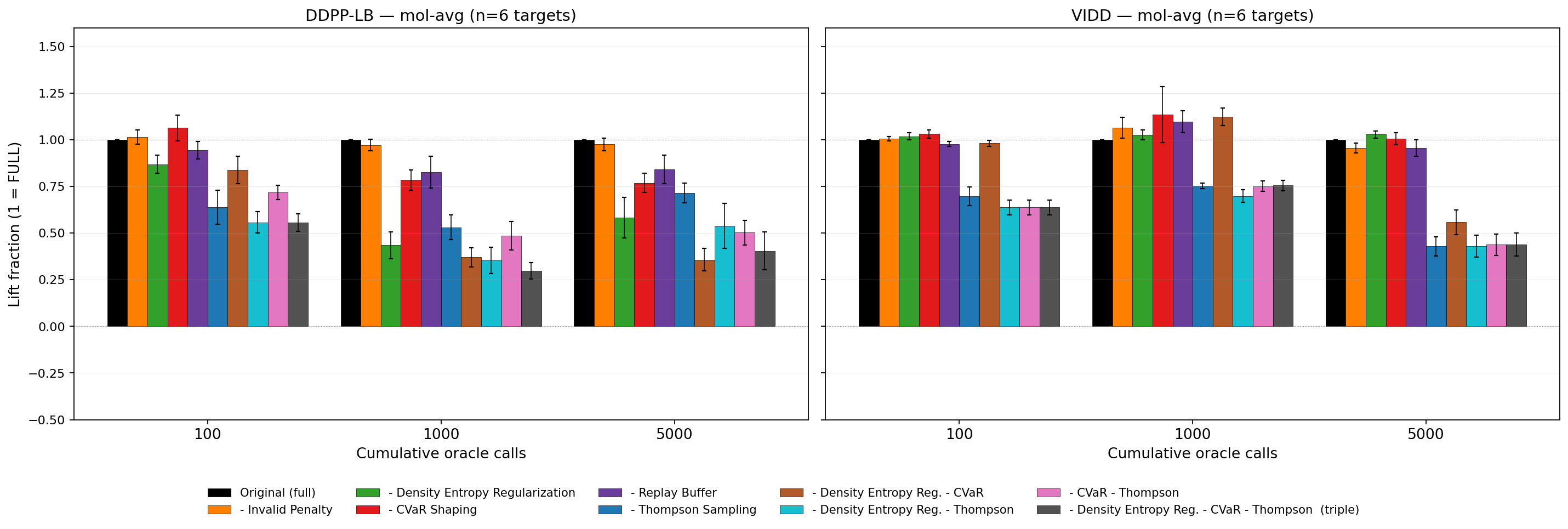}
    \caption{\textbf{Multi-knob ablation: normalised reward at oracle-call slices.} Same normalization as Fig.~\ref{fig:barplot_ablation}; six FA-oracle targets averaged, error bars $\pm 1$ SEM across targets. No pair removal sits above either of its constituent single removals at the 5\,000-call slice joint losses compound rather than substitute.}
    \label{fig:multiknob_lift_bars}
\end{figure}

\begin{figure}[ht]
    \centering
    \includegraphics[width=\linewidth]{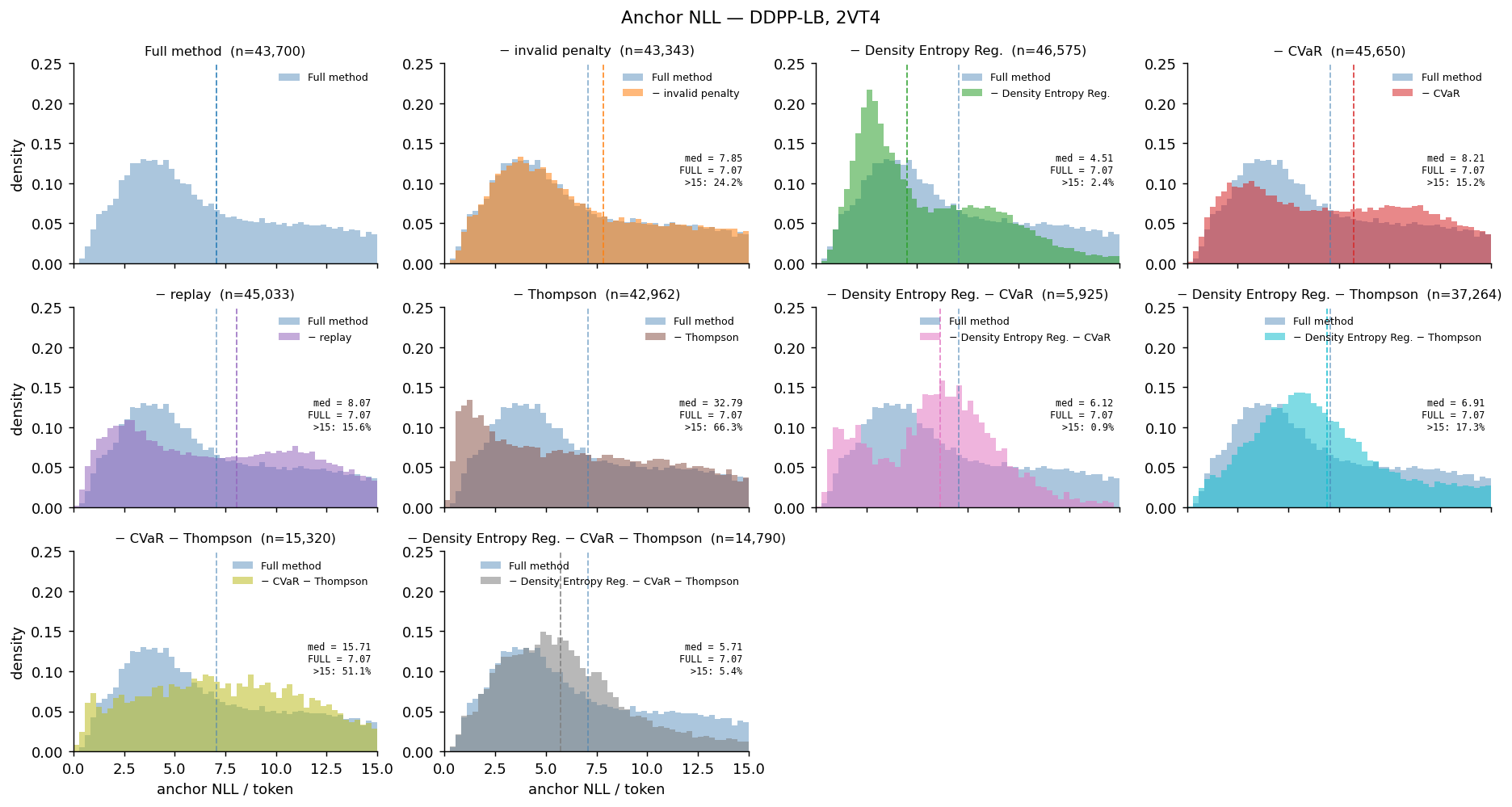}
    \caption{\textbf{Anchor-NLL distribution per multi-knob ablation cell (DDPP-LB, 2VT4).} Per-token anchor NLL $-\log p_{\text{pre}}(x_0 \mid x_t)$ under the frozen pretrained generator on all oracle-evaluated SMILES (5 seeds), rough estimator with $T=10$ random-mask MC averaging; FULL (steelblue) overlaid as reference. Every cell with Density Entropy Regularization removed collapses mass back toward the prior, identifying DE as the dominant driver of distribution shift away from $p_{\text{pre}}$ regardless of the other knobs.}
    \label{fig:multiknob_anchor_nll}
\end{figure}

\clearpage

\subsection{Cross-Finetuner Ablations: DDPP-LB and VIDD}\label{app:vidd-ablation}

\subsubsection{Numerical summary across tasks}

\begin{table}[t]
\centering
\footnotesize
\setlength{\tabcolsep}{4.5pt}
\caption{\textbf{Ablation normalised reward (mean $\pm$ s.e.).} Each row removes one design choice from the full method. Values are averaged across 6 molecular targets (Mol) and 3 protein families (Prot), normalised by the per-target range $[C_t, F_t]$ shared with the bar chart in Fig.~\ref{fig:barplot_ablation}.}
\label{tab:ablation_numbers}
\begin{tabular}{@{}lrcccc@{}}
\toprule
 & & \multicolumn{2}{c}{\textbf{DDPP}} & \multicolumn{2}{c}{\textbf{VIDD}} \\
\cmidrule(lr){3-4}\cmidrule(lr){5-6}
\textbf{Ablation} & \textbf{Calls} & \textbf{Mol} & \textbf{Prot} & \textbf{Mol} & \textbf{Prot} \\
\midrule
\multirow{3}{*}{Full method}
  & 100   & $.25{\pm}.02$ & $.72{\pm}.03$ & $.41{\pm}.03$ & $.66{\pm}.05$ \\
  & 1k    & $.85{\pm}.02$ & $.91{\pm}.01$ & $.49{\pm}.03$ & $.79{\pm}.04$ \\
  & 5k    & $.97{\pm}.01$ & $.94{\pm}.01$ & $.89{\pm}.02$ & $.87{\pm}.03$ \\
\midrule
\multirow{3}{*}{$-$ Invalid Pen.}
  & 100   & $.26{\pm}.02$ & $.72{\pm}.03$ & $.41{\pm}.03$ & $.66{\pm}.05$ \\
  & 1k    & $.83{\pm}.02$ & $.90{\pm}.01$ & $.51{\pm}.03$ & $.79{\pm}.04$ \\
  & 5k    & $.94{\pm}.02$ & $.95{\pm}.02$ & $.86{\pm}.03$ & $.87{\pm}.03$ \\
\midrule
\multirow{3}{*}{$-$ CVaR Shaping}
  & 100   & $.27{\pm}.02$ & $.72{\pm}.03$ & $.42{\pm}.02$ & $.67{\pm}.04$ \\
  & 1k    & $.67{\pm}.03$ & $.83{\pm}.03$ & $.51{\pm}.02$ & $.82{\pm}.03$ \\
  & 5k    & $.75{\pm}.03$ & $.85{\pm}.03$ & $.90{\pm}.03$ & $.90{\pm}.02$ \\
\midrule
\multirow{3}{*}{$-$ Density Entr.\ Reg.}
  & 100   & $.22{\pm}.01$ & $.77{\pm}.03$ & $.41{\pm}.02$ & $.66{\pm}.05$ \\
  & 1k    & $.39{\pm}.04$ & $.89{\pm}.01$ & $.49{\pm}.03$ & $.76{\pm}.04$ \\
  & 5k    & $.57{\pm}.04$ & $.97{\pm}.01$ & $.91{\pm}.02$ & $.84{\pm}.03$ \\
\midrule
\multirow{3}{*}{$-$ Replay Buffer}
  & 100   & $.24{\pm}.02$ & $.72{\pm}.03$ & $.41{\pm}.03$ & $.65{\pm}.05$ \\
  & 1k    & $.71{\pm}.03$ & $.83{\pm}.02$ & $.51{\pm}.03$ & $.85{\pm}.02$ \\
  & 5k    & $.82{\pm}.04$ & $.87{\pm}.02$ & $.85{\pm}.03$ & $.97{\pm}.01$ \\
\midrule
\multirow{3}{*}{$-$ Thompson Samp.}
  & 100   & $.18{\pm}.02$ & $.72{\pm}.03$ & $.31{\pm}.02$ & $.58{\pm}.04$ \\
  & 1k    & $.47{\pm}.03$ & $.78{\pm}.02$ & $.39{\pm}.02$ & $.66{\pm}.05$ \\
  & 5k    & $.70{\pm}.03$ & $.82{\pm}.01$ & $.44{\pm}.03$ & $.71{\pm}.04$ \\
\bottomrule
\end{tabular}
\end{table}

\subsubsection{Per-task VIDD trajectories}

\begin{figure}[H]
  \centering
  \includegraphics[width=0.95\textwidth]{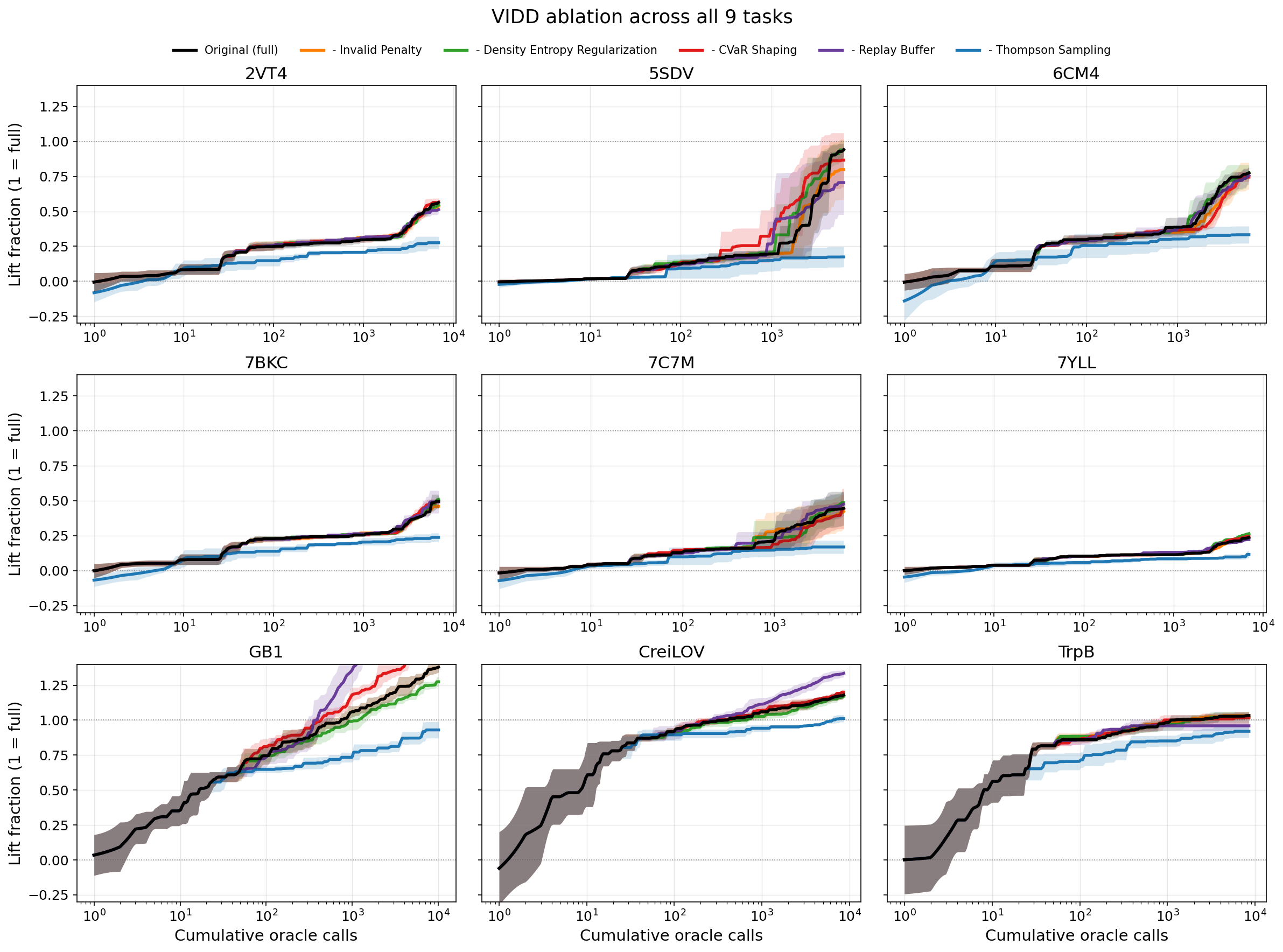}
  \caption{\textbf{VIDD harness ablation across all $9$ tasks.} Mirror of Fig.~\ref{fig:ablation_ddpp} with VIDD substituted for DDPP-LB as the finetuning objective $\mathcal{L}_{\text{ft}}$ in the harness; normalised reward is per-target $(1 =$ full VIDD$)$. Knob ordering matches DDPP-LB on most targets, supporting the harness claim of Sec.~\ref{sec:design-space} that knob intent transfers across finetuners. Spread between ablation cells is consistently smaller under VIDD: the student-teacher distillation step damps the policy drift that the exploration / objective-mismatch knobs amplify under DDPP-LB, so each knob's marginal contribution is also damped.}
  \label{fig:ablation_vidd}
\end{figure}

\clearpage

\section{Search Baselines and Compute Accounting}\label{app:search}

This section separates comparisons of optimization strategies from their computational cost. Each performance comparison reports both cumulative oracle calls and GPU hours.

\subsection{Online adaptation and search baselines}

\begin{figure}[H]
  \centering
  \includegraphics[width=\textwidth]{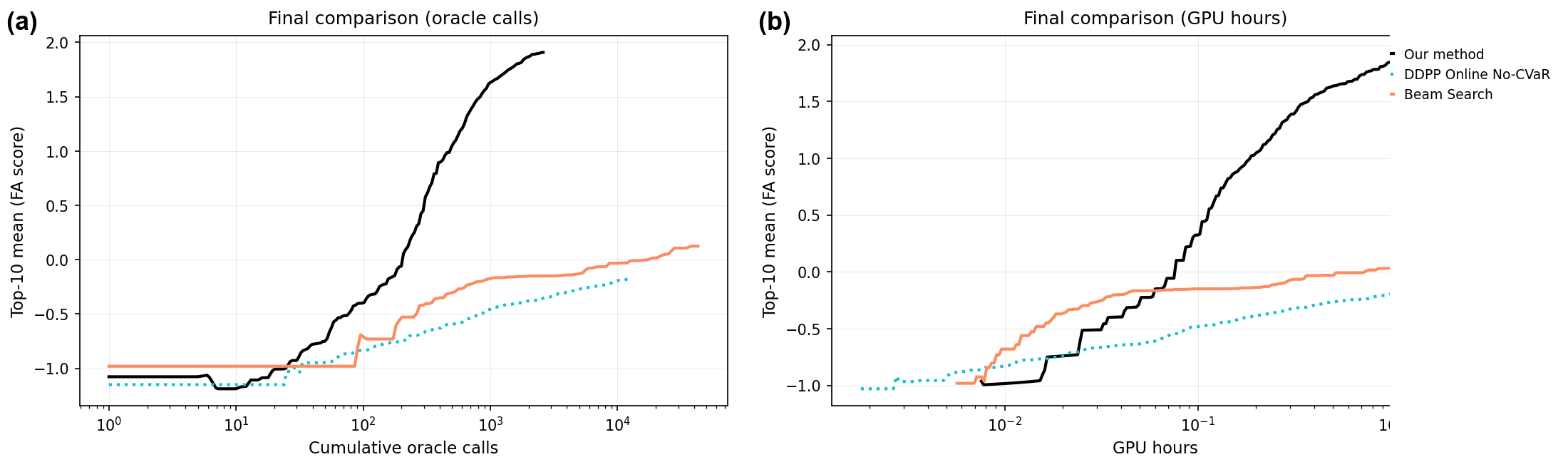}
  \caption{\textbf{Final comparison.} Top-10 mean FA score vs.\ (a) cumulative oracle calls and (b) GPU hours, comparing the full online active loop (Our method: DDPP-LB $+$ CVaR $+$ debiasing $+$ Thompson $+$ buffer $+$ invalid penalty) against the strongest pure-search baseline (Beam Search) and untreated online finetuning (DDPP Online No-CVaR). The full stack dominates both baselines under matched compute on the FA oracle.}
  \label{fig:final_comparison}
\end{figure}

\begin{figure}[H]
  \centering
  \includegraphics[width=0.49\textwidth]{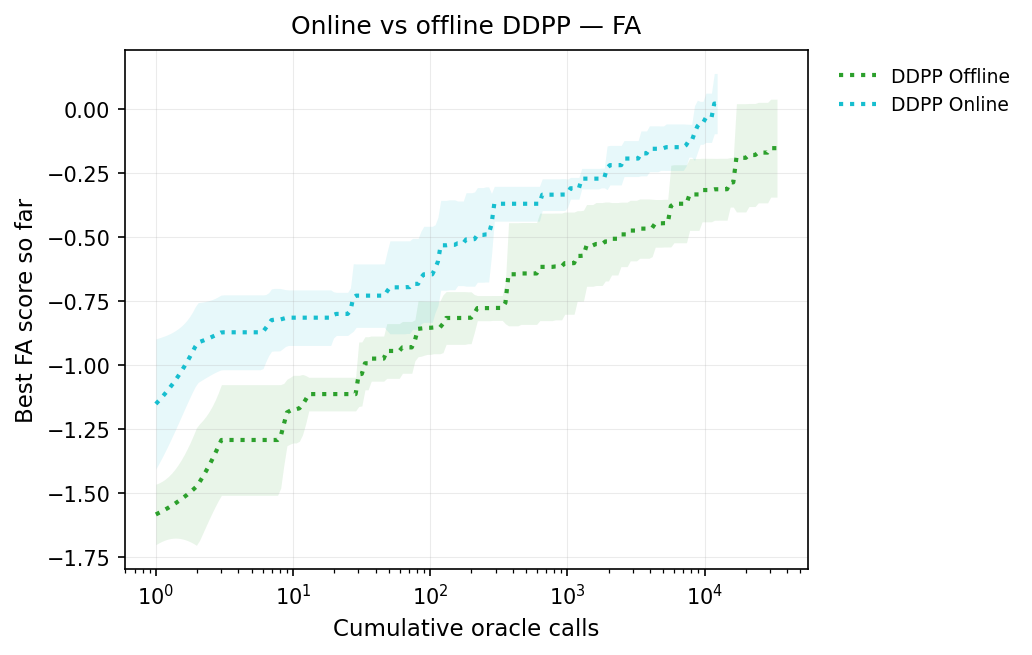}\hfill
  \includegraphics[width=0.49\textwidth]{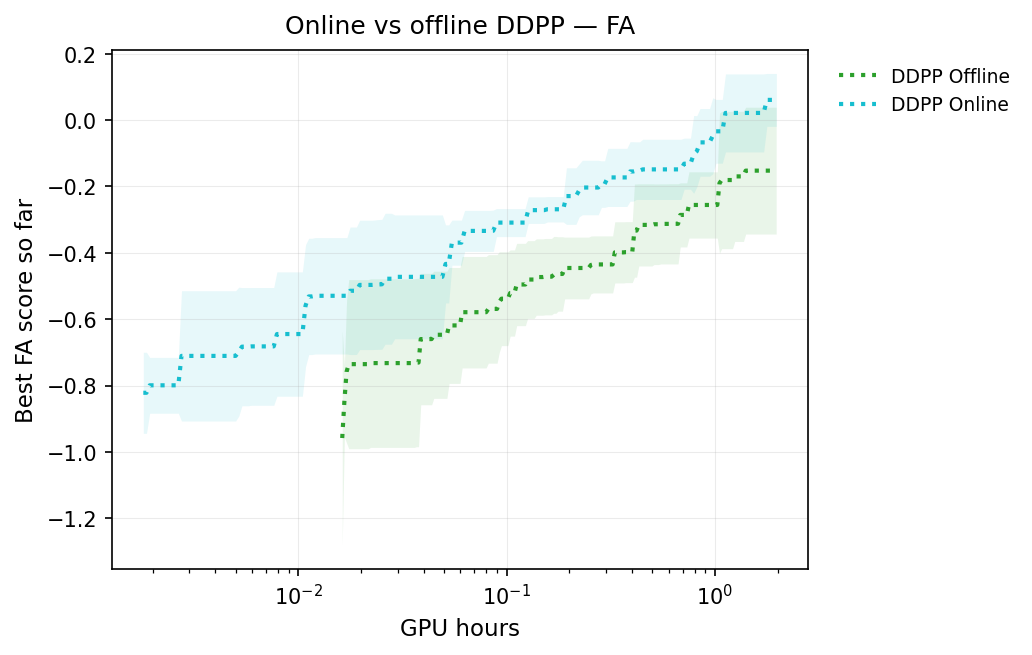}
  \caption{\textbf{Online vs.\ offline DDPP-LB} on top-1 best FA score so far, against cumulative oracle calls (left) and GPU hours (right). Both runs use the same DDPP-LB loss and no reward shaping; the only difference is whether the buffer of $(x, r(x))$ pairs is collected up front and trained on once (offline) or refreshed continuously by the current $q_\theta$ between gradient steps (online). Seed-averaged ($n{=}5$) bands. Online dominates on both axes: the larger number of gradient updates and the larger volume of policy-correlated data pushed through them push the sampling distribution closer to $p^\star$ than a one-shot offline schedule does.}
  \label{fig:online_vs_offline}
\end{figure}

\subsection{Surrogate screening and Tweedie shortcuts in frozen-generator search}

\begin{figure}[H]
  \centering
  \includegraphics[width=0.85\textwidth]{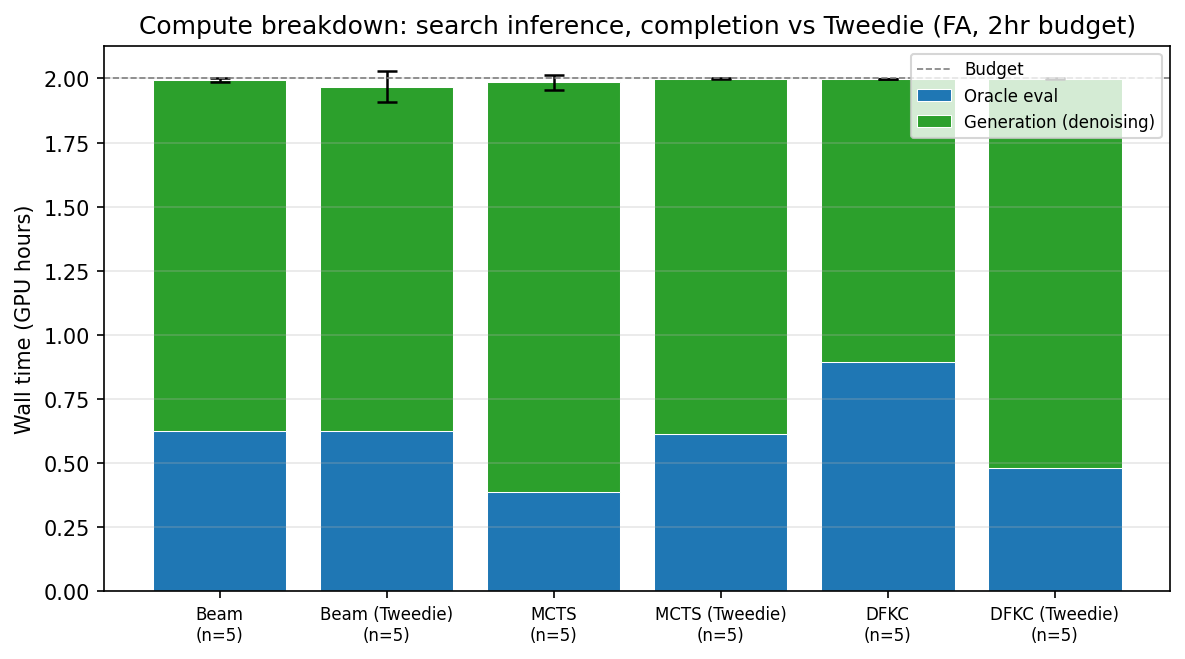}
  \caption{\textbf{Compute breakdown for inference-time search, completion vs.\ Tweedie.} Stacked bars are seed-averaged ($n{=}5$) wall-clock spend at the $2$-hour FA budget; oracle compute is estimated as $n_{\text{calls}} \times c_{\text{FA}}$ with $c_{\text{FA}} \approx 50$\,ms (matched to the per-phase log on the active runs). Tweedie shifts the per-family balance but does not recover enough wall time to compensate for the score loss in Fig.~\ref{fig:tweedie_search}: for MCTS it produces more oracle calls in the same budget, for DFKC it produces fewer. Generation dominates total wall time in every cell.}
  \label{fig:tweedie_inference_compute}
\end{figure}

\subsection{Search inside the online adaptation loop}

\begin{figure}[H]
  \centering
  \includegraphics[width=0.49\textwidth]{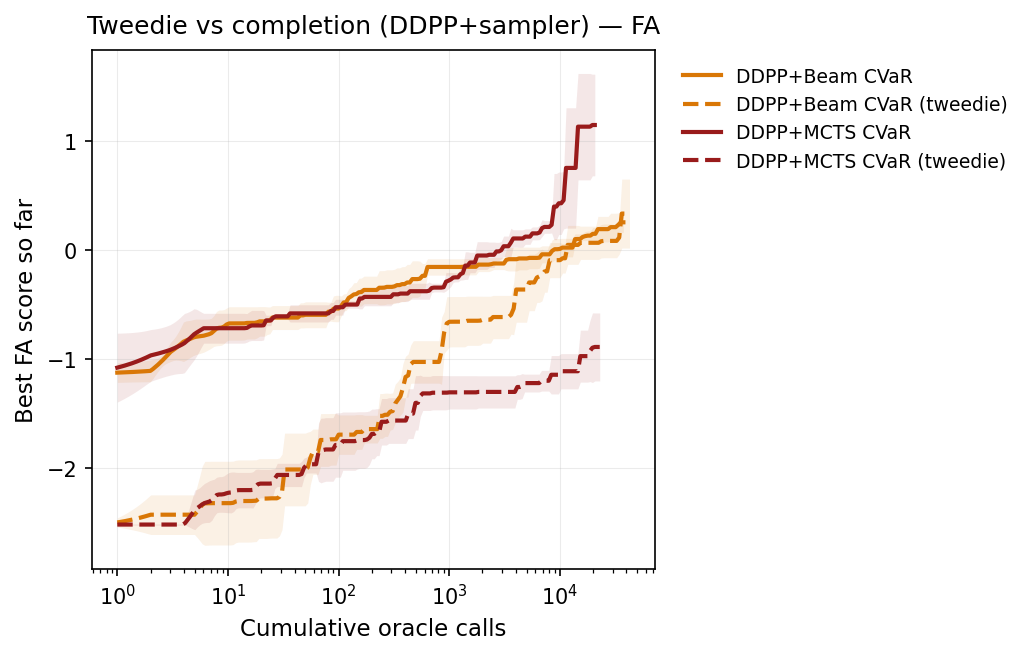}\hfill
  \includegraphics[width=0.49\textwidth]{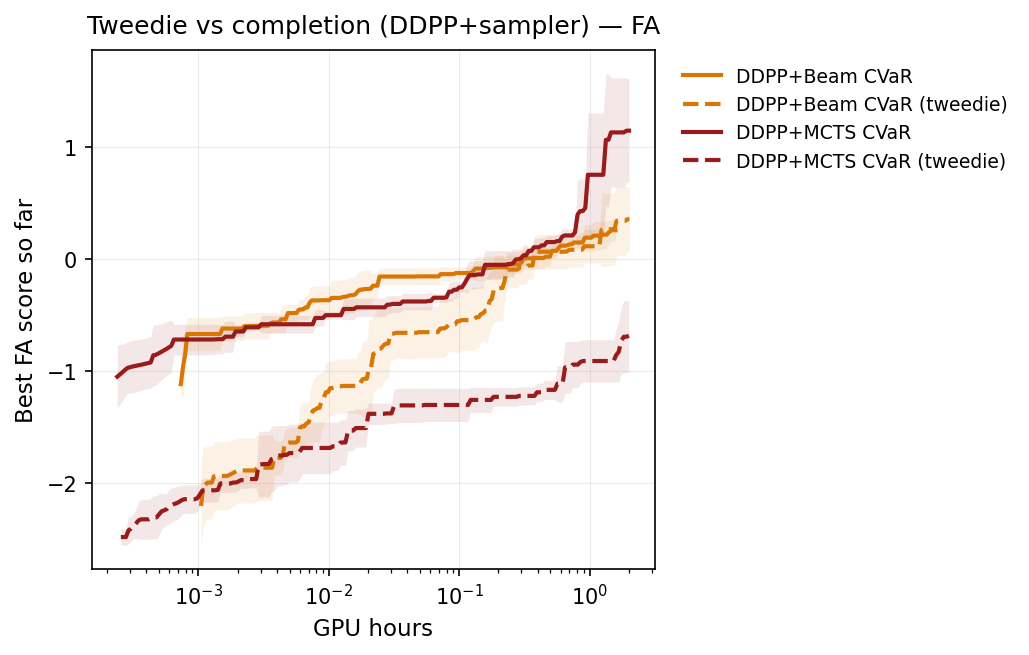}
  \caption{\textbf{Tweedie vs.\ completion (active hybrids).} Same Tweedie/completion comparison but applied inside the online DDPP-LB hybrids of Sec.~\ref{sec:results} (beam and MCTS only; DFKC is not paired here). Solid: completion. Dashed: Tweedie. Seed-averaged ($n{=}5$). The Tweedie variants underperform their completion counterparts here as well, despite the saved generation passes nominally freeing time for fine-tuning.}
  \label{fig:tweedie_active}
\end{figure}

\begin{figure}[H]
  \centering
  \includegraphics[width=0.85\textwidth]{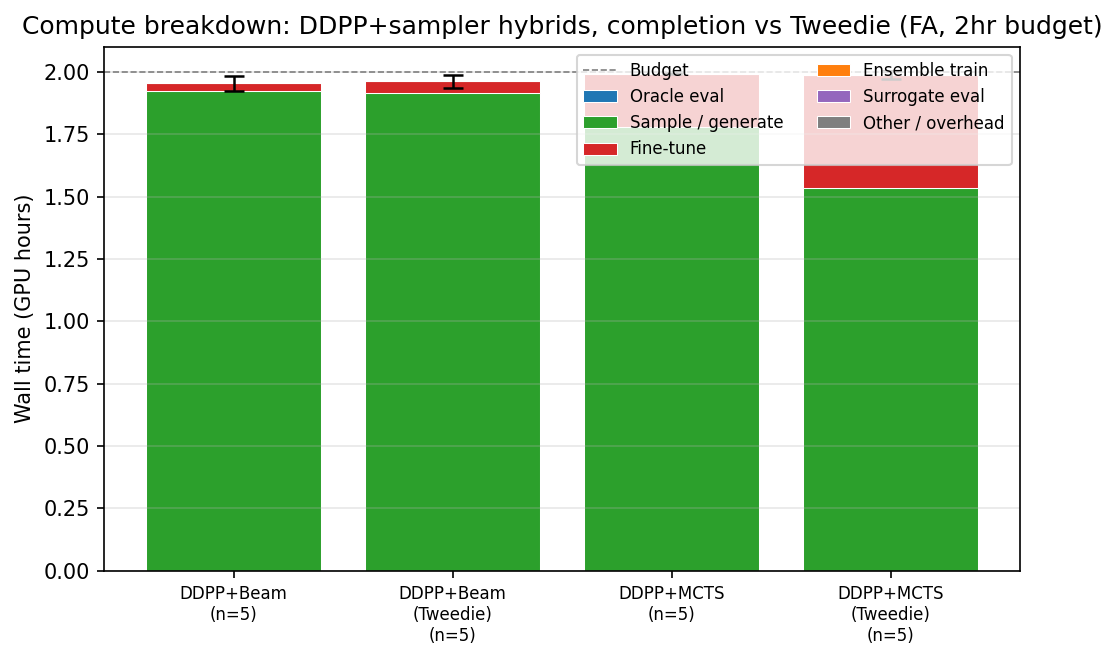}
  \caption{\textbf{Compute breakdown for active hybrids, completion vs.\ Tweedie.} Per-phase wall times pulled from \texttt{active\_loop\_log.jsonl}; seed-averaged ($n{=}5$). Generation dominates for both completion and Tweedie variants, leaving very little wall time for fine-tuning ($\leq 0.1$\,hr) regardless of whether the search uses the Tweedie shortcut or completion. The fine-tuning budget reduction observed for completion-based hybrids in Fig.~\ref{fig:search_overhead} is not relieved by the Tweedie shortcut.}
  \label{fig:tweedie_active_compute}
\end{figure}

\subsection{Compute accounting and oracle-cost sensitivity}

\begin{figure}[H]
  \centering
  \includegraphics[width=\textwidth]{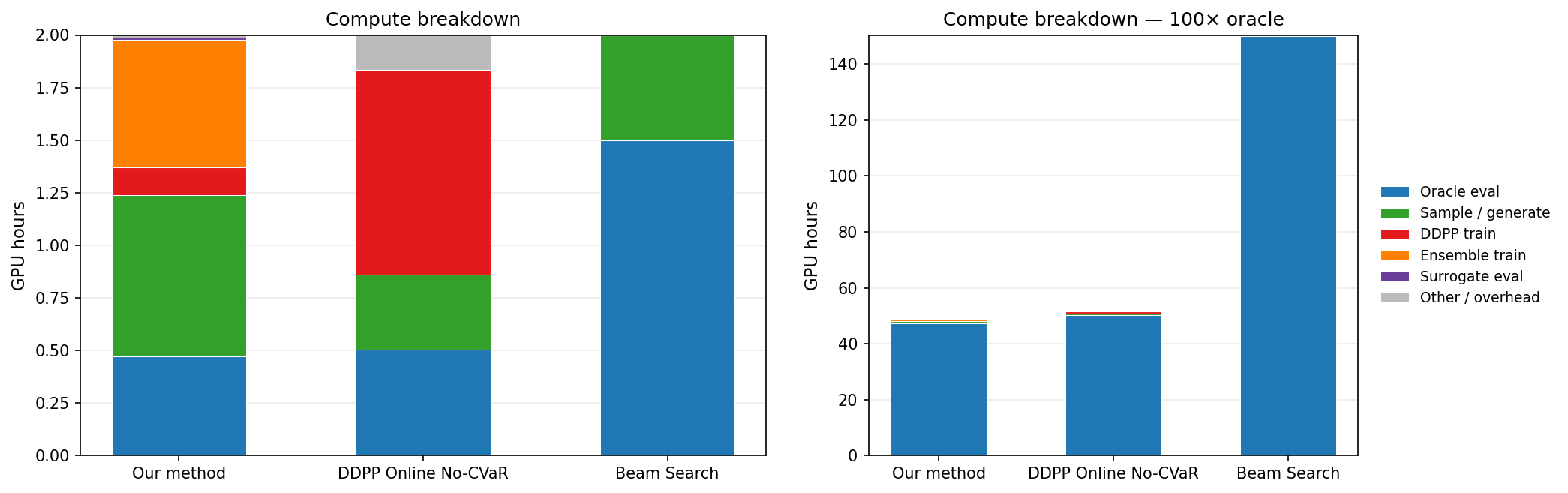}
  \caption{\textbf{Compute breakdown.} Per-method decomposition of wall-clock spend at standard FA oracle cost (left, $1\times$) and under a retroactive $100\times$ oracle-cost rescaling (right). Pure search spends almost all of its budget on oracle calls, so its absolute compute scales nearly linearly with $c_{\text{oracle}}$. Online finetuning amortizes generation and gradient updates across the same oracle budget, so its share of oracle compute is smaller and the absolute compute under the $100\times$ rescaling grows much less.}
  \label{fig:compute_breakdown}
\end{figure}

To pull apart why test-time finetuning beats inference-time search, it helps to look at where each method actually spends its wall-clock budget. We log per-call elapsed time alongside the oracle score in every run (Sec.~\ref{sec:exp-setup}), which lets us decompose each method's wall time into the components of Eq.~\ref{eq:wall-decomp}: oracle evaluation ($N \cdot c_{\text{oracle}}$), candidate generation, fine-tuning gradient steps, surrogate ensemble training, and bookkeeping. The same per-call timeline supports a retroactive cost analysis: scaling the per-call oracle cost $c_{\text{oracle}}$ by $100\times$ simulates a switch from a fast oracle (FA) to a slow oracle (Boltz-2-class) without re-running anything. Appendix Fig.~\ref{fig:compute_breakdown} shows the resulting breakdowns side-by-side.

The takeaway is that finetuning's advantage over search comes mostly from not spending the budget on oracle calls. Each gradient step extracts more usable signal per oracle call than a one-shot search evaluation does, so the same oracle budget does more work. The advantage widens further as the oracle gets more expensive: the $100\times$ panel is closer to where Boltz-2-class oracles actually live, and there the full-loop compute breakdown is dominated by generation and surrogate training rather than oracle evaluation.

\begin{figure}[H]
  \centering
  \includegraphics[width=0.49\textwidth]{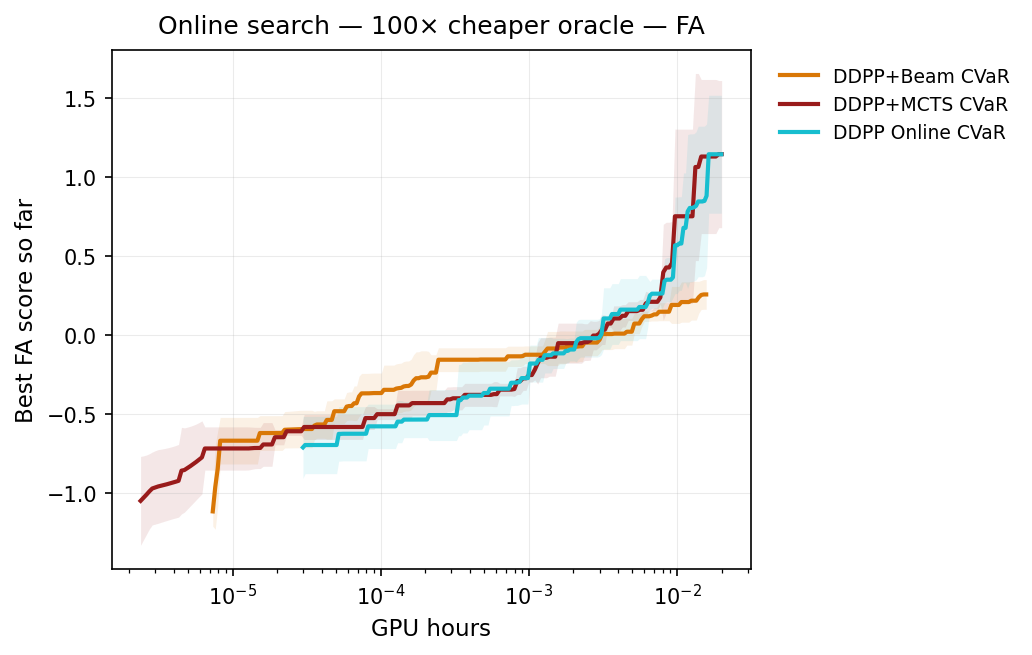}\hfill
  \includegraphics[width=0.49\textwidth]{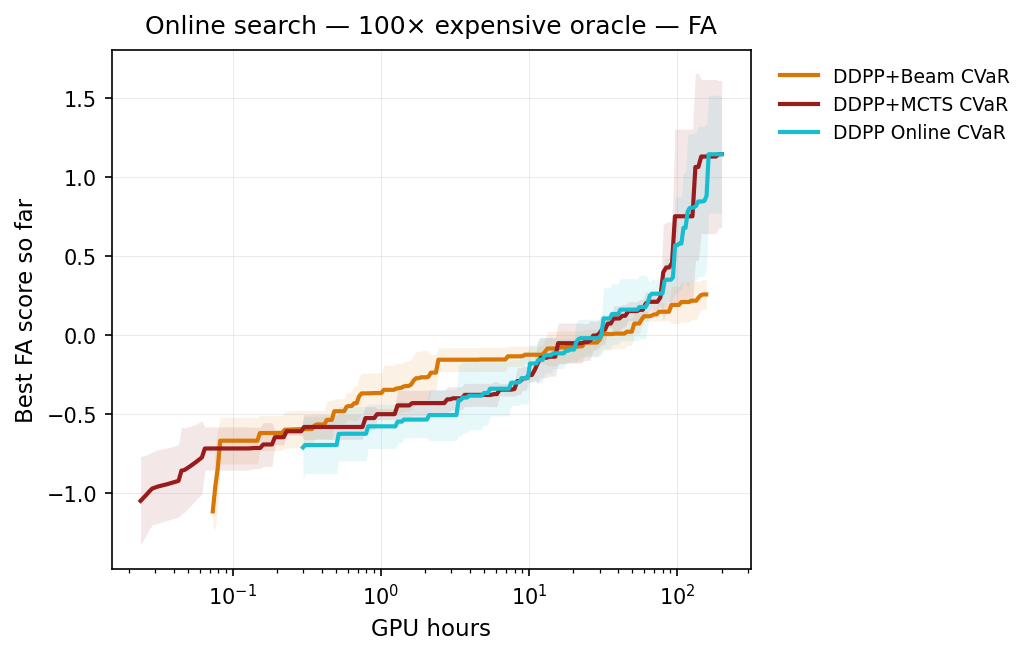}
  \caption{\textbf{Retroactive oracle-cost rescaling on online search hybrids.} Same three online methods (\texttt{ddpp\_online\_cvar} vs.\ \texttt{ddpp\_beam\_cvar} vs.\ \texttt{ddpp\_mcts\_cvar}) plotted on the GPU-hours axis with the per-call oracle cost rescaled by $1/100$ (left) and by $100$ (right). Rescaling shifts the GPU-hours axis by four orders of magnitude in either direction without altering the relative ordering: plain online DDPP-LB matches or exceeds both search-augmented variants under both regimes, supporting the claim in Sec.~\ref{sec:results} that the marginal value of inference-time search on top of online finetuning is robust to oracle expense.}
  \label{fig:oracle_rescale}
\end{figure}

\clearpage

\section{Task-Level and Aggregate Results}\label{app:results-detail}

This section first preserves target-specific variation and then aggregates the same comparisons across domains and compute axes.

\subsection{Per-Task Side-by-Side Ablation Grids}\label{app:per-task-ablation}

The main-text figures (Fig.~\ref{fig:ablation_ddpp}, Fig.~\ref{fig:ablation_search}, and Fig.~\ref{fig:ablation_vidd} in Appendix~\ref{app:vidd-ablation}) hold the finetuner / search-family fixed and step through the $9$ tasks. The figures below transpose that view: each $3 \times 3$ grid groups rows by task and places DDPP-LB, VIDD, and search-only baselines side by side per task. Same normalised-reward definition and conventions as Sec.~\ref{sec:exp-setup}.

\begin{figure}[H]
  \centering
  \includegraphics[width=0.95\textwidth]{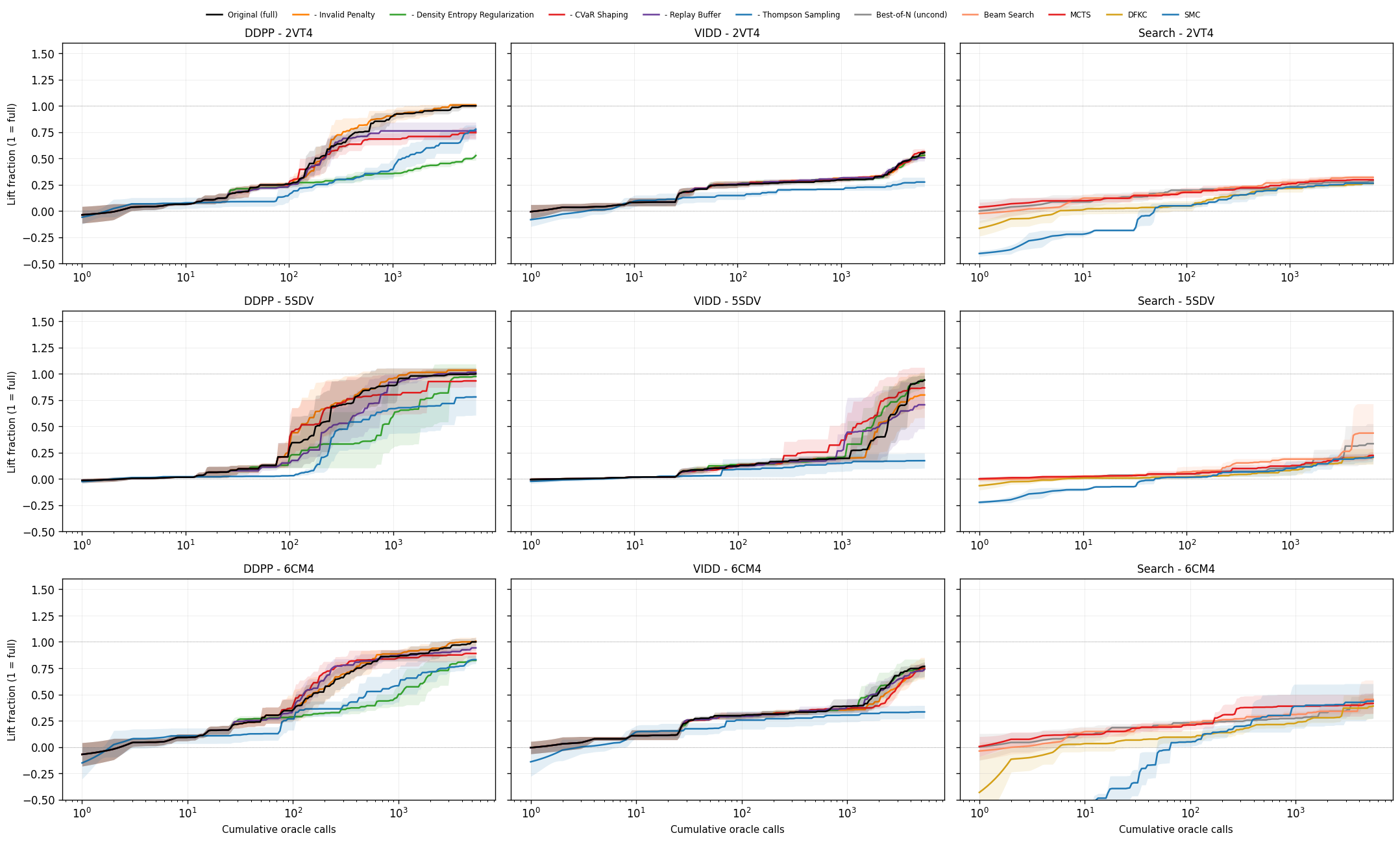}
  \caption{\textbf{Per-task ablation, small molecules (set 1: 2VT4, 5SDV, 6CM4).} Rows are FA-oracle targets; columns are DDPP-LB ablations (left), VIDD ablations (centre), search baselines (right). The VIDD column is consistently more compressed than DDPP-LB on the same target, matching the cross-finetuner observation in Fig.~\ref{fig:ablation_vidd}.}
  \label{fig:per_task_mol_set1}
\end{figure}

\begin{figure}[H]
  \centering
  \includegraphics[width=0.95\textwidth]{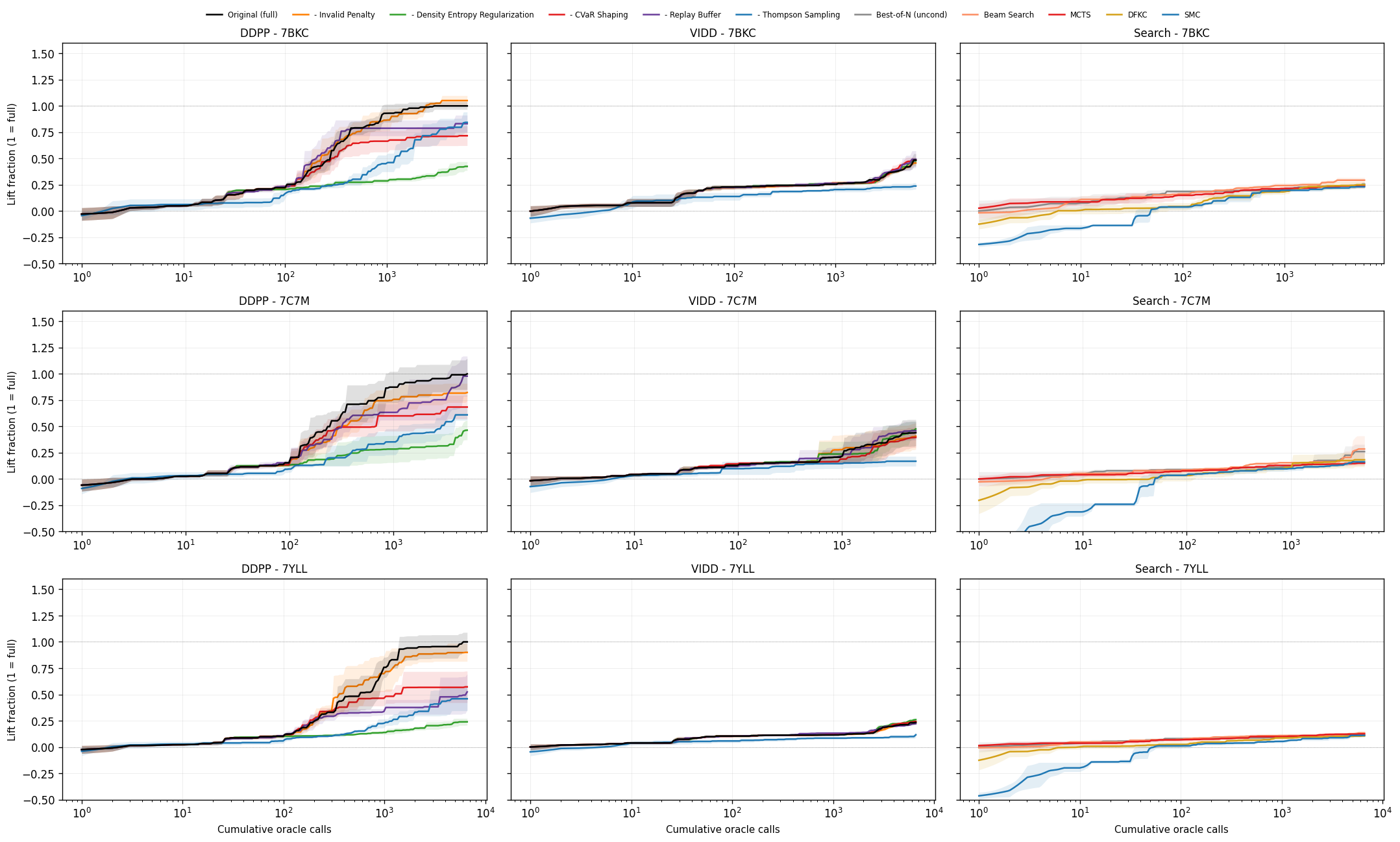}
  \caption{\textbf{Per-task ablation, small molecules (set 2: 7BKC, 7C7M, 7YLL).} Continuation of Fig.~\ref{fig:per_task_mol_set1} for the remaining three FA-oracle targets. The 7C7M row is extreme: removing \textit{- density entropy regularization} drops DDPP-LB to a small fraction of the full method, while the same removal under VIDD has a much smaller effect --- consistent with the student-teacher damping of policy drift discussed in Appendix~\ref{app:vidd-ablation}.}
  \label{fig:per_task_mol_set2}
\end{figure}

\begin{figure}[H]
  \centering
  \includegraphics[width=0.95\textwidth]{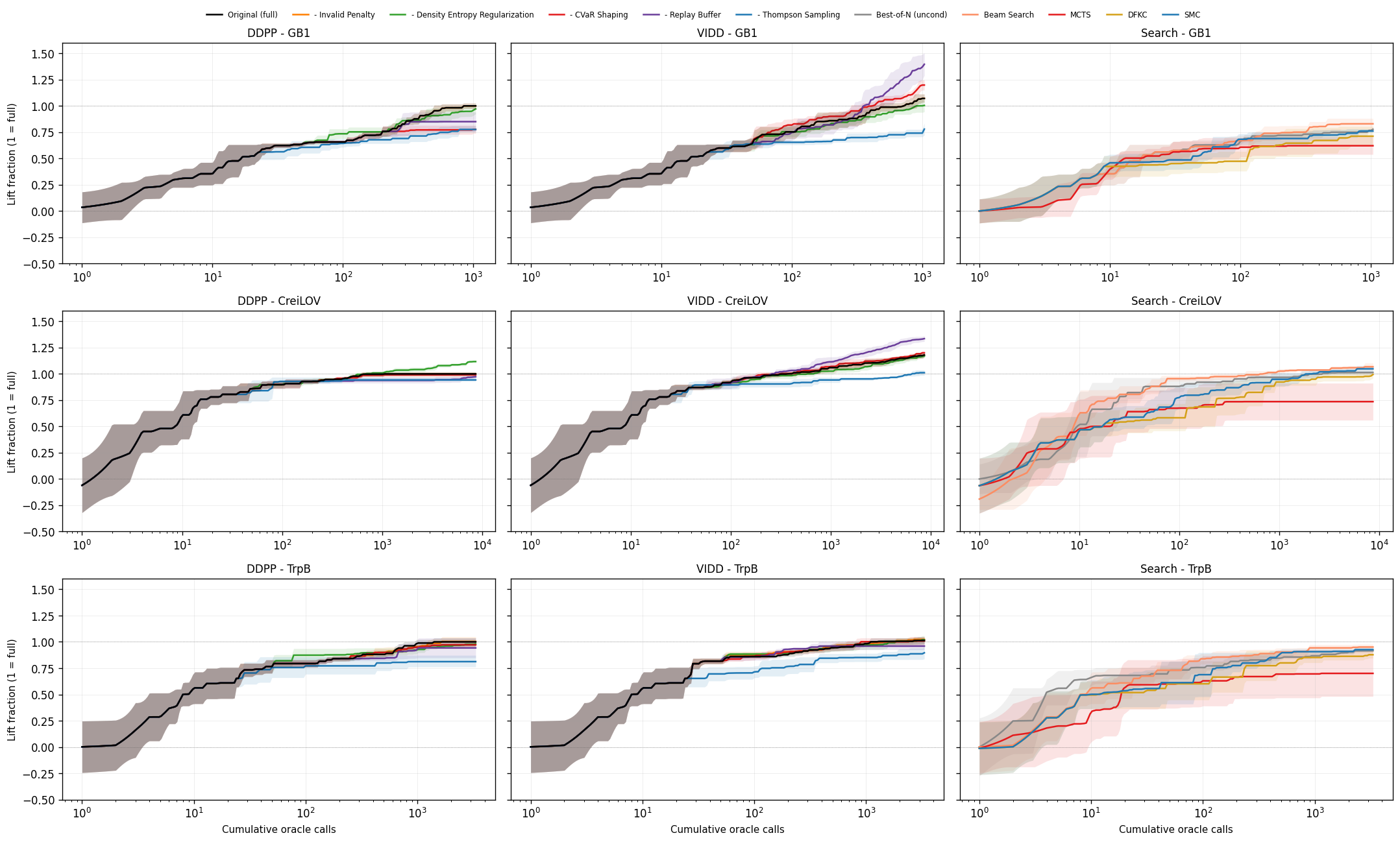}
  \caption{\textbf{Per-task ablation, protein-fitness targets (GB1, CreiLOV, TrpB).} Rows are protein-fitness families; columns match Fig.~\ref{fig:per_task_mol_set1}. CI bands are wider than the small-molecule rows because each family has a single trained surrogate / generator pair (no within-family pooling).}
  \label{fig:per_task_protein}
\end{figure}

\begin{figure}[H]
  \centering
  \includegraphics[width=\textwidth]{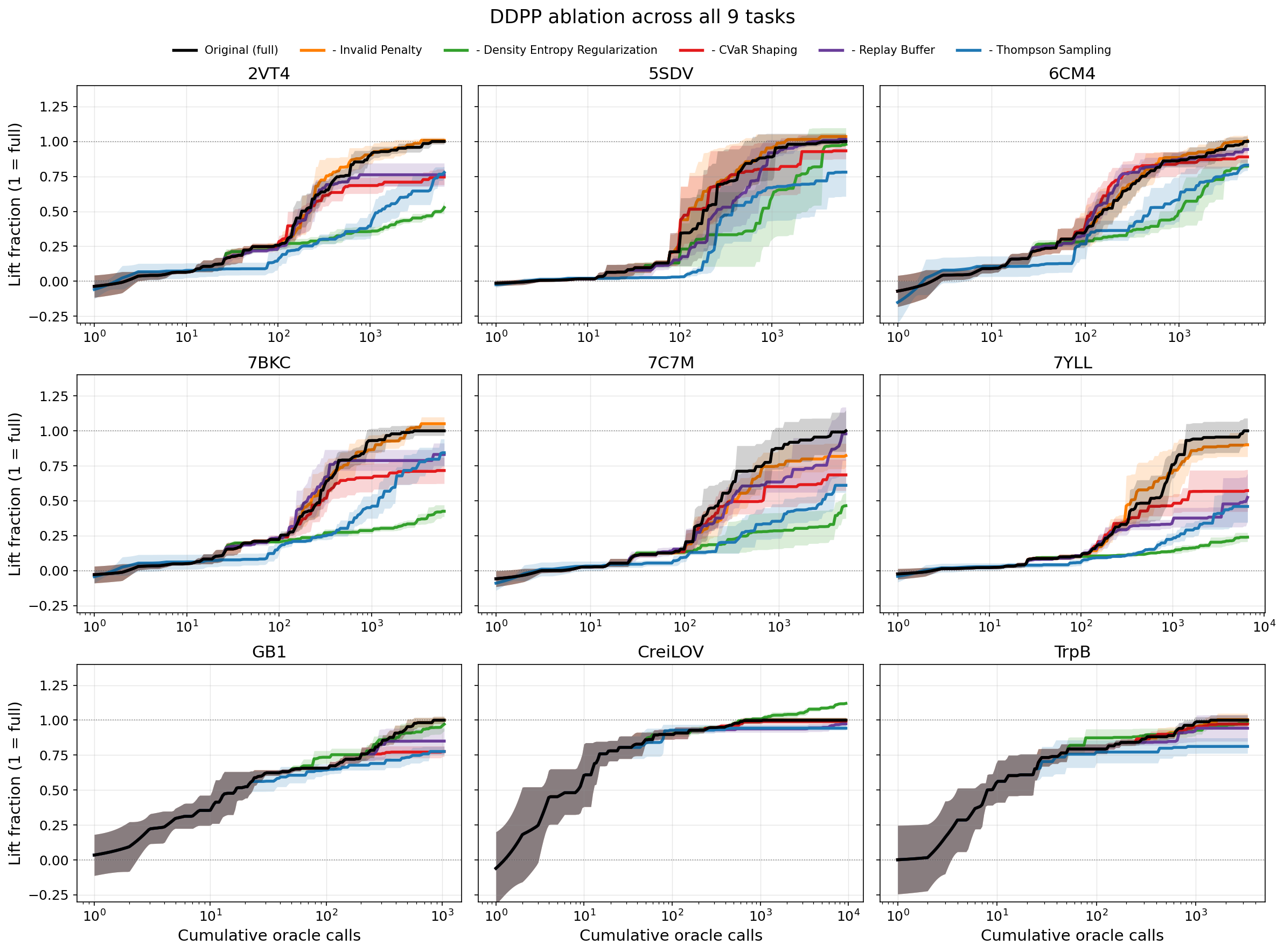}
  \caption{\textbf{DDPP-LB harness ablation across all $9$ tasks.} Normalised reward $(1 =$ full DDPP-LB on that target$)$ for each leave-one-out ablation of the knobs in Table~\ref{tab:methods}; six FA-oracle small molecules (top two rows) and three protein-fitness families (bottom row). Seeds, CI bands, and budget cap follow Sec.~\ref{sec:exp-setup}. Each knob's contribution varies across tasks, but knob ordering is broadly preserved across targets. VIDD twin: Appendix~\ref{app:vidd-ablation}. Per-task side-by-side: Appendix~\ref{app:per-task-ablation}.}
  \label{fig:ablation_ddpp}
\end{figure}

\begin{figure}[H]
  \centering
  \includegraphics[width=\textwidth]{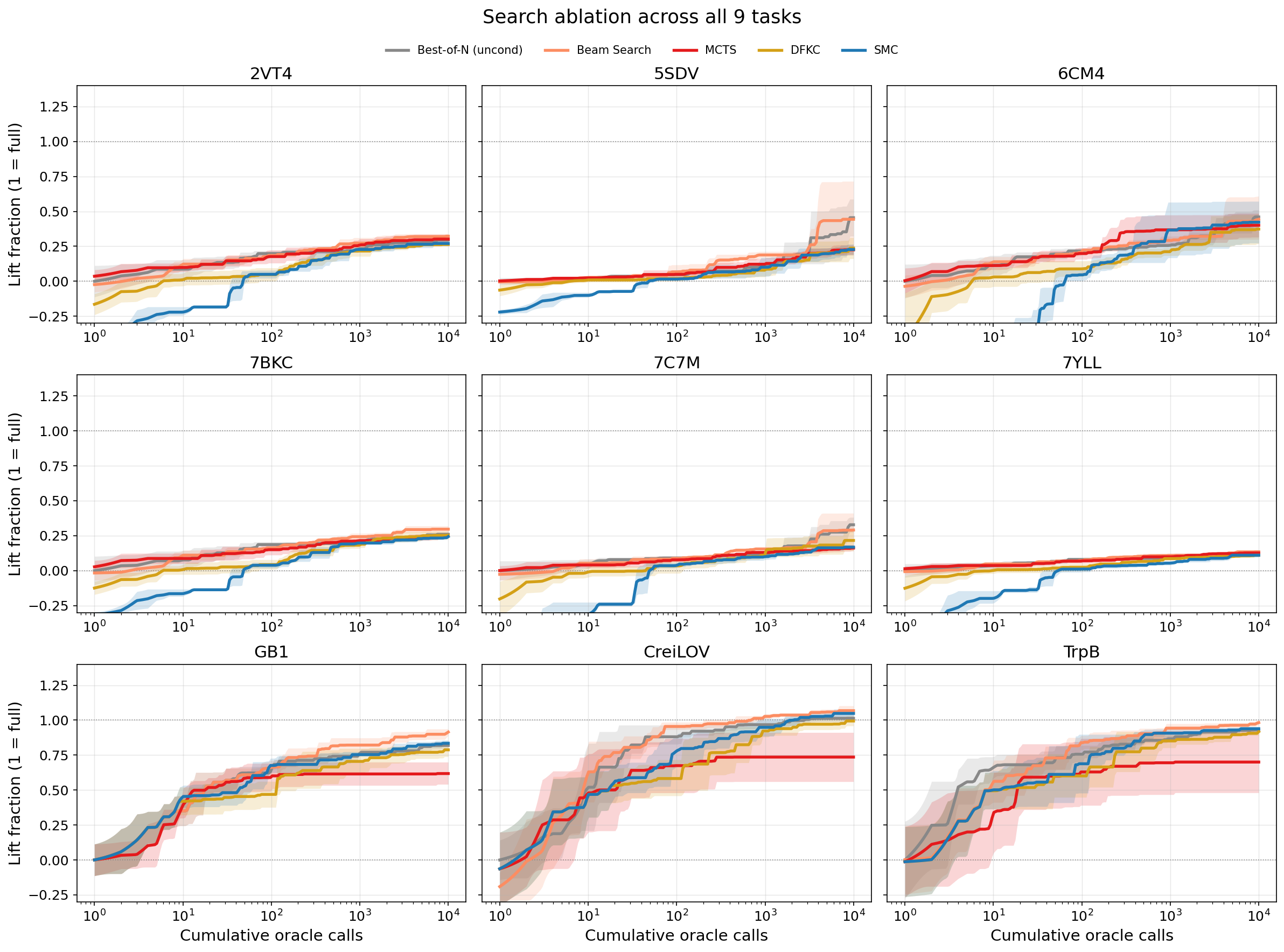}
  \caption{\textbf{Inference-time search-only baselines across all $9$ tasks.}} Same normalised-reward definition as Fig.~\ref{fig:ablation_ddpp} applied to the search-only methods (best-of-$N$, beam, MCTS, DFKC, SMC) at matched cumulative-oracle budget.
  \label{fig:ablation_search}
\end{figure}

\subsection{Aggregated Ablation Grids}\label{app:avg-ablation}

The per-task figures above retain the cross-target spread that motivates the manifold / mode-density discussion in Sec.~\ref{sec:results}. The figures below collapse that spread by averaging the normalised-reward curves across the $6$ FA-oracle small molecules (top row) and across the $3$ protein-fitness families (bottom row), giving a single curve per knob per axis-type. CI bands are across-target ($\pm 1.96 \cdot \mathrm{SEM}$), so they reflect target-level rather than seed-level variability.

\begin{figure}[H]
  \centering
  \includegraphics[width=0.95\textwidth]{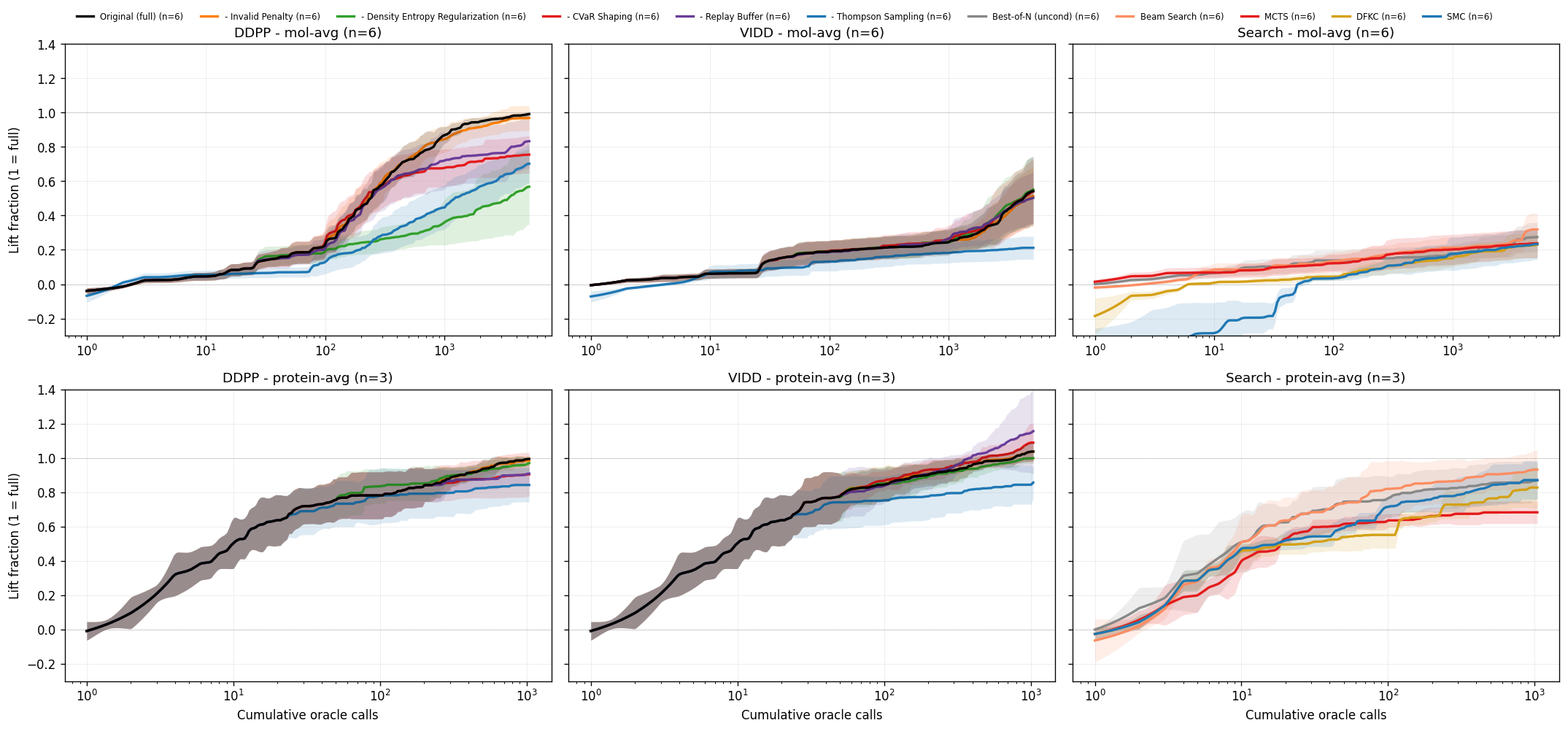}
  \caption{\textbf{Aggregated ablation grid, oracle-call view.} $2 \times 3$ grid: row $1$ averages over all six FA-oracle small molecules, row $2$ averages over the three protein-fitness families; columns are DDPP-LB ablations, VIDD ablations, and search-only baselines. Same normalised-reward definition as Fig.~\ref{fig:ablation_ddpp}. Across-target CI bands (target-level variability) are wider than the per-target seed bands in the figures above.}
  \label{fig:avg_ablation_calls}
\end{figure}

\begin{figure}[H]
  \centering
  \includegraphics[width=0.95\textwidth]{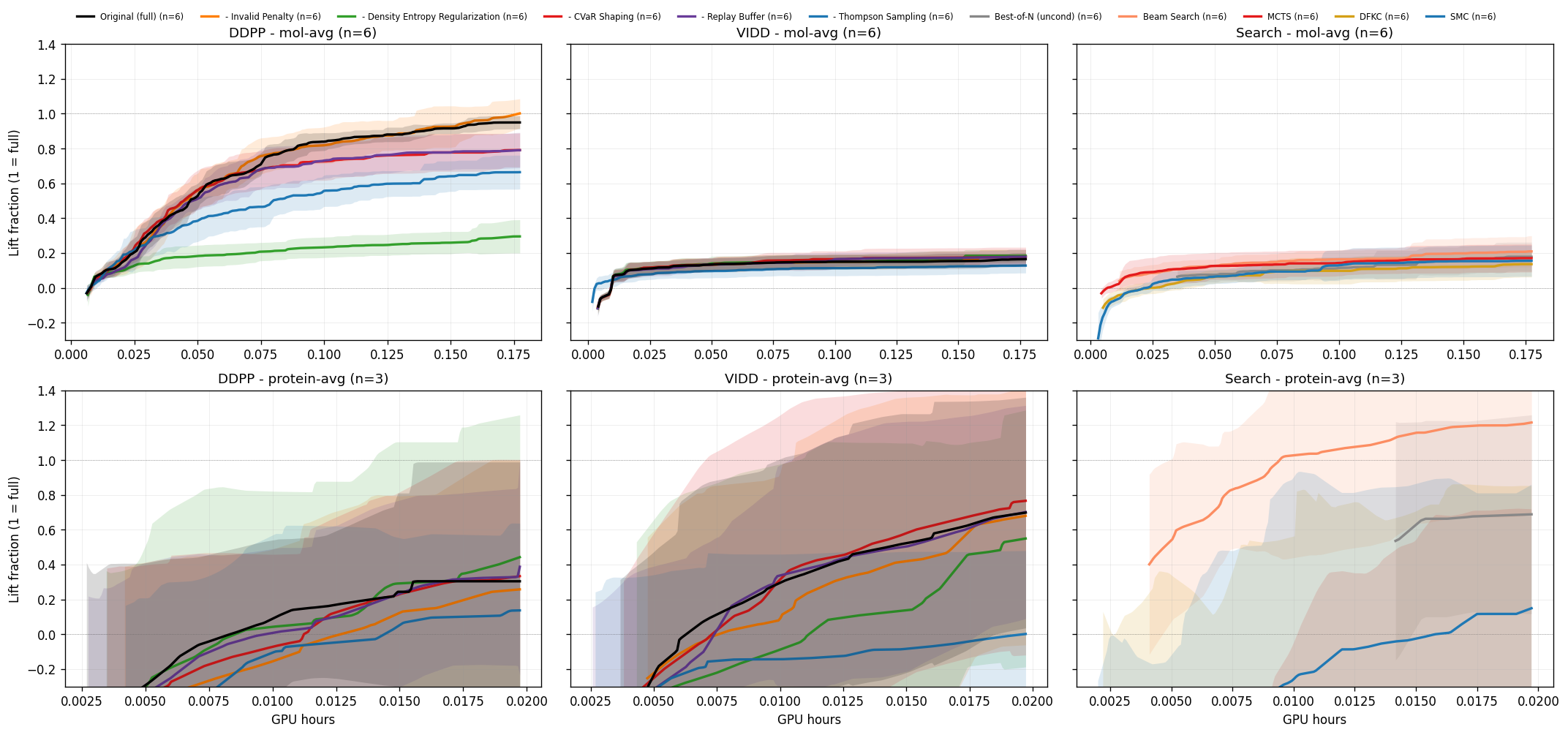}
  \caption{\textbf{Aggregated ablation grid, GPU-hours view.} GPU-hours twin of Fig.~\ref{fig:avg_ablation_calls}; same averaging.}
  \label{fig:avg_ablation_gpuhrs}
\end{figure}

\subsection{GPU-Hour View of the Ablation Grids}\label{app:gpuhrs-ablation}

The main-text and per-task ablation grids (Figs.~\ref{fig:ablation_ddpp}, \ref{fig:ablation_search}, \ref{fig:ablation_vidd}, \ref{fig:per_task_mol_set1}, \ref{fig:per_task_mol_set2}, \ref{fig:per_task_protein}) use cumulative oracle calls on the x-axis to isolate feedback efficiency. This appendix replots the same data against GPU hours, including method-side cost (generation, scoring, surrogate update, finetuning); see Sec.~\ref{sec:exp-setup} for the motivation. Same normalised-reward definition throughout.

\begin{figure}[H]
  \centering
  \includegraphics[width=0.95\textwidth]{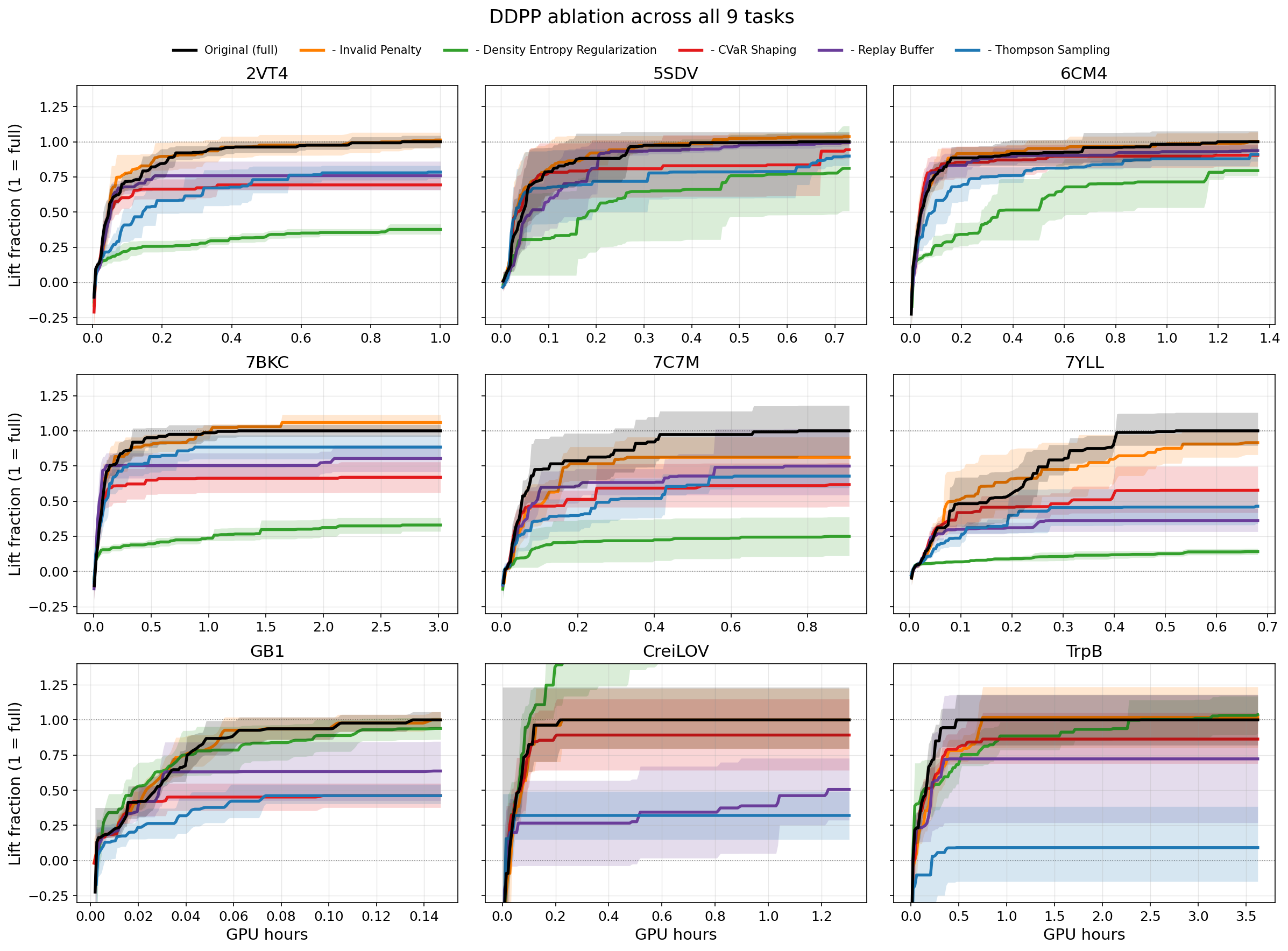}
  \caption{\textbf{DDPP-LB harness ablation, GPU-hours view.} GPU-hours twin of Fig.~\ref{fig:ablation_ddpp}. Knob ordering is largely preserved, but the gap between the full method and \textit{- Thompson sampling} narrows in wall time --- the Thompson step adds non-trivial method-side cost.}
  \label{fig:ablation_ddpp_gpuhrs}
\end{figure}

\begin{figure}[H]
  \centering
  \includegraphics[width=0.95\textwidth]{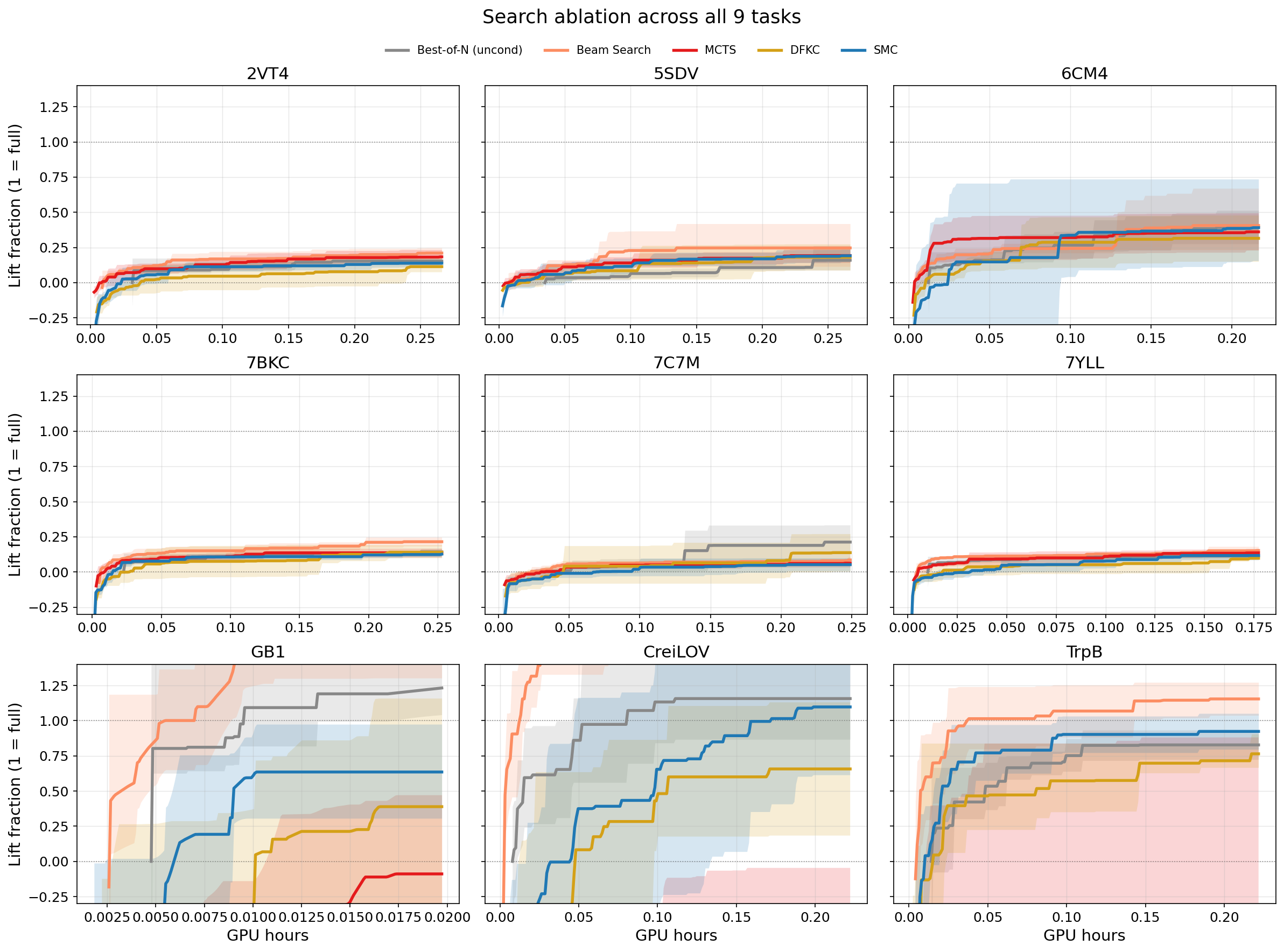}
  \caption{\textbf{Search-only baselines, GPU-hours view.} GPU-hours twin of Fig.~\ref{fig:ablation_search}. Plain best-of-$N$ wins on wall time more often than on oracle-call efficiency --- no per-call surrogate / search overhead.}
  \label{fig:ablation_search_gpuhrs}
\end{figure}

\begin{figure}[H]
  \centering
  \includegraphics[width=0.95\textwidth]{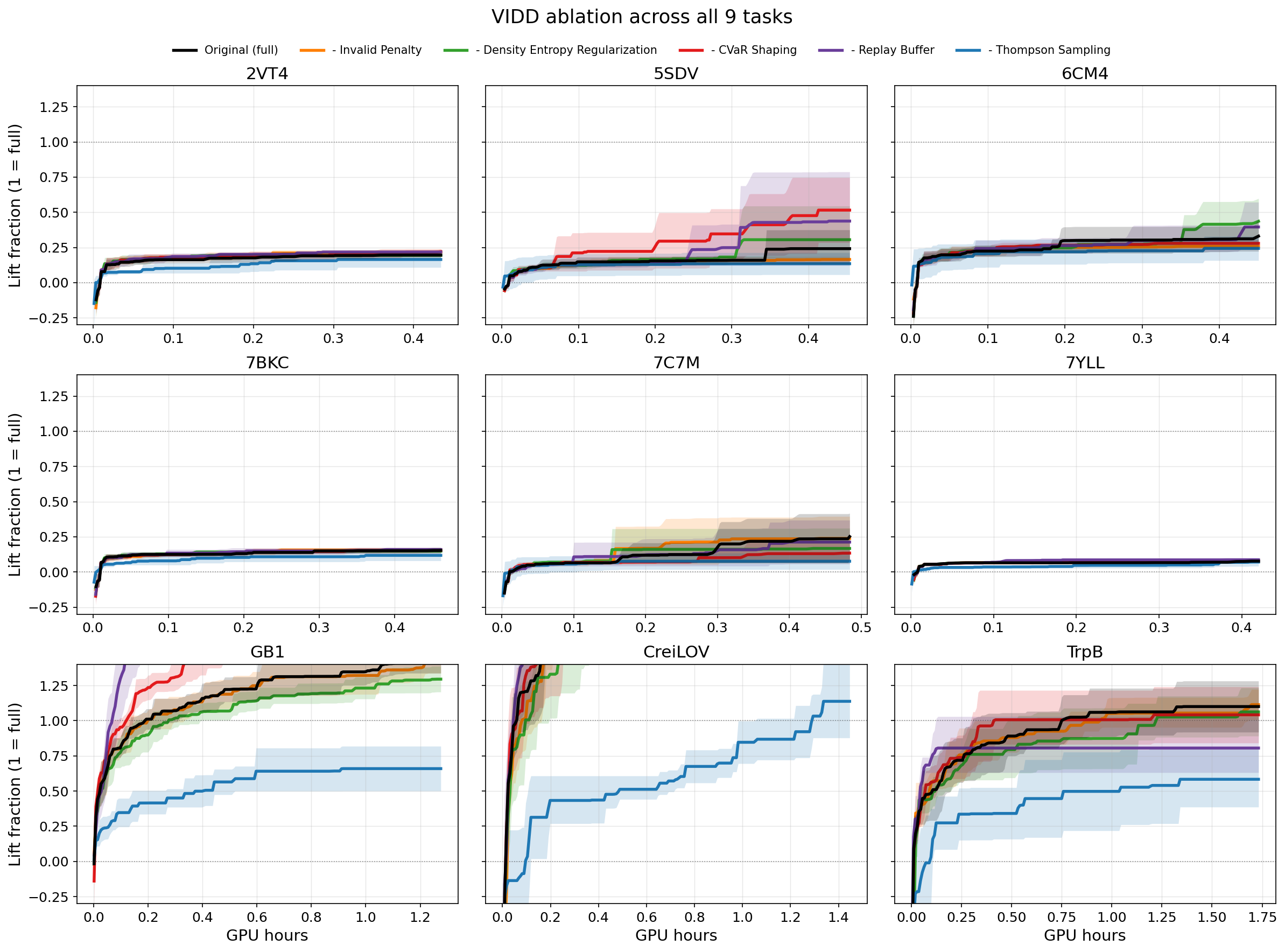}
  \caption{\textbf{VIDD harness ablation, GPU-hours view.} GPU-hours twin of Fig.~\ref{fig:ablation_vidd}. The distillation step adds a fixed per-round cost, so GPU-hour caps are tighter than the corresponding DDPP-LB caps on the same target; knob ordering is unchanged.}
  \label{fig:ablation_vidd_gpuhrs}
\end{figure}

\begin{figure}[H]
  \centering
  \includegraphics[width=0.95\textwidth]{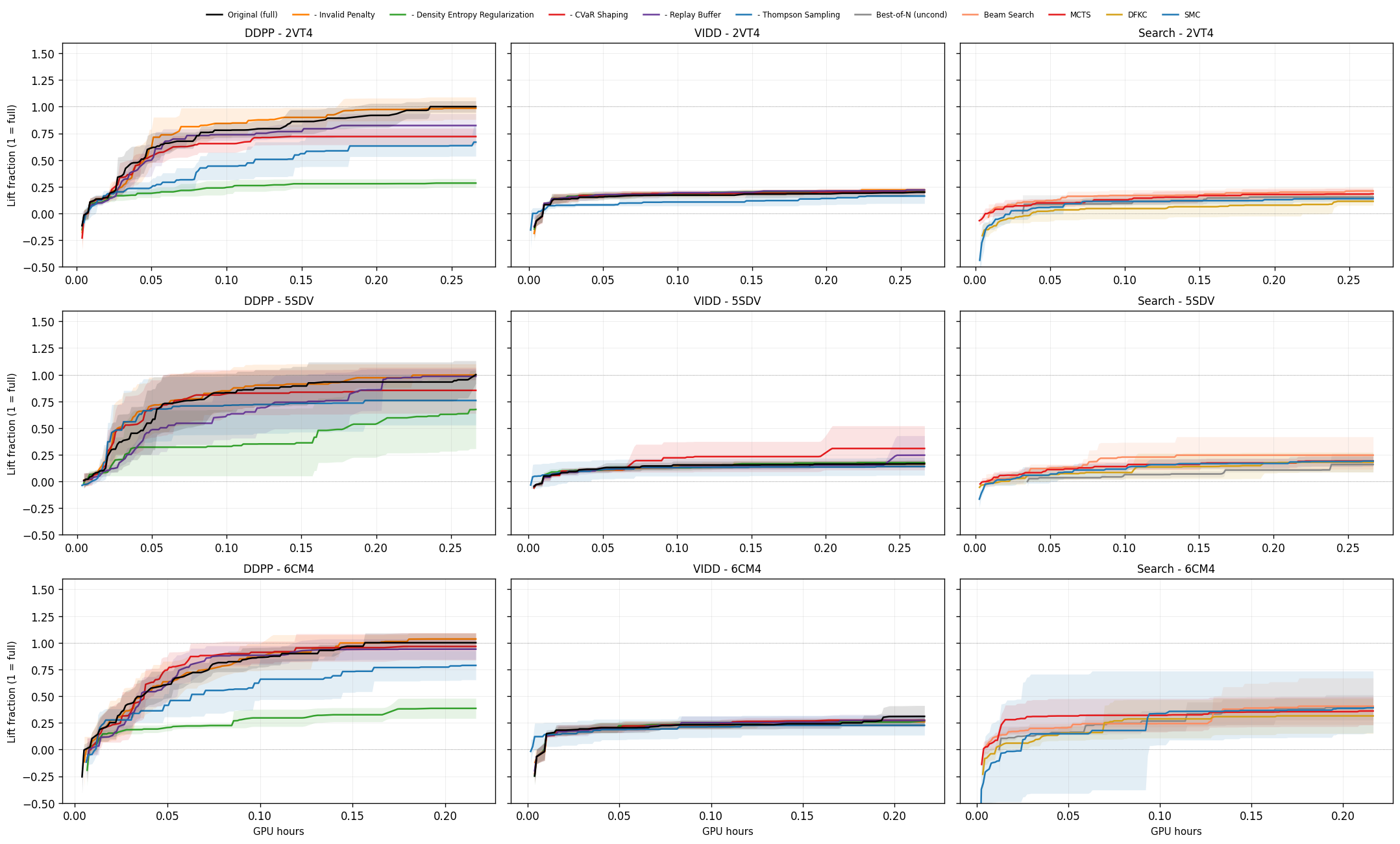}
  \caption{\textbf{Per-task ablation (mol set 1), GPU-hours view.} GPU-hour twin of Fig.~\ref{fig:per_task_mol_set1}.}
  \label{fig:per_task_mol_set1_gpuhrs}
\end{figure}

\begin{figure}[H]
  \centering
  \includegraphics[width=0.95\textwidth]{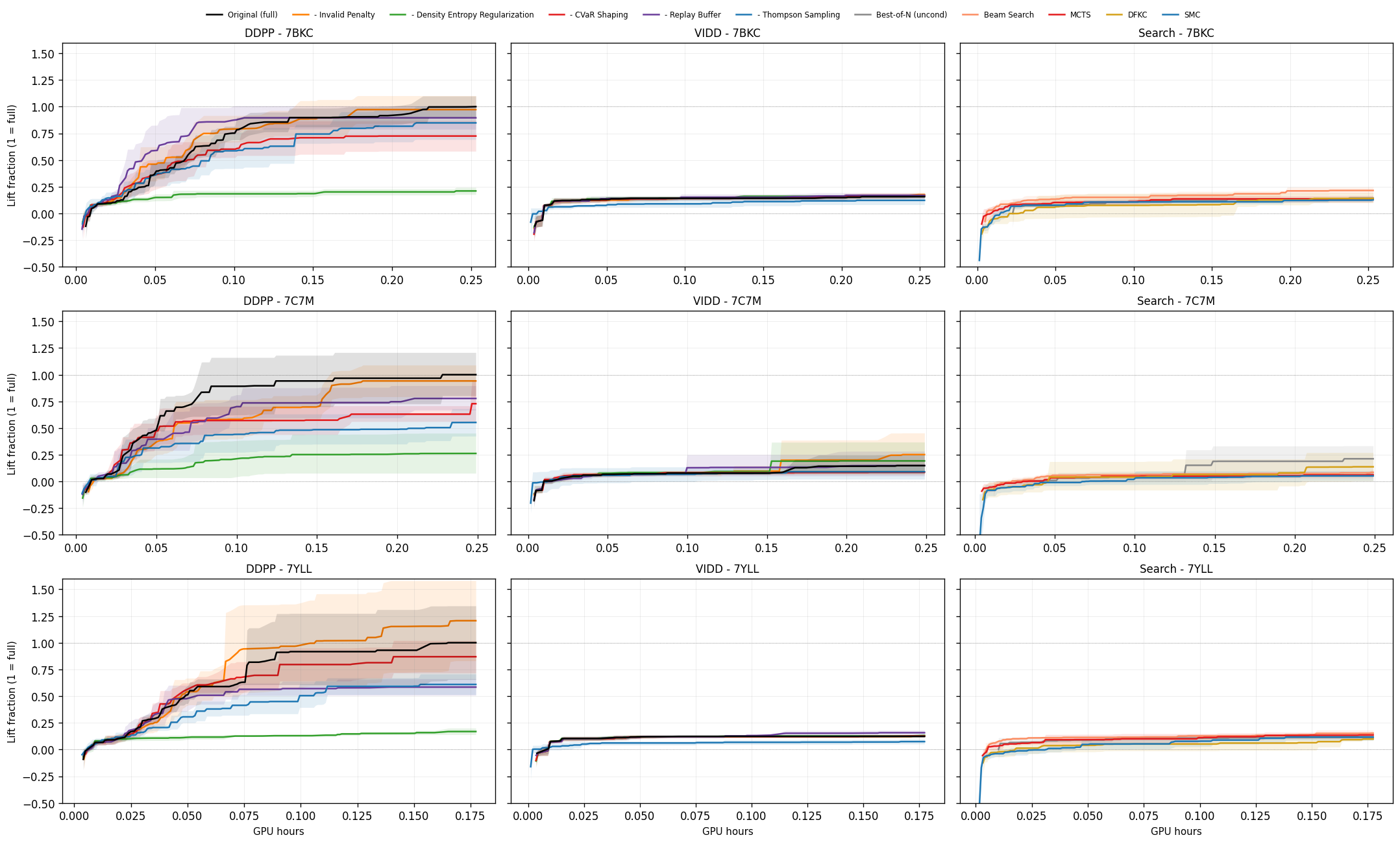}
  \caption{\textbf{Per-task ablation (mol set 2), GPU-hours view.} GPU-hour twin of Fig.~\ref{fig:per_task_mol_set2}.}
  \label{fig:per_task_mol_set2_gpuhrs}
\end{figure}

\begin{figure}[H]
  \centering
  \includegraphics[width=0.95\textwidth]{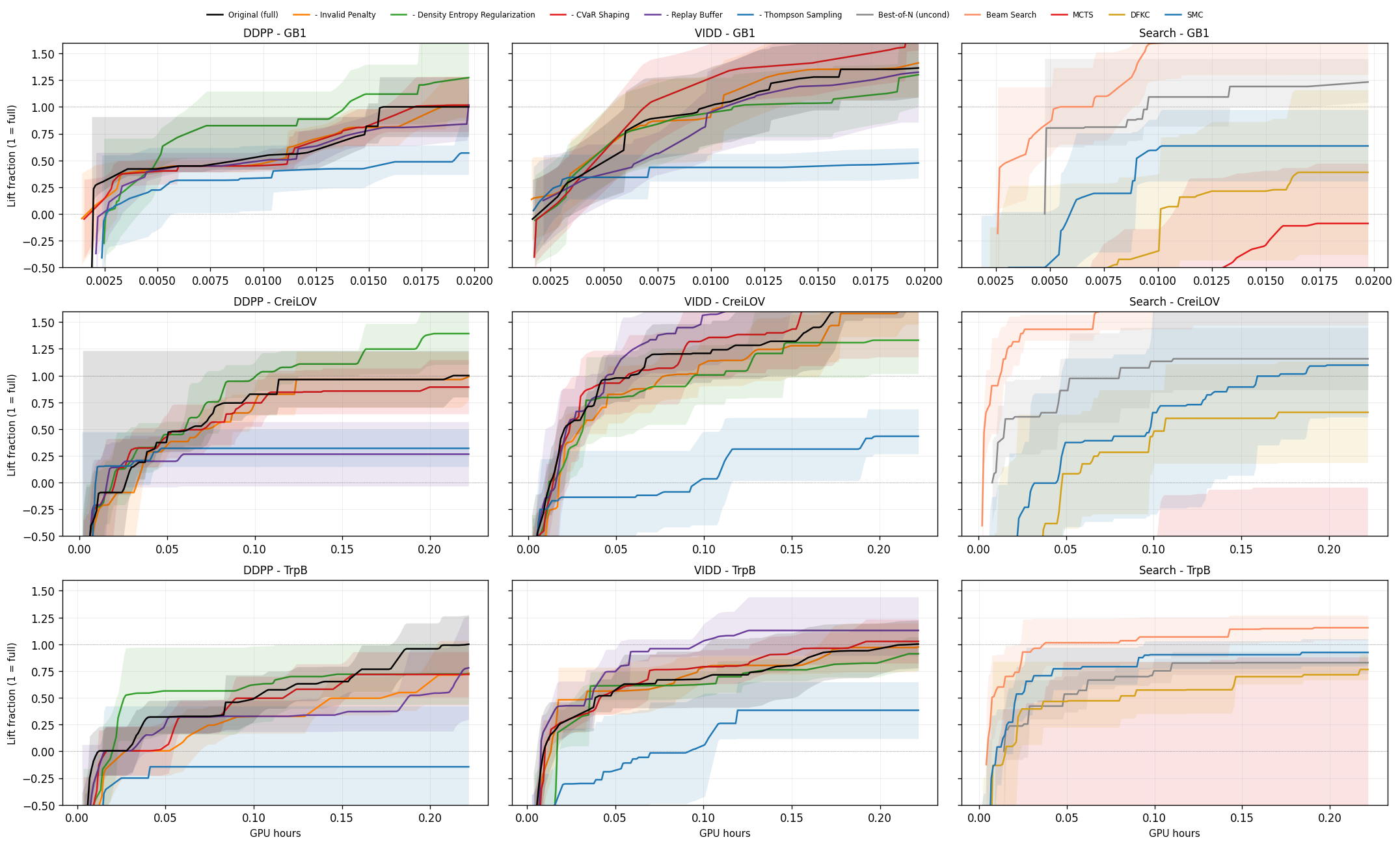}
  \caption{\textbf{Per-task ablation (proteins), GPU-hours view.} GPU-hour twin of Fig.~\ref{fig:per_task_protein}.}
  \label{fig:per_task_protein_gpuhrs}
\end{figure}


\subsection{Guide to the Main-Text Bar Plots}\label{app:per-target-grids}

The canonical bar plots are Fig.~\ref{fig:barplot_ablation} (component ablations) and Fig.~\ref{fig:barplot_inference_search} (full configuration versus search baselines) in the main text. They are referenced directly rather than reproduced here.

\clearpage

\section{Alternative Oracle and Qualitative Examples}\label{app:external-validation}

These experiments test whether the conclusions persist beyond the primary FlashAffinity evaluation and show representative valid outputs.

\subsection{Boltz-2 Oracle Results}\label{app:boltz}

The body claims that the design-space conclusions transfer from FlashAffinity (FA) to a structure-prediction-based oracle that is two orders of magnitude more expensive (Sec.~\ref{sec:exp-setup}). This appendix replicates each comparison from Appendices~\ref{app:search} and~\ref{app:knobs} under Boltz-2, with a $10$-hour wall budget per seed. The Boltz-2 panels use the same five seeds (\{0, 1, 2, 3, 42\}) and the same per-knob configurations as their FA counterparts; ablation cells that did not get re-run on Boltz-2 (annealed-CVaR variants, search-side Tweedie variants) are omitted. We show top-1 best-so-far against cumulative oracle calls (left) and GPU hours (right) only --- the top-10 / top-10\% rows from Appendix~\ref{app:knobs} are not separately rendered for Boltz-2. Across every panel below the relative ordering of methods is preserved, supporting the claim that the recipe is robust to oracle expense.

\begin{figure}[H]
  \centering
  \includegraphics[width=0.49\textwidth]{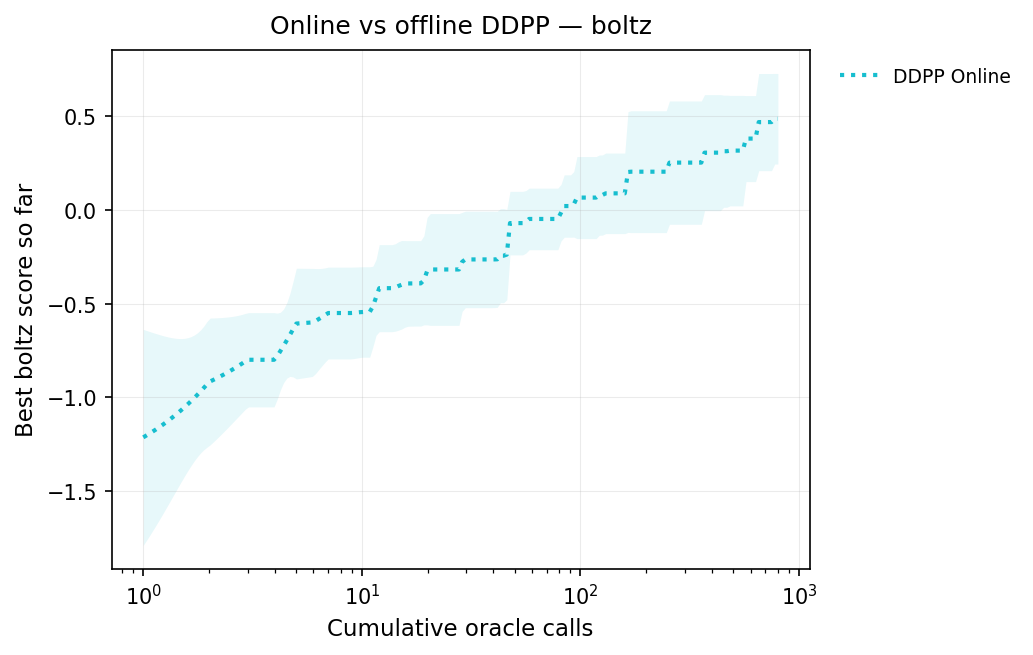}\hfill
  \includegraphics[width=0.49\textwidth]{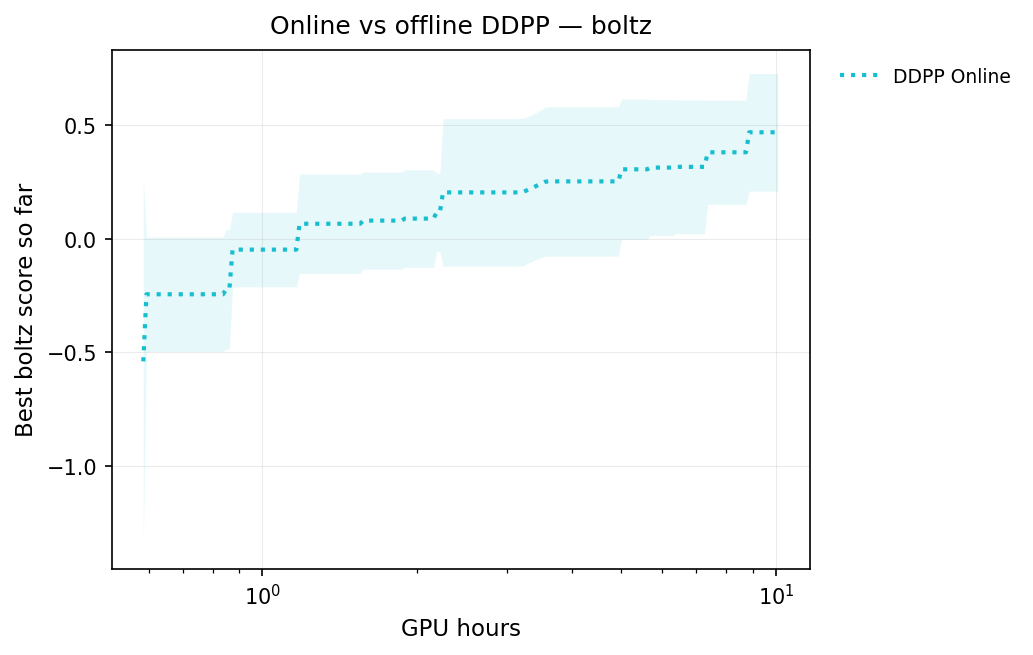}
  \caption{\textbf{Online vs.\ offline DDPP-LB --- Boltz-2.} Companion to Fig.~\ref{fig:online_vs_offline}. Same conclusion: online dominates offline at matched compute on top-1 best-so-far Boltz-2 score.}
  \label{fig:boltz_online_offline}
\end{figure}

\begin{figure}[H]
  \centering
  \includegraphics[width=0.49\textwidth]{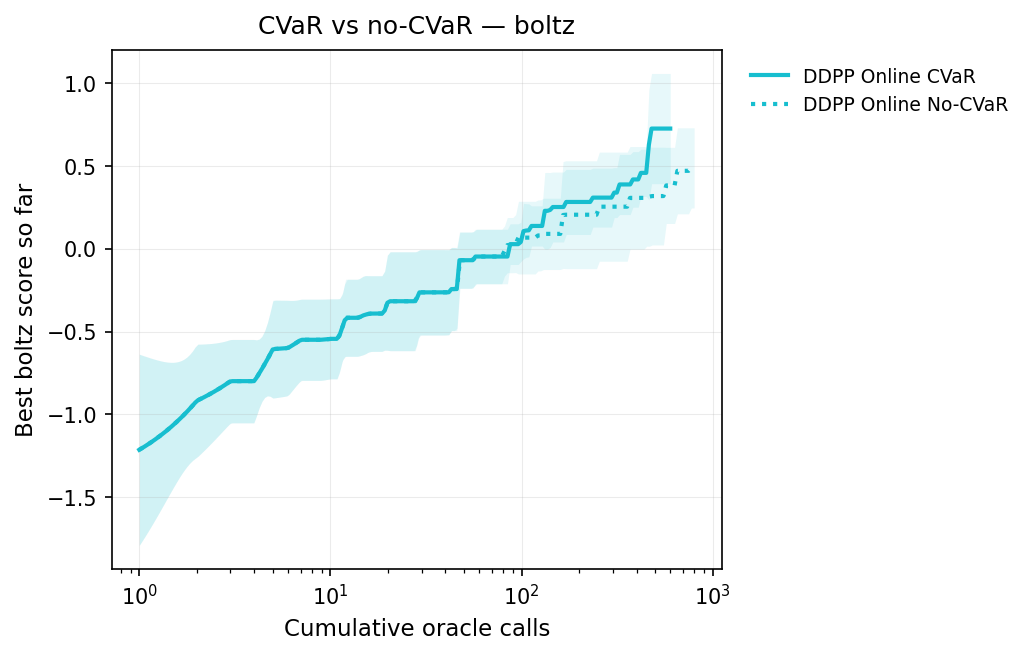}\hfill
  \includegraphics[width=0.49\textwidth]{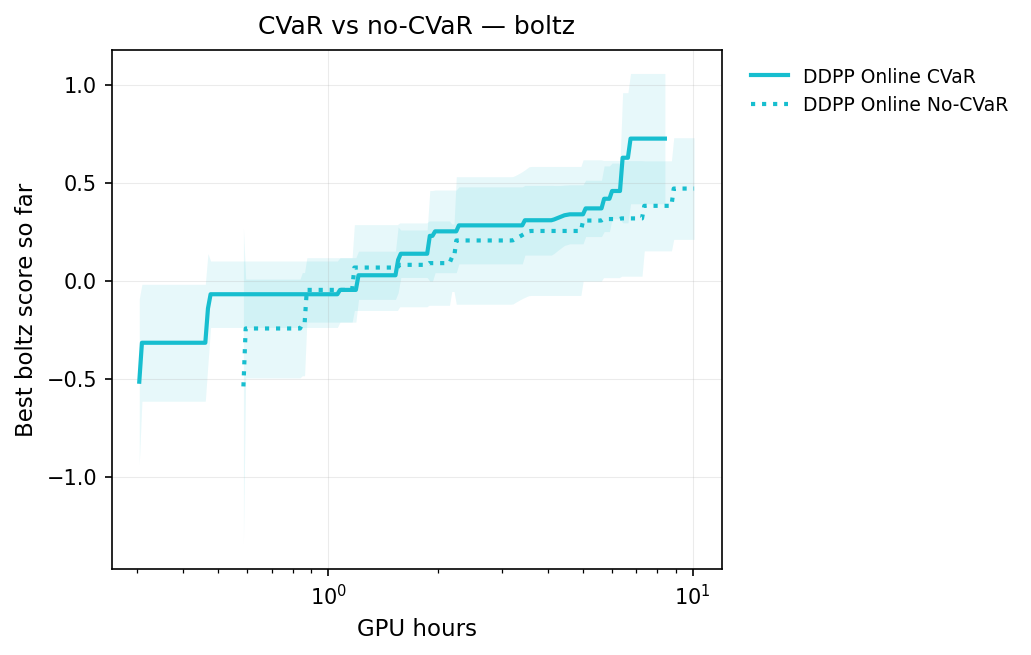}
  \caption{\textbf{CVaR ablation --- Boltz-2.} Companion to Fig.~\ref{fig:cvar_ablation}, online cells only (offline DDPP was not re-run on Boltz-2). CVaR-on solid, CVaR-off dotted. CVaR continues to lift top-1 by a comparable margin under Boltz-2 as under FA.}
  \label{fig:boltz_cvar}
\end{figure}

\begin{figure}[H]
  \centering
  \includegraphics[width=0.49\textwidth]{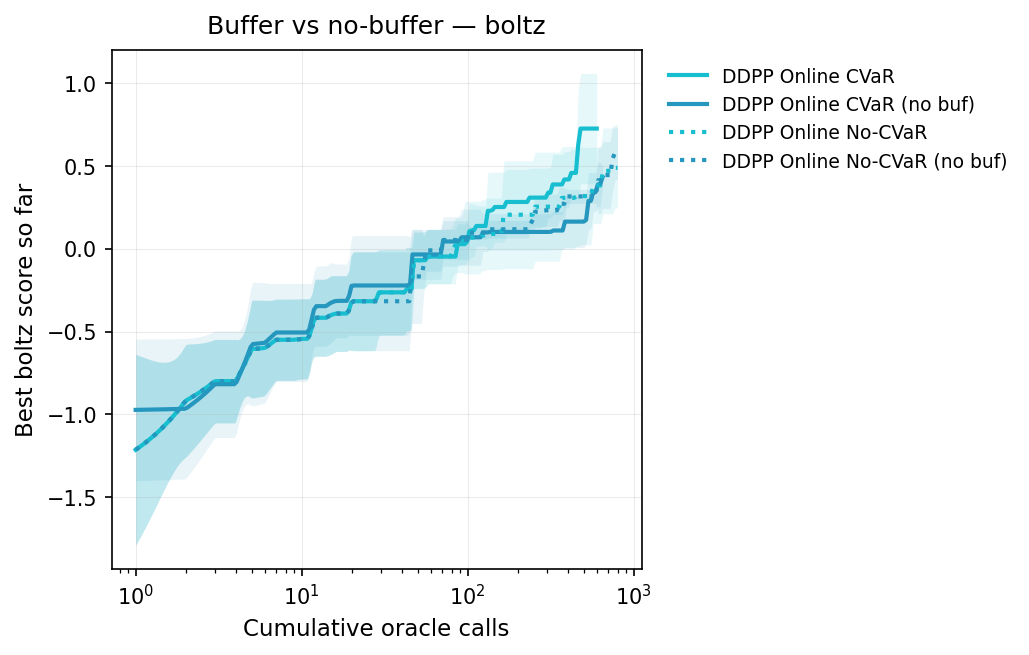}\hfill
  \includegraphics[width=0.49\textwidth]{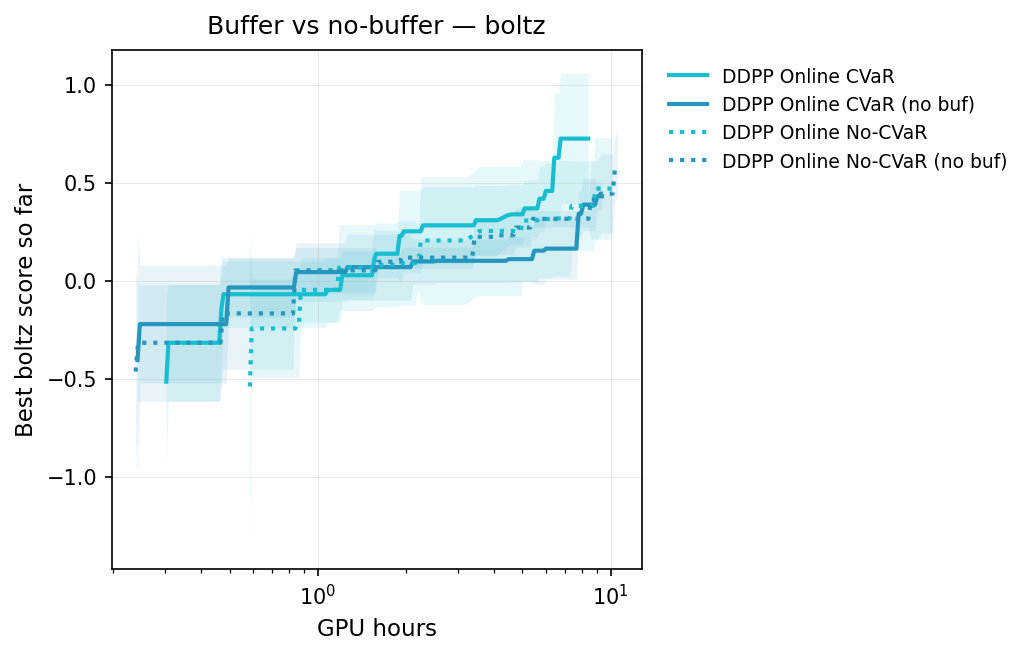}
  \caption{\textbf{Replay-buffer ablation --- Boltz-2.} Companion to Fig.~\ref{fig:buffer_ablation}. Same four cells as the FA version (buffer~$\times$~CVaR). The buffer-on / CVaR-on cell continues to lead.}
  \label{fig:boltz_buffer}
\end{figure}

\begin{figure}[H]
  \centering
  \includegraphics[width=0.49\textwidth]{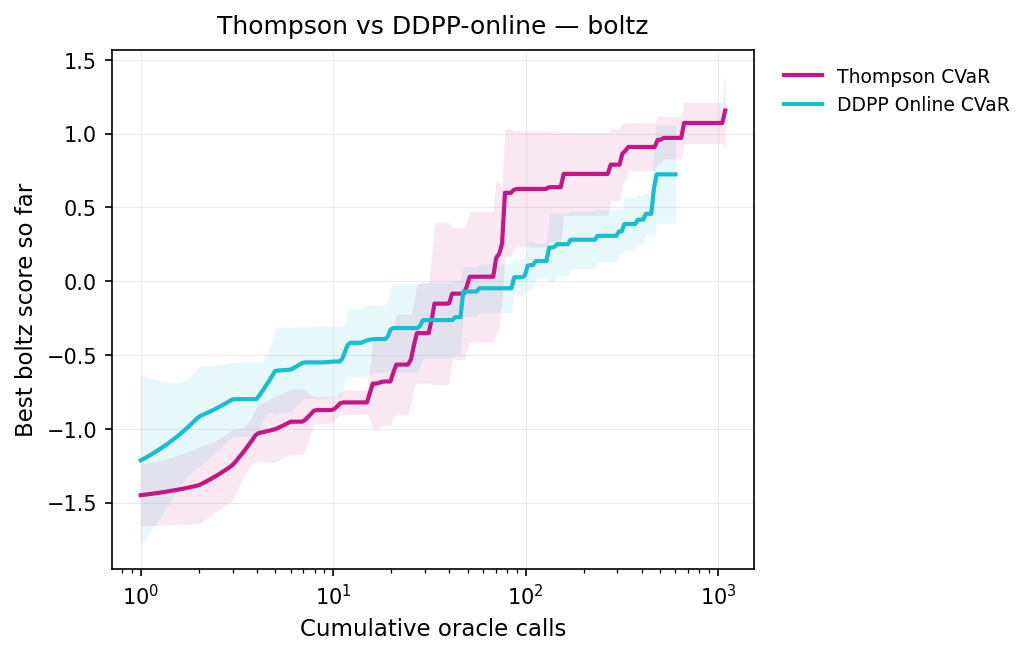}\hfill
  \includegraphics[width=0.49\textwidth]{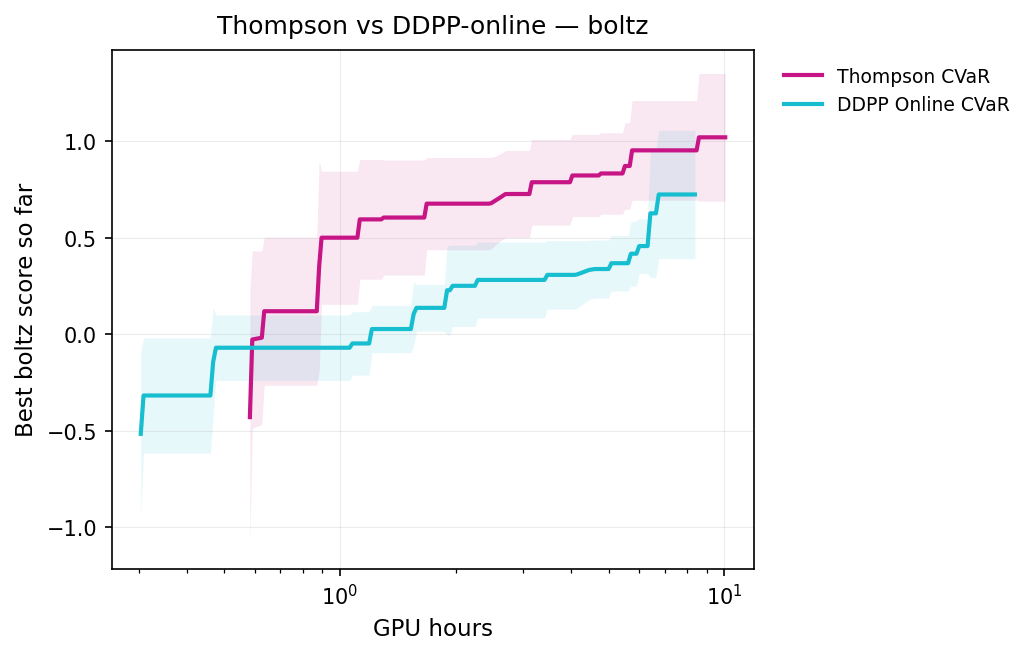}
  \caption{\textbf{Thompson-sampling ablation --- Boltz-2.} Companion to Fig.~\ref{fig:thompson_ablation}. Thompson on top of online DDPP-LB$+$CVaR vs.\ deterministic top-$K$ selection. Thompson retains its lead on top-1 under the slower oracle.}
  \label{fig:boltz_thompson}
\end{figure}

\begin{figure}[H]
  \centering
  \includegraphics[width=0.49\textwidth]{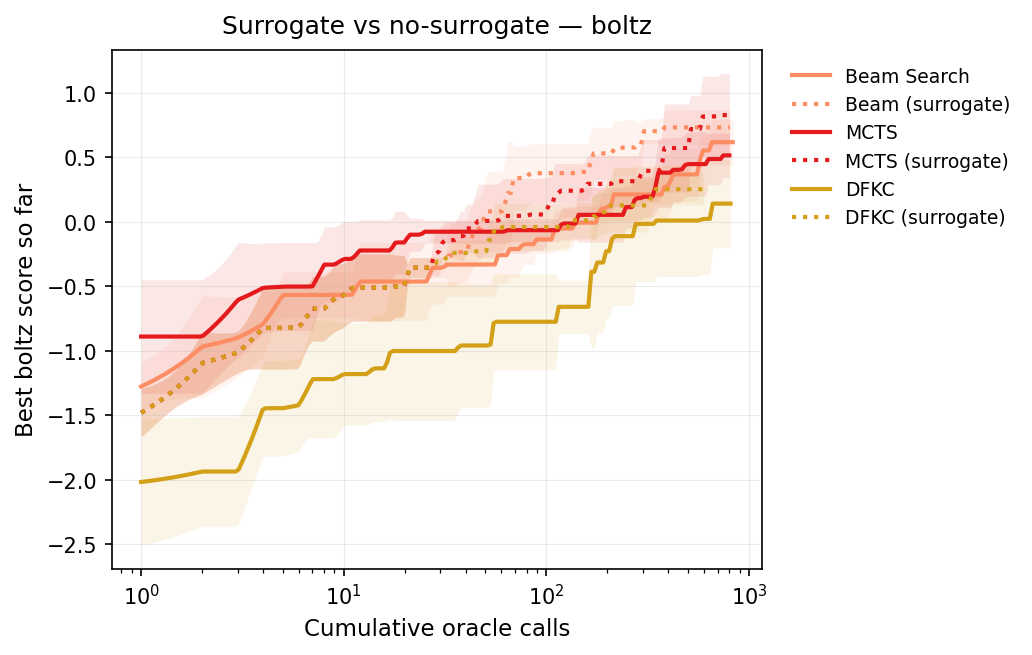}\hfill
  \includegraphics[width=0.49\textwidth]{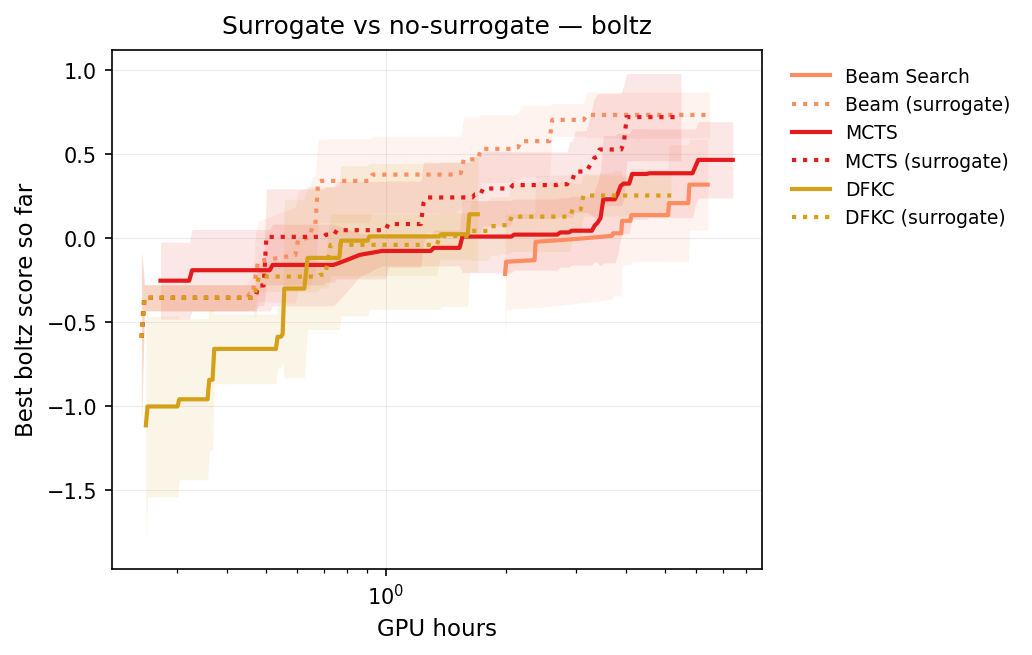}
  \caption{\textbf{Surrogate screening on inference-time search --- Boltz-2.} Companion to Fig.~\ref{fig:surrogate_search}. Beam, MCTS, and DFKC each compared against their surrogate-pre-filter variant. Surrogate screening continues to dominate plain search across all three families under Boltz-2.}
  \label{fig:boltz_surrogate}
\end{figure}

\begin{figure}[H]
  \centering
  \includegraphics[width=0.49\textwidth]{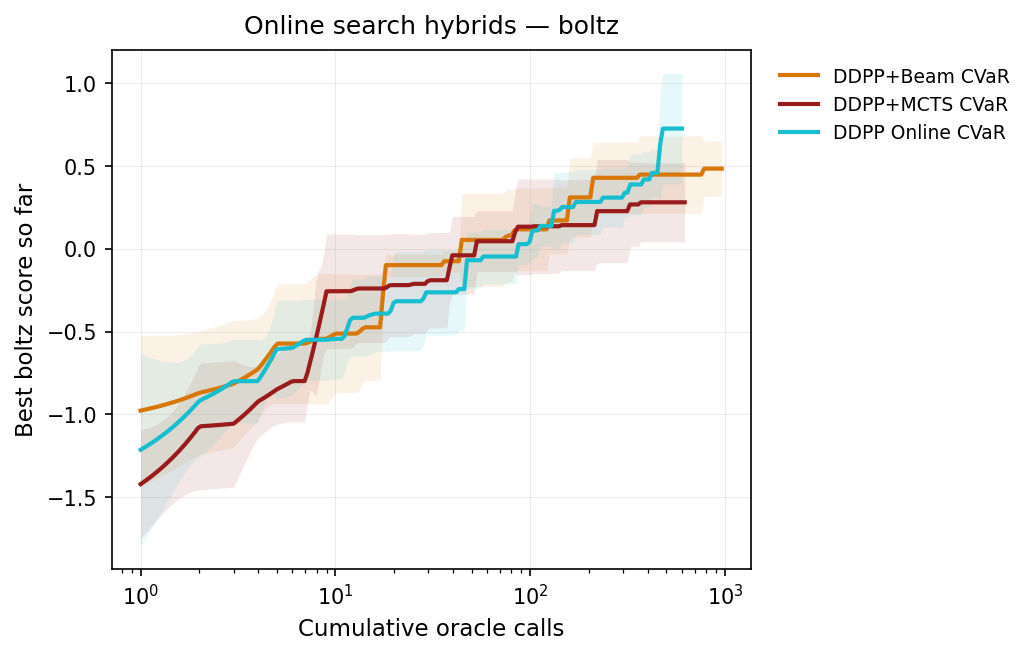}\hfill
  \includegraphics[width=0.49\textwidth]{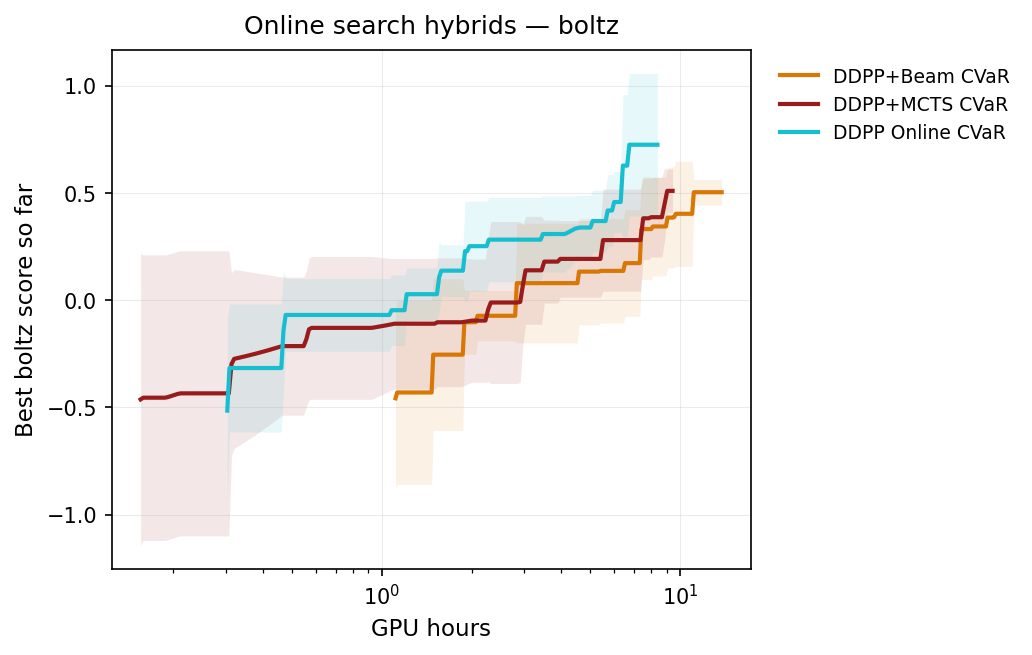}
  \caption{\textbf{Online search hybrids at standard Boltz-2 oracle cost.} Plain online DDPP-LB$+$CVaR vs.\ DDPP$+$beam$+$CVaR vs.\ DDPP$+$MCTS$+$CVaR. Plain online matches or exceeds both search-augmented variants on top-1 Boltz-2, mirroring the FA conclusion in Sec.~\ref{sec:results}.}
  \label{fig:boltz_oracle_1x}
\end{figure}

\begin{figure}[H]
  \centering
  \includegraphics[width=0.49\textwidth]{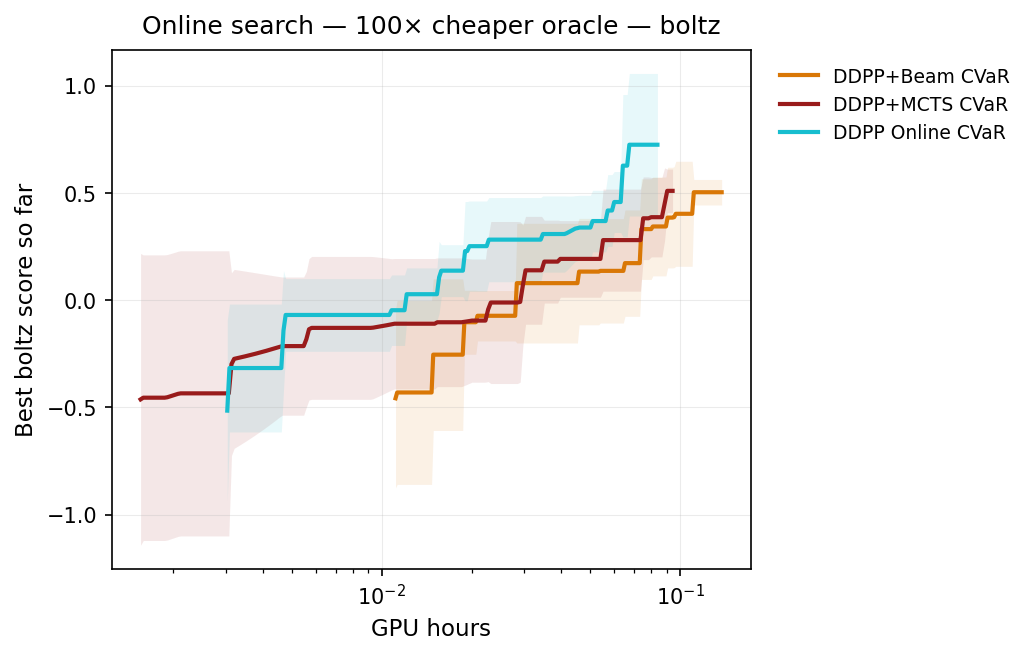}\hfill
  \includegraphics[width=0.49\textwidth]{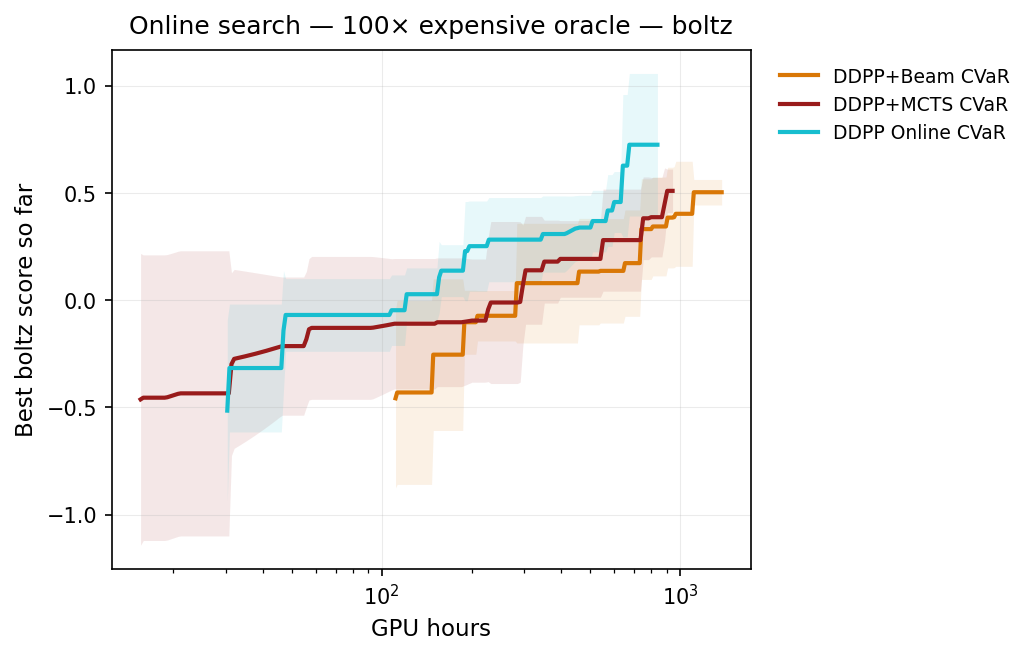}
  \caption{\textbf{Retroactive Boltz-2 oracle-cost rescaling.} Companion to Fig.~\ref{fig:oracle_rescale}, restricted to the GPU-hours axis. Per-call Boltz-2 cost rescaled by $1/100$ (left) and $100$ (right). The rescaling shifts the GPU-hours axis by four orders of magnitude in either direction without altering the relative ordering: plain online DDPP-LB$+$CVaR continues to match or exceed the search-augmented variants under both regimes, supporting the claim that the marginal value of inference-time search on top of online finetuning is robust to oracle expense even when the oracle is itself the expensive variant.}
  \label{fig:boltz_oracle_rescale}
\end{figure}

\subsection{Valid Molecule Samples Across Targets and Methods}\label{app:valid-samples}

For the FA-oracle small-molecule targets, we render
representative oracle-evaluated SMILES generated by the \textbf{DDPP-LB}
and \textbf{VIDD} full methods. Per (method, target) we display
three \emph{high-score} picks (top decile of all oracle calls) and three
\emph{low-score} picks (bottom decile). 

\begin{figure}[H]
\centering
\includegraphics[width=0.31\textwidth]{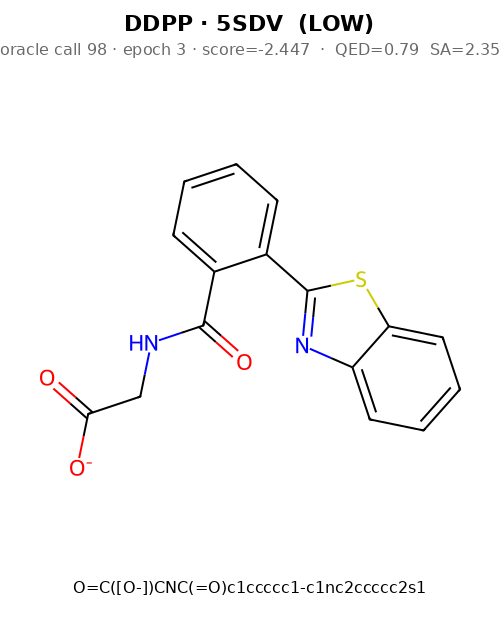}\hfill
\includegraphics[width=0.31\textwidth]{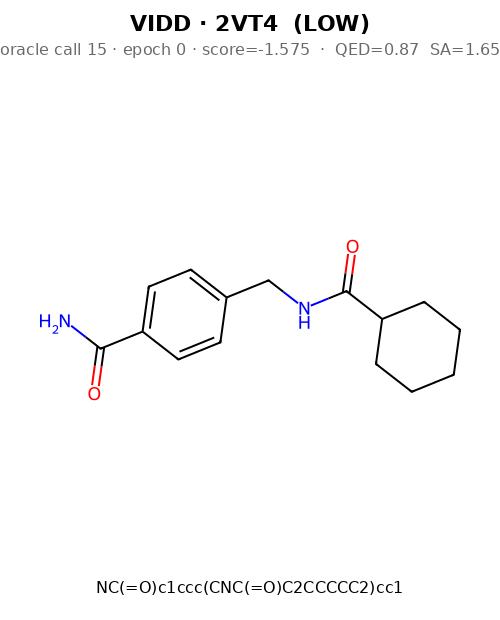}\hfill
\includegraphics[width=0.31\textwidth]{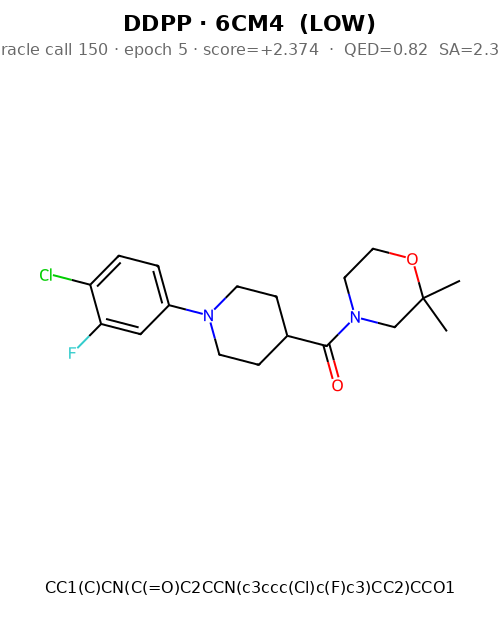}

\vspace{0.6em}

\includegraphics[width=0.31\textwidth]{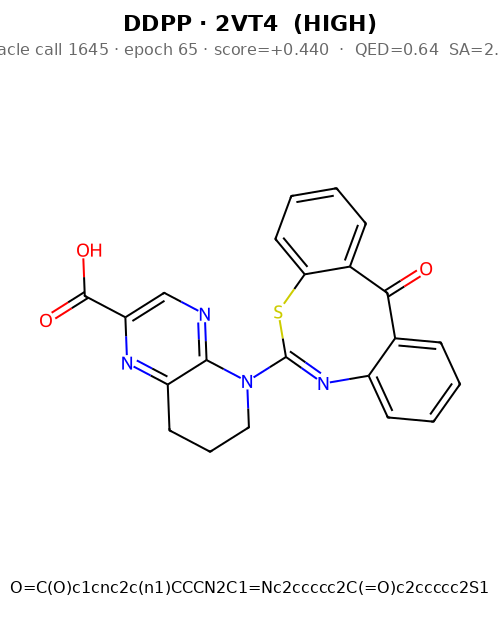}\hfill
\includegraphics[width=0.31\textwidth]{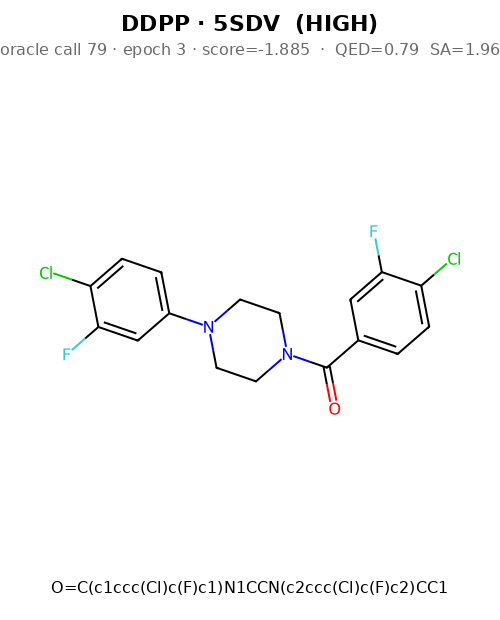}\hfill
\includegraphics[width=0.31\textwidth]{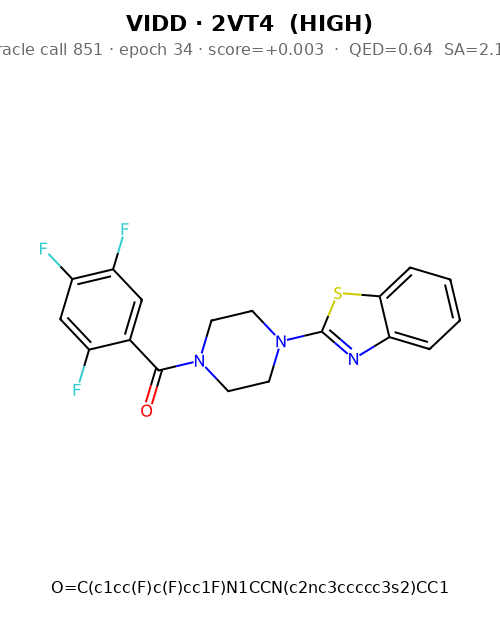}
\end{figure}

\clearpage

\section{Declaration of LLM Usage}
LLMs were used to write code, interpret results, assist with writing, and assist with figure making.

\end{document}